\documentclass[hidelinks,12pt]{article}
\usepackage{amsmath}

\usepackage[margin=1in]{geometry}

\usepackage{amssymb}

\usepackage{epsfig}
\usepackage{subcaption}
\usepackage{float}

\usepackage{chapterbib}
\usepackage{natbib}

\usepackage{etoc}

\bibpunct{(}{)}{;}{a}{,}{,}
\usepackage{setspace}

\newcommand{\blind}{1}

\usepackage{mathabx}

\usepackage{amsthm}

\newtheorem{theorem}{Theorem}
\newtheorem{lemma}[theorem]{Lemma}
\newtheorem{proposition}[theorem]{Proposition}
\newtheorem{corollary}[theorem]{Corollary}
\theoremstyle{definition}
\newtheorem{definition}{Definition}[section]

\theoremstyle{remark}

\newcommand{\ones}{1}

\DeclareMathOperator*{\argmin}{arg\,min}

\newcommand{\tr}[1]{\operatorname{tr}\left(#1\right)}
\newcommand{\HWMatProj}{\Phi}

\newcommand{\groupCenteredGenes}{\widetilde{J}_0}

\newcommand{\BFixed}{B_{n, m}}
\newcommand{\DeltaFixed}{\Delta_{n, m}}
\newcommand{\BinvFixed}{B^{-1}_{n, m}}
\newcommand{\OmegaFixed}{\Omega_{n, m}}

\newenvironment{subeqnarray}
{\begin{subequations}\begin{eqnarray}}%
{\end{eqnarray}\end{subequations}\hskip-4.0pt}

\def\fatnorm#1{|\kern-.2ex|\kern-.2ex| #1 |\kern-.2ex|\kern-.2ex|}
\newcommand{\twonorm}[1]{\left\lVert#1\right\rVert_2}

\newcommand{\fnorm}[1]{\left\lVert#1\right\rVert_{F}}
\newcommand{\norm}[1]{\left\lVert#1\right\rVert}
\newcommand{\abs}[1]{\left\lvert#1\right\rvert}

\newcommand{\Cond}{\kappa}
\newcommand{\cov}{\textsf{Cov}}

\newcommand{\off}{\text{off}}

\newcommand{\inv}[1]{\frac{1}{#1}}
\def\conv{\mathop{\text{\rm conv}\kern.2ex}}

\newcommand{\offzero}[1]{\ensuremath{\left|#1\right|_{0,\text{\off}}}}

\newcommand{\expct}[1]{\ensuremath{\mathbb E}\left(#1\right)}
\newcommand{\silent}[1]{}
\newcommand{\mvec}[1]{\rm{vec}\left\{\,#1\,\right\}}

\newcommand{\ve}{\varepsilon}

\def\qed{\hskip1pt $\;\;\scriptstyle\Box$}
\def\Ber{\mathop{\text{Bernoulli}\kern.2ex}}
\def\supp{\mathop{\text{supp}\kern.2ex}}
\def\corr{\mathop{\text{corr}\kern.2ex}}
\def\prec{\mathop{\text{precision}\kern.2ex}}
\def\recall{\mathop{\text{recall}\kern.2ex}}
\def\cov{\mathop{\text{Cov}\kern.2ex}}
\def\mnorm{\mathcal{N}_{f,m}\kern.2ex}
\def\var{\mathop{\text{Var}\kern.2ex}}
\def\ess{\mathop{\text{ess}\kern.2ex}}
\def\dom{\mathop{\text{dom}\kern.2ex}}
\def\lin{\mathop{\text{lin}\kern.2ex}}

\let\hat\widehat

\def\E{{\mathbb E}}

\def\supp{\mathop{\text{\rm supp}\kern.2ex}}
\def\argmin{\mathop{\text{arg\,min}\kern.2ex}}

\newcommand{\prob}[1]{\ensuremath{\mathbb P}\left(#1\right)}

\newcommand{\expf}[1]{\exp\left\{#1\right\}}
\newcommand{\beq}{\begin{equation}}
\newcommand{\eeq}{\end{equation}}
\newcommand{\ben}{\begin{eqnarray}}
\newcommand{\een}{\end{eqnarray}}
\newcommand{\bnum}{\begin{enumerate}}
\newcommand{\enum}{\end{enumerate}}
\newcommand{\bit}{\begin{itemize}}
\newcommand{\eit}{\end{itemize}}
\newcommand{\bens}{\begin{eqnarray*}}
\newcommand{\eens}{\end{eqnarray*}}
\newcommand{\X}{{\mathcal X}}

\newcommand{\init}{\text{\rm init}}
\newcommand{\e}{\epsilon}

\newcommand{\vp}{\varphi}

\def\qed{\hskip1pt $\;\;\scriptstyle\Box$}

\newenvironment{proofof}[1]{\hspace*{20pt}{\it Proof}{ of #1}.\hskip10pt}{\qed\vskip5pt}
\newenvironment{proofof2}{\hskip10pt}{\qed\vskip5pt}

\newcommand{\SOne}{\mathcal{E}_{\operatorname{I}}}
\newcommand{\STwo}{\mathcal{E}_{\operatorname{II}}}

\newcommand{\Parens}[1]{\left( #1 \right) }
\newcommand{\Brackets}[1]{\left[ #1 \right] }
\newcommand{\Braces}[1]{\left \{ #1 \right \} }
\newcommand{\InfNorm}[1]{\left\lVert#1\right\rVert_{\infty}}
\newcommand{\MaxNorm}[1]{\left \lVert #1\right \rVert_{\max}}
\newcommand{\OneNorm}[1]{ \left \lVert #1 \right \rVert_1}
\newcommand{\FroNorm}[1]{ \left \lVert #1 \right \rVert_F}
\newcommand{\Trace}[1]{\operatorname{tr}\left( #1 \right)}
\newcommand{\Vectorize}[1]{\operatorname{vec}\left\{ #1 \right \} }
\newcommand{\Ones}[1]{1_{#1}}
\newcommand{\Identity}[1]{I_{#1}}

\newcommand{\GroupIndicators}{\delta_n }

\newcommand{\GlobalCenIndices}{J_1}
\newcommand{\GroupCenIndices}{J_0}

\newcommand{\MGlobal}{M_{J_1}}

\newcommand{\SBGlobal}{S_1(B)}

\newcommand{\PGlobal}{P_1}
\newcommand{\ResidGlobal}{I - \PGlobal}
\newcommand{\NumGlobal}{m_1}

\newcommand{\SBGroup}{S_{2}(B)}

\newcommand{\PGroup}{P_2}
\newcommand{\ResidGroup}{I - \PGroup}
\newcommand{\NumGroup}{m_0}

\newcommand{\GlobalTwo}{S_{\operatorname{II}}}
\newcommand{\GlobalThree}{S_{\operatorname{III}}}

\newcommand{\EntryGlobalOne}[1]{(S_\text{I})_{#1}}

\newcommand{\EntryGlobalThree}[1]{(S_\text{III})_{#1}}

\newcommand{\NumFixedFalseNeg}{m_{01}}

\newcommand{\gammaGlobal}{\gamma_{J_1}}

\newcommand{\AGlobal}{A_{J_1}}
\newcommand{\EntryTrueA}[1]{a_{#1}}
\newcommand{\TrueB}{B}

\newcommand{\QuadraticFormGlobalCen}{\Phi_{i, j}}
\newcommand{\VecGlobalTwo}{\psi_{ij}}

\begin{document}
% \pdfminorversion=4

\if1\blind
{
  \title{Joint mean and covariance estimation with unreplicated
    matrix-variate data
\footnote{
Michael Hornstein is Ph.D. candidate, Department of Statistics, 
University of Michigan, Ann Arbor, MI 48109.
Roger Fan, Kerby Shedden, and Shuheng Zhou are alphabetically listed.
This work is done while all authors are affiliated with Department of Statistics, University of Michigan.
The research is supported in part by NSF under Grant DMS-1316731 and
the Elizabeth Caroline Crosby Research Award from the Advance Program 
at the University of Michigan. 
The authors thank the Editor, the Associate editor and three referees 
for their constructive comments that led to improvements in the paper. 
}}
  \author{Michael Hornstein, Roger Fan, Kerby Shedden, Shuheng Zhou
\hspace{.2cm} \\
 Department of Statistics, University of Michigan}
  \maketitle
} \fi

\if0\blind
{
  \bigskip
  \bigskip
  \bigskip
  \begin{center}
    {\LARGE\bf Joint mean and covariance estimation with unreplicated
    matrix-variate data}
\end{center}
  \medskip
} \fi

\def\spacingset#1{\renewcommand{\baselinestretch}%
{#1}\small\normalsize} \spacingset{1}

\bigskip
\begin{abstract} \normalsize
It has been proposed that complex populations, such as those that arise in genomics studies, may exhibit dependencies among observations as well as among variables.  This gives rise to the challenging problem of analyzing unreplicated high-dimensional data with unknown mean and dependence structures.  Matrix-variate approaches that impose various forms of (inverse) covariance sparsity allow flexible dependence structures to be estimated, but cannot directly be applied when the mean and covariance matrices are estimated jointly.  We present a practical method utilizing generalized least squares and penalized (inverse) covariance estimation to address this challenge.  We establish consistency and obtain rates of convergence for estimating the mean parameters and covariance matrices.  The advantages of our approaches are: (i) dependence graphs and covariance structures can be estimated in the presence of unknown mean structure, (ii) the mean structure becomes more efficiently estimated when accounting for the dependence structure among observations; and (iii) inferences about the mean parameters become correctly calibrated.  We use simulation studies and analysis of genomic data from a twin study of ulcerative colitis to illustrate the statistical convergence and the performance of our methods in practical settings.  Several lines of evidence show that the test statistics for differential gene expression produced by our methods are correctly calibrated and improve power over conventional methods.
\end{abstract}

\noindent%
{\it Keywords:}  two-group comparison, sparsity, genomics, generalized least squares, graphical modeling
\vfill

\newpage
\spacingset{1} % DON'T change the spacing!

%!TEX root = submit.arxiv.tex
\section{Introduction}
\label{sec:intro}

Understanding how changes in gene expression are related to changes in biological state
is one of the fundamental tasks in genomics research, and is a prototypical example of ``large scale inference'' \citep{efron2010large}.
While some genomics datasets have within-subject replicates or other known clustering factors that
could lead to dependence among observations, most are viewed as population cross-sections or convenience samples,
and are usually analyzed by taking observations (biological samples) to be statistically independent of each other.  Countering this conventional view, \citet{Efr09} proposed that there may be unanticipated correlations between samples even when the study design would not suggest it.  To identify and adjust for unanticipated sample-wise correlations,
\citet{Efr09} proposed an empirical Bayes approach utilizing the sample moments of the data.  In particular, sample-wise correlations may lead to inflated evidence for mean differences, and could be one explanation for the claimed lack of reproducibility in genomics research ~\citep{leek2010tackling, allen2012inference, sugden2013assessing}.

A persistent problem in genomics research is that test statistics for mean parameters (e.g. t-statistics for two-group comparisons) often appear to be incorrectly calibrated \citep{efron:05, allen2012inference}.  When this happens, for example when test statistics are uniformly overdispersed relative to their intended reference distribution, this is usually taken to be an indication of miscalibration, rather than reflecting a nearly global pattern of differential effects ~\citep{efron2007correlation}.  
Adjustments such as genomic control ~\citep{devlin1999genomic} can be used to account for this; a related approach is that of~\citet{allen2012inference}.
In this work we address unanticipated sample-wise dependence, which can exhibit a strong effect on statistical inference.  We propose a new method to jointly estimate the mean and covariance with a single instance of the data matrix, as is common in genetics.  The basic idea of our approach is to alternate for a fixed number of steps between mean and covariance estimation.  We exploit recent developments in two-way covariance estimation for matrix-variate data \citep{Zhou14a}.  We crucially combine the classical idea of generalized least squares (GLS)  ~\citep{aitken1936iv} with thresholding for model selection and estimation of the mean parameter vector.  Finally, we use Wald-type statistics to conduct inference.
We motivate this approach using differential expression analysis in a genomics context,
but the method is broadly applicable to matrix-variate data
having unknown mean and covariance structures, with or without replications.  We illustrate, using theory and data examples, including a genomic study of ulcerative colitis, that estimating and accounting for the sample-wise dependence can systematically improve the calibration of test statistics, therefore reducing or eliminating the need for certain post-hoc adjustments.

With regard to variable selection, another major challenge we face is that variables (e.g. genes or mRNA transcripts) have a complex dependency structure that exists together with any dependencies among observations.
As pointed out by \citet{Efr09} and others,  the presence of correlations among the samples makes
it more difficult to estimate correlations among variables, and vice versa.
A second major challenge is that due to dependence among both observations and variables,
there is no independent replication in the data, that is, we have a single matrix to conduct covariance estimation along both axes.  This challenge is addressed in \citet{Zhou14a} when the mean structure is taken to be zero.  A third major challenge that is unique to our framework is that covariance structures
can only be estimated after removing the mean structure,  a fact that is generally not considered
in most work on high dimensional covariance and graph estimation, where the population mean is taken to be zero.
We elaborate on this challenge next.

\subsection{Our approach and contributions}
Two obvious approaches for removing the mean structure in our setting are
to globally center each column of the data matrix (containing the data for one variable),
or to center each column separately within each group of sample points to be compared (subsequently referred to as ``group centering''). Globally centering each column, by ignoring the mean structure, may result in an estimated covariance matrix that is not consistent.
% as we illustrate in Figure \ref{fig::gemini_comp}.  
Group centering all genes, by contrast, leads to consistent covariance estimation, as shown in Theorem \ref{mainTheoremGroupCentering} with regard to Algorithm 1.
However, group centering all genes introduces extraneous noise when the true vector of mean differences is sparse.
We find that there is a complex interplay between the mean and covariance estimation tasks,
such that overly flexible modeling of the mean structure can introduce large systematic
errors in the mean structure estimation.
To mitigate this effect, we aim to center the data using a model selection strategy.
More specifically, we adopt  a model selection centering approach
in which only mean parameters having a sufficiently large effect size (relative to the dimension of the data) are targeted for removal.  This refined approach has theoretical guarantees and performs well in simulations.  The estimated covariance matrix can be used in uncertainty assessment and formal testing of mean parameters, thereby improving calibration of the inference.

In Section~\ref{sec::covEstimation}, we define the two group mean model, which is commonly used in the genomics literature, and introduce the GLS algorithm in this context.  We bound the statistical error for estimating each column of the mean matrix using the GLS procedure so long as each column of $X$ shares the same covariance matrix $B$, for which we have a close approximation.  It is commonly known that genes are correlated, so correlations exist across columns as well as rows of the data matrix.  In particular, in Theorem~\ref{thm::GLSFixedB} in Section~\ref{sec::GLSFixedBtheorem}, we establish consistency for  the GLS estimator given a deterministic $\widehat{B}$ which is close to $B$ in the operator norm, and present the rate of convergence for mean estimation for data generated according to a subgaussian model to be defined in Definition~\ref{defSubgaussian}.  Moreover, we do not impose a separable covariance model in the sense of \eqref{eq::matrix-normal-rep-intro}.  

 What distinguishes our model from those commonly used in the genomics literature is that we do not require that individuals are independent. Our approach to covariance modeling builds on the Gemini method \citep{Zhou14a},
which is designed to estimate a separable covariance matrix for data with two-way dependencies.
For matrices $A \in \mathbb{R}^{m \times m}$ and $B \in \mathbb{R}^{n \times n}$, 
the Kronecker product $A \otimes B \in \mathbb{R}^{mn \times nm}$ is the block matrix for which the $(i, j)$th block is $a_{ij}B$, for $i, j \in \{ 1, \ldots, m\}$. 
We say that an $n \times m$ random matrix $X$ follows a
matrix variate distribution with mean $M \in \mathbb{R}^{n \times m}$ and a separable covariance matrix
% $\Sigma = A \otimes B$, denoted as
\ben
\label{eq::matrix-normal-rep-intro}
X_{n\times m} \sim \mathcal{L}_{n,m}(M, A_{m \times m} \otimes
B_{n \times n}),
\een
 if $\mvec{X}$ has  mean
$\mvec{M}$ and covariance $\Sigma = A \otimes B$.
Here $\mvec{X}$ is formed
by stacking the columns of $X$ into a vector in $\mathbb{R}^{mn}$.
For the mean matrix 
$M$, we focus on the two-group setting to be defined in
\eqref{meanMatrixTwoGroups}.  Intuitively, $A$ describes the covariance between columns
while $B$  describes the covariance between rows of $X$.
%$E[ \varepsilon \varepsilon^T]$, and $E[\varepsilon^T \varepsilon]$, 
% $A$ and $B$ are not identifiable, because for any $\eta > 0$, $A 
% \otimes B = \eta A \otimes B/\eta$.  
Even with perfect knowledge of $M$,  we can only estimate $A$ and $B$ up to a scaling factor, as $A \eta \otimes \inv{\eta}B = A 
\otimes B$ for any $\eta > 0$, and hence this will be our goal and precisely what we mean when we say we are 
interested in estimating covariances $A$ and $B$.  However, this lack of 
 identifiability does not affect the GLS estimate, because the GLS 
 estimate is invariant to rescaling the estimate of $B^{-1}$. 

% In Section~\ref{sec::GroupCenter},
% we use Theorem~\ref{thm::GLSFixedB} to develop an algorithm in the setting where $\mvec{X}$ now follows a separable
% covariance model, for which we also propose two centering algorithms.  We illustrate their interactions with the iterative covariance estimation procedures in Section~\ref{sec::covEstimation} as well as in simulation studies.  In Section~\ref{sec::mainTheorem} we present rates of convergence for these methods, including a rate of convergence for the mean estimation using GLS in Theorem~\ref{sec::GLSFixedBtheorem}, a joint rate of convergence for the mean and covariance estimation
% using Algorithm 1 in Theorem~\ref{mainTheoremGroupCentering}, and the same for Algorithm 2 in Theorem~\ref{mainTheoremModSel}.
% Finally, in the analysis of the observation-wise sample covariance,
% we characterize an interesting bias-variance trade-off in the entry-wise error,
% which we make explicit in the supplement in Section~\ref{app::entrywise_sample_corr} (c.f. \eqref{BiasMatrixB} and \eqref{VarianceMatrixB}).
% There, we elaborate that the source of the bias comes from the statistical error in estimating
% the high-dimensional mean vectors.  To illustrate our methodology and theory, we apply the estimated mean and covariance matrices to gene expression data in Section \ref{sec::UCData}.

%For details, including simulations comparing confounder adjustment with our methods, see Section \ref{secComparison}.

\subsection{Related work} \label{sec::RelatedWork}

\citet{Efr09} introduced an approach for inference on mean differences
in data with two-way dependence.  His approach uses empirical
Bayes ideas and tools from large scale inference, and also explores
how challenging the problem of conducting inference on mean parameters is when there are uncharacterized dependences among samples.  We combine GLS and variable selection with matrix-variate techniques.
% \citet{allen2012inference} also consider this setting, but they use OLS throughout their iterations.
\citet{allen2012inference} also consider this question and develop a different approach that uses ordinary least squares (OLS) through the iterations, first decorrelating the residuals and then using OLS techniques again on this adjusted dataset.  
The confounder adjustment literature in genomics, including \citet{sun2012multiple} and \citet{wang2015confounder}, can also be used to perform large-scale mean comparisons in similar settings that include similarity structures among observations. These methods use the same general matrix decomposition framework, where the mean and noise are separated.  They exploit low-rank structure in the mean matrix, as well as using sparse approximation of OLS estimates, for example where thresholding.  Our model introduces row-wise dependence through matrix-variate noise, while the confounder adjustment literature instead assumes that a small number of latent factors also affect the mean expression, resulting in additional low-rank structure in the mean matrix. Web Supplement Section \ref{secComparison} contains detailed comparisons between our approach and these alternative methods.  

Our inference procedures are based on Z-scores and associated FDR
values for mean comparisons of individual variables.  While we account for sample-wise correlations, gene-gene correlations remain, which we regard as a nuisance parameter.  Our estimated correlation matrix among the genes can be used in future work in combination with the line of work that addresses FDR in the presence of gene correlations.  This relies on earlier work for false discovery rate estimation using correlated data, including \citet{owen2005variance, benjamini2001control, tony2011optimal, li2014rate, benjamini1995controlling,
  storey2003positive}.  Taking a different approach, \citet{hall2010innovated} develop the innovated higher criticism test
statistics to detect differences in means in the presence of correlations between genes.  Our estimated gene-gene correlation matrix can be used in combination with this approach; we leave this as future work.  Another line of relevant research has focused on hypothesis testing of high-dimensional means, exploiting assumed sparsity of effects, and developing theoretical results using techniques from high dimensional estimation theory.  Work of this type includes \citet{caihigh,
  chen2014twoSampleThresh, bai1996effect,chen2010two}. \citet{Hoff11} adopts a Bayesian approach, using a model that is a generalization of the matrix-variate normal distribution. 

Our method builds on the Gemini estimator introduced by
\citet{Zhou14a}, which estimates covariance matrices when both rows and
columns of the data matrix are dependent.  In the setting where correlations exist along only one axis of the array, researchers have
proposed various covariance estimators and studied their theoretical
and numerical properties \citep{BGA08, FFW09, FHT07, LF07, MB06,
  PWZZ09, RWRY08, RBLZ08, YL07, ZLW08}. Although we focus on the setting of Kronecker products, or separable
covariance structures, \citet{cai2015joint} proposed a covariance
estimator for a model with several populations, each of which may have
a different variable-wise covariance matrix.  Our methods can be generalized to this setting.  \citet{tan2014sparse} use a similar matrix-variate data setting as in~\eqref{eq::matrix-normal-rep-intro}, 
but perform biclustering instead of considering a regression problem with a known design matrix.

\subsection{Notation and organization}

Before we leave this section, we introduce the notation needed for the technical sections.  Let $e_1,\ldots, e_p$ be the canonical basis of $\mathbb{R}^p$.
For a matrix $A = (a_{ij})_{1\le i,j\le m}$, let $|A|$ denote the determinant and ${\rm tr}(A)$ be the trace of $A$.
Let $\norm{A}_{\max} =
\max_{i,j} |a_{ij}|$ denote  the entry-wise max norm.
Let $\norm{A}_{1} = \max_{j}\sum_{i=1}^m\abs{a_{ij}}$
denote the matrix $\ell_1$ norm.
The Frobenius norm is given by $\norm{A}^2_F = \sum_i\sum_j a_{ij}^2$.
Let $\varphi_{i}(A)$ denote the $i$th largest eigenvalue of $A$, with $\varphi_{\max}(A)$ and $\varphi_{\min}(A)$ denoting the largest and smallest eigenvalues, respectively. Let $\kappa(A)$ be the condition
number for matrix $A$.
Let $| A |_{1, \text{off}} = \sum_{i \neq j} |a_{ij} |$ denote the sum of the absolute values of the off-diagonal entries and let $| A |_{0, \text{off}}$ denote the number of non-zero off-diagonal entries.  Let $a_{\max} = \max_i a_{ii}$.
Denote by $r(A)$ the stable rank ${\fnorm{A}^2 }/{\twonorm{A}^2}$.
We write $\text{diag}(A)$ for a diagonal matrix with the same diagonal as
$A$.  Let $I$ be the identity matrix. We let $C, C_1, c, c_1, \ldots$ be positive constants which may change from line to line.
For two numbers $a, b$, $a \wedge b := \min(a, b)$ and $a \vee b := \max(a, b)$.
Let $(a)_+ := a \vee 0$.  For sequences $\{ a_n \}, \{ b_n\}$, we write $a_n =O(b_n)$ if $|a_n| \le C |b_n|$ for some positive absolute constant $C$ which is independent of $n$ and $m$ or sparsity parameters, and write $a_n \asymp b_n$ if $c|a_n| \le |b_n| \le C|a_n|$.  We write $a_n = \Omega(b_n)$ if $|a_n| \geq C|b_n|$ for some positive absolute constant $C$ which is independent of $n$ and $m$ or sparsity parameters.  We write $a_n = o(b_n)$ if $\lim_{n \rightarrow \infty} a_n/b_n = 0$.
For random variables $X$ and $Y$, let $X \sim Y$ denote that $X$ and $Y$ follow the same distribution.

The remainder of the paper is organized as follows.
In Section~\ref{sec::covEstimation}, we present our matrix-variate modeling framework and methods on joint mean and covariance estimation.  In particular, we propose two algorithms for testing mean differences based on two centering strategies.
In Section~\ref{sec::mainTheorem}, we present convergence rates for these methods.
In Theorems~\ref{mainTheoremGroupCentering} and~\ref{mainTheoremModSel}, we provide joint rates of convergence for mean and covariance estimation using Algorithms 1 and 2, respectively.
We also emphasize the importance of the design effect (c.f.\ equation \eqref{glsTestStats}) in testing and present theoretical results for estimating this quantity in Corollary~\ref{theoremInference} and Corollary~\ref{corDesignEffectAlg2}.
In Section~\ref{sec::Simulations}, we demonstrate through simulations that 
our algorithms can outperform OLS estimators in terms of accuracy and variable selection consistency.
In Section~\ref{sec::UCData}, we analyze a gene expression dataset, and show that our
method corrects test statistic overdispersion that is clearly present when using
sample mean based methods (c.f.\ Section \ref{inferenceGamma}).
We conclude in Section~\ref{sec::conclude}.  We place all technical proofs and additional simulation and data analysis results in the Web Supplement, which is organized as follows.  Sections~\ref{sec::simulationAppend} and ~\ref{sec::dataAppend} 
contain additional simulation and data analysis results. 
Section~\ref{sec::apppreliminary} contains some preliminary results and notation.
In Section~\ref{sec::proofsOfTheorems}, we prove Theorem~\ref{thm::GLSFixedB}.
In Sections~\ref{sec::ProofMainThmPartI} and~\ref{sec::proofsforTheorem2}
we prove Theorem~\ref{mainTheoremGroupCentering}.
In Section~\ref{app::entrywise_sample_corr}, we derive entry-wise rates
of convergence for the sample covariance matrices.
In Sections~\ref{sec::proofTheorem3} and \ref{sec::LemmasForTheorem3} we prove Theorem~\ref{mainTheoremModSel} and its auxiliary results.  In Section~\ref{secComparison} we provide additional comparisons between our method and some related methods on both simulated and real data.

%!TEX root = submit.arxiv.tex

\section{Models and methods}
\label{sec::covEstimation}

%We will state an initial theoretical characterization of the GLS-based approach for mean estimation in Section~\ref{sec::methodology}.  We present the rate of convergence for  mean estimation in the two-group model for subgaussian data in Theorem \ref{thm::GLSFixedB}.    We will show that  our theory and analysis works with a model much more general than the matrix variate normal model, which we will define in Section~\ref{sec::methodology}.  
In this section we present our model and method for joint mean and
covariance estimation.  Our results apply to subgaussian data.  Before
we present the model, we define subgaussian random vectors and the
$\psi_2$ norm.
The $\psi_2$  condition on a scalar random variable $V$ is equivalent to 
the subgaussian tail decay of $V$, which means 
$P(|V| >t) \leq 2 \exp(-t^2/c^2), \; \; \text{for all} \; \; t>0.$ 
For a vector $y = (y_1, \ldots, y_p) \in \mathbb{R}^p$, denote by
$\lVert y \rVert_2 = \sqrt{\sum_{i = 1}^p y_i^2}$.
\begin{definition}
\label{defSubgaussian}
Let $Y$ be a random vector in $\mathbb{R}^p$.
(a) $Y$ is called isotropic if for every $y \in \mathbb{R}^p$, $E[ | \langle Y, y \rangle |^2] = \lVert y \rVert_2^2$.
(b) $Y$ is $\psi_2$ with a constant $\alpha$ if   for every $y \in 
  \mathbb{R}^p$,
$$\lVert \langle Y, y \rangle \rVert_{\psi_2} := \;
\inf \{t: E[\exp( \langle Y,y \rangle^2/t^2) ] \leq 2 \}
\; \leq \; \alpha \lVert y \rVert_2.$$
\end{definition}
Our goal is to estimate the group mean vectors $\beta^{(1)}, \beta^{(2)}$, the vector of mean differences between two groups $\gamma = \beta^{(1)} - \beta^{(2)} \in \mathbb{R}^m$, the row-wise covariance matrix $B \in \mathbb{R}^{n \times n}$, and the column-wise covariance matrix $A \in \mathbb{R}^{m \times m }$.  In our motivating genomics applications, the people by people covariance matrix $B$ is often incorrectly anticipated to have a simple known structure,
for example, $B$ is taken to be diagonal if observations are assumed to be uncorrelated.
However, we show by example in Section \ref{sec::UCData} that departures from the anticipated
diagonal structure may occur, corroborating earlier claims of this type by \citet{Efr09} and others.  Motivated by this example, we define the two-group mean model and the GLS algorithm, which takes advantage of the covariance matrix $B$.

\noindent{\bf The model.}
Our model for the matrix-variate data $X$ can be expressed as a mean matrix plus a noise term,
\begin{equation}
\label{modelMeanCov}
  X = M +  \varepsilon,
\end{equation}
where columns (and rows) of $\varepsilon$ are subgaussian.  Let $u, v, \in \mathbb{R}^n$ be defined as
\begin{equation}
\label{groupIndicators}
u = ( \underbrace{1, \ldots, 1}_{n_1}, \underbrace{0, \ldots,
  0}_{n_2}) \in \mathbb{R}^n \quad \text{and} 
\quad v = ( \underbrace{0, \ldots, 0}_{n_1}, \underbrace{1, \ldots, 1}_{n_2}) \in \mathbb{R}^n.
\end{equation}
Let $\mathbf{1}_n \in \mathbb{R}^n$ denote a vector of ones.  
For the two-group model, we take the mean matrix to have the form
\begin{equation}  
\label{meanMatrixTwoGroups}
M = D \beta = \begin{bmatrix}
\mathbf{1}_{n_1} \beta^{(1)T} \\
\mathbf{1}_{n_2} \beta^{(2)T}
\end{bmatrix} \in \mathbb{R}^{n \times m}, \quad \text{where} \quad D = \begin{bmatrix}
	u & v
	\end{bmatrix} \in \mathbb{R}^{n \times 2}
\end{equation}
is the design matrix and $\beta = (\beta^{(1)}, \beta^{(2)})^T \in \mathbb{R}^{2 \times m}$
is a matrix of group means.  Let $\gamma = \beta^{(1)} - \beta^{(2)} \in \mathbb{R}^m$ denote the vector of mean differences.  Let $d_0 = |\operatorname{supp}(\gamma)| = | \{ j: \gamma_j \neq 0 \}|$ denote the size of the support of $\gamma$.  To estimate the group means, we use a GLS estimator,
\begin{equation} 
\label{GLSestimator}
  \widehat{\beta}(\widehat{B}^{-1}) := (D^T \widehat{B}^{-1} D)^{-1} D^T \widehat{B}^{-1} X \in \mathbb{R}^{2 \times m},
\end{equation}
where $\widehat{B}^{-1}$ is an estimate of the observation-wise inverse covariance matrix.
Throughout the paper, we denote by $\widehat{\beta}(B^{-1})$ the
oracle GLS estimator, since it depends on the unknown true covariance
$B$.  Also, we denote the estimated vector of mean differences as
$\widehat{\gamma}(\widehat{B}^{-1}) = \delta^T
\widehat{\beta}(\widehat{B}^{-1}) \in \mathbb{R}^m$, where $\delta =
(1, -1) \in \mathbb{R}^2$.  

\subsection{Matrix-variate covariance modeling}
In the previous section,  we have not yet explicitly constructed an estimator of $B^{-1}$.  To address this need, we model the data matrix $X$ with a
matrix-variate distribution having a separable covariance matrix, namely,
the covariance of $\mvec{X}$
 follows a Kronecker product covariance model.
When  $\varepsilon$ \eqref{modelMeanCov} follows a matrix-variate normal 
distribution $\mathcal{N}_{n, m}(0, A \otimes B)$, as considered in 
~\citet{Zhou14a}, the support of $B^{-1}$ encodes conditional 
independence relationships between samples, and 
likewise, the support of $A^{-1}$ encodes conditional independence
relationships among genes.
The inverse covariance matrices $A^{-1}$ and $B^{-1}$ have the same supports as their respective correlation matrices, 
so edges of the dependence graphs are identifiable under the model
$\text{Cov}(\text{vec}(\varepsilon)) = A \otimes B$.  When the data is
subgaussian, the method is still valid for obtaining consistent
estimators of $A$, $B$, and their inverses, but the interpretation in
terms of conditional independence does not hold in general.

Our results do not assume normally distributed data; we analyze the
subgaussian correspondent of the matrix variate normal model instead.
In the Kronecker product  covariance model we consider in the present
work, the noise term has the form $ \varepsilon =  B^{1/2} Z A^{1/2}$
for a mean-zero random matrix $Z$ with independent 
subgaussian entries satisfying $1 = \E Z_{ij}^2 \le \norm{Z_{ij}}_{\psi_2} \leq K$.
Clearly, $\mvec{\ve} = A \otimes B$.   Here, the matrix $A$ represents
the shared covariance among variables for each sample, while $B$
represents the covariance among observations which in turn is shared
by all genes.

For identifiability, and convenience, we define
\begin{equation} 
\label{KroneckerIdentifiability}
	A^* = \frac{m}{\operatorname{tr}(A)}A \quad \text{ and }  \quad B^* = \frac{\operatorname{tr}(A)}{m} B,
\end{equation}
where the scaling factor is chosen so that $A^*$ has trace $m$.  
For  the rest of the paper $A$ and $B$ refer to $A^*$ and $B^*$, as defined in \eqref{KroneckerIdentifiability}. 
Let $S_A$ and $S_B$ denote sample covariance matrices to be
specified. Let the corresponding sample correlation matrices be defined as
\ben
\label{defGammaAGammaB}
\widehat{\Gamma}_{ij}(A) = \frac{(S_A)_{ij} }{ \sqrt{(S_A)_{ii}(S_A)_{jj}}}
\quad \text{ and } \quad \widehat{\Gamma}_{ij}(B) = \frac{(S_B)_{ij} }{ \sqrt{(S_B)_{ii}(S_B)_{jj}}}.
\een
The baseline Gemini estimators ~\citep{Zhou14a} are defined as follows,
using a pair of penalized estimators for the correlation matrices $\rho(A) = (a_{ij}/\sqrt{a_{ii} a_{jj}})$
and $\rho(B) = (b_{ij}/\sqrt{b_{ii} b_{jj}})$,
\begin{subeqnarray}
\label{geminiObjectiveFnA}
	\widehat{A}_\rho &= &
\argmin_{A_\rho \succ 0} \left\{ \tr{\widehat{\Gamma}(A) A_\rho^{-1}}
  + \log |A_\rho|
+ \lambda_B |A_\rho^{-1}|_{1, \text{off}} \right\}, \; \text{ and }  \\
\label{geminiObjectiveFnB}
\widehat{B}_\rho &= &
\argmin_{B_\rho \succ 0} \left\{ \tr{\widehat{\Gamma}(B) B_\rho^{-1}}
  + \log |B_\rho| + \lambda_A |B_\rho^{-1}|_{1, \text{off}} \right\},
\end{subeqnarray}
where the input are a pair of sample correlation matrices as defined in \eqref{defGammaAGammaB}.

Let $\widehat{M}$ denote the estimator of the mean matrix $M$ in
\eqref{eq::matrix-normal-rep-intro}.  
Denote the centered data matrix and the sample covariance matrices as
\ben
\nonumber
X_{\operatorname{cen}} & = & X - \widehat{M}, \quad \text{for
  $\widehat{M}$ to be specified in Algorithms 1 and 2,} \\
 \label{SampleCovAB}
S_B & = & X_{\operatorname{cen}} X_{\operatorname{cen}}^T / m, 
\quad \text{ and } \quad S_A = X_{\operatorname{cen}}^T X_{\operatorname{cen}} / n.
\een
Define the diagonal matrices of sample standard deviations as
\ben
 \label{W1hatW2hat}
&& \widehat{W}_1 = \sqrt{n} \text{diag}(S_A)^{1/2} \in \mathbb{R}^{m
  \times m}, \quad \widehat{W}_2 = \sqrt{m}
\text{diag}(S_B)^{1/2} \in \mathbb{R}^{n \times n},\\
&&
\label{estimatorAKroneckerB} \text{ and } \;
\widehat{A \otimes B} = \left( \widehat{W}_1 
          \widehat{A}_{\rho} \widehat{W}_1 \right) \otimes 
\left( \widehat{W}_2 \widehat{B}_{\rho} \widehat{W}_2 \right) /
\lVert X_{\operatorname{cen}} \rVert_F^2.
\een
\subsection{Group based centering method}
\label{sec::GroupCenter}
We now discuss our first method for estimation and
inference with respect to the vector of mean differences $\gamma =
\beta^{(1)} - \beta^{(2)}$, for $\beta^{(1)}$ and $\beta^{(2)}$ as in
\eqref{meanMatrixTwoGroups}.  Our approach in Algorithm 1 is to remove all
possible mean effects by centering each variable within every group.

\noindent{\textbf{Algorithm 1: GLS-Global group centering} }\\
\noindent\rule{16cm}{0.4pt} \\
Input: $X$; and $\mathcal{G}(1), \mathcal{G}(2)$: indices of group one and two, respectively. \\
Output: $\widehat{A}^{-1}$, $\widehat{B}^{-1}$, $\widehat{A \otimes
  B}$, $\widehat{\beta}(\widehat{B}^{-1})$, $\widehat{\gamma}$, $T_j$ for all $j$ \\
\noindent\rule{16cm}{0.4pt}

\begin{description}
\item [1. Group center the data.]  Let $Y_i$ denote the $i$th row of the data matrix.
To estimate the group mean vectors $\beta^{(1)}, \beta^{(2)} \in \mathbb{R}^m$: Compute sample mean vectors
\ben
\label{eq::sampleMean}
\widetilde{\beta}^{(1)}  =  \frac{1}{n_1} \sum_{i \in \mathcal{G}(1)}
&&
Y_i
\quad \text{ and } \quad \widetilde{\beta}^{(2)} =  \frac{1}{n_2}
\sum_{i \in \mathcal{G}(2)} Y_i;
\quad \text{set} \quad \widehat{\gamma}^{\operatorname{OLS}} =
\widetilde{\beta}^{(1)} - \widetilde{\beta}^{(2)}.  \\
\nonumber
\text{Center the data by } &&  X_{\text{cen}} = X - \widehat{M}, \text{ with } \widehat{M} =
\begin{bmatrix} 1_{n_1} \widetilde{\beta}^{(1)T} \\
1_{n_2} \widetilde{\beta}^{(2)T}
\end{bmatrix}. \notag
\een
\item [2.  Obtain regularized correlation estimates.]
\begin{enumerate} 
\item[(2a)]
The centered data matrix used to calculate $S_A$ and $S_B$ for
Algorithm 1 is $X_{\operatorname{cen}} = (I - P_2)X$, where $P_2$ is the projection matrix that performs within-group centering, 
\begin{equation} 
\label{def:withinGroupProjection}
	P_2 = \begin{bmatrix}
	n_1^{-1} 1_{n_1} 1_{n_1}^T & 0 \\
	0 & n_2^{-1} 1_{n_2} 1_{n_2}^T 
	\end{bmatrix} = uu^T / n_1 + vv^T / n_2,
\end{equation}
with $u$ and $v$ as defined in \eqref{groupIndicators}.
 Compute sample covariance matrices based on group-centered 
  data: 
$S_A=  \inv{n}X_{\text{cen}}^T X_{\text{cen}} = \inv{n}X^T (I - P_2) 
X$\; \text{ and} \\
$S_B =  \inv{m} X_{\text{cen}} X_{\text{cen}}^T = \inv{m}(I -
P_2)XX^T(I - P_2)$.
 \item[(2b)]
Compute~\eqref{defGammaAGammaB} 
to obtain penalized correlation matrices $\widehat{A}_\rho$ and $\widehat{B}_\rho$
using the Gemini estimators as defined in (\ref{geminiObjectiveFnA}) and
 (\ref{geminiObjectiveFnB}) with tuning parameters to be defined
 in~\eqref{GLassoPenaltyB}.
\end{enumerate}
\item [3. Rescale the estimated correlation matrices to  obtain  penalized covariance]
\ben
\label{BiHat}
	\widehat{B}^{-1} = m \widehat{W}_2^{-1} \widehat{B}_{\rho}
        \widehat{W}_2^{-1} \; 
\text{ and  }  \; \widehat{A}^{-1} = (\lVert X_{\operatorname{cen}}
\rVert_F^2 / m) \widehat{W}_1^{-1} \widehat{A}_{\rho}
\widehat{W}_1^{-1}.
\een
\item [4. Estimate the group mean matrix] using the GLS estimator as defined in \eqref{GLSestimator}.
\item [5. Obtain test statistics.]  The $j$th test statistic is defined as
\begin{equation} 
\label{glsTestStats}
	T_j = \frac{\widehat{\gamma}_j(\widehat{B}^{-1}) }{ \sqrt{\delta^T (D^T \widehat{B}^{-1} D)^{-1} } \delta}, \qquad \text{with } \delta = (1, -1) \in \mathbb{R}^2,
\end{equation}
and  $\widehat{\gamma}_j(\widehat{B}^{-1}) = \delta^T \widehat{\beta}_j(\widehat{B}^{-1})$, for $j = 1, \ldots, m$.
Note that $T_j$ as defined in \eqref{glsTestStats} is essentially a Wald test and the denominator is a plug-in standard error of $\widehat{\gamma}_j(B^{-1})$.
\end{description}

\subsection{Model selection centering method}
\label{sec::ModelBasedCenter}

In this section we present Algorithm 2,
which aims to remove mean effects that are strong enough to
have an impact on covariance estimation. The strategy here  is to
use a model selection step to identify variables with strong mean
effects.

{\noindent \textbf {Algorithm 2: GLS-Model selection centering} }\\
\noindent\rule{16cm}{0.4pt} \\
Input: $X$,  and $\mathcal{G}(1), \mathcal{G}(2)$: indices of group one and two, respectively. \\
Output: $\widehat{A}^{-1}$, $\widehat{B}^{-1}$, $\widehat{A \otimes
  B}$, $\widehat{\beta}(\widehat{B}^{-1})$,  $\widehat{\gamma}$, $T_j$ for all $j$ \\
\noindent\rule{16cm}{0.4pt}

\begin{description}
\item[1. Run Algorithm 1.] Use the group centering method to obtain initial estimates
$\widehat{\gamma}^{\text{init}}_j = \widehat{\beta}_j^{(1)} -
\widehat{\beta}_j^{(2)}$ for all $j = 1, \ldots, m$.
Let $\hat{B}^{-1}_{\init}$ and $\hat{B}_{\init}$ be as obtained in \eqref{BiHat}.
\item[2. Select genes with large estimated differences in means.]
Let $\groupCenteredGenes = \{j: | \widehat{\gamma}^{\text{init}}_j | >
2\widehat{\tau}_{\text{init}} \}$
denote the set of genes which we consider as having strong mean effects, where
\begin{equation} 
\label{modSelThresh}
	\widehat{\tau}_{\text{init}} \asymp \left(
          \frac{\log^{1/2}m}{\sqrt{m}} + \frac{\lVert \hat{B}_{\init}
            \rVert_1}{n_{\min}} \right)
\sqrt{ \frac{n_{\text{ratio}} |\hat{B}_{\init}^{-1}|_{0, \text{off}} }{n_{\min}}}
+ \sqrt{\log m } \lVert (D^T \hat{B}_{\init}^{-1}D)^{-1} \rVert_2^{1/2},
\end{equation}
with $n_{\min} = n_1 \wedge n_2$,  $n_{\max} =n_1 \vee n_2$, and $n_{\operatorname{ratio}} = n_{\max} / n_{\min}$.
\item[3. Calculate Gram matrices based on model selection centering.]
Global centering can be expressed in terms of the projection matrix
$P_1 = n^{-1} \mathbf{1}_n \mathbf{1}_n^T$.
Compute the centered data matrix
\[
	X_{\operatorname{cen}, j} = \begin{cases}
	X_j - P_2 X_j & \text{if $j \in \groupCenteredGenes$} \\
	X_j - P_1 X_j & \text{if $j \in \groupCenteredGenes^c$},
	\end{cases}
\]
where $X_{\operatorname{cen}, j}$
 denotes the $j$th column of the
centered data matrix $X_{\operatorname{cen}}$.
Compute the sample covariance and correlation matrices with
$X_{\operatorname{cen}}$ following \eqref{SampleCovAB} and \eqref{defGammaAGammaB}.
\item[4.  Estimate covariances and means.]
\begin{enumerate}
\item[(4a)] Obtain the penalized correlation matrices
  $\widehat{B}_{\rho}$ and $\widehat{A}_{\rho}$ using Gemini
  estimators as defined in (\ref{geminiObjectiveFnA}) and
  \eqref{geminiObjectiveFnB} with tuning parameters of the same
  order as those in \eqref{GLassoPenaltyB}.
\item[(4b)] Obtain inverse covariance estimates $\widehat{B}^{-1}$, $\widehat{A}^{-1}$ using \eqref{BiHat}.
\item[(4c)] Calculate the GLS estimator $\widehat{\beta}(\widehat{B}^{-1})$
  as in \eqref{GLSestimator},
as well as the vector of mean differences
$\widehat{\gamma}(\widehat{B}^{-1}) = \delta^T
\widehat{\beta}(\widehat{B}^{-1})$,
for $\delta = (1, -1) \in \mathbb{R}^2$.
\end{enumerate}
\item [5. Obtain test statistics.]  Calculate test statistics as in \eqref{glsTestStats}, now using $\widehat{B}^{-1}$ as estimated in Step 4.
\end{description}
\noindent{\bf Remarks.}
%Although \eqref{modSelThresh} gives the order of the theoretical
%threshold, in reality we do not know the values of the constants that
%depend on $B$. In practice, we suggest two alternatives for
%performing this thresholding.
In the case that $\gamma$ is sparse, we show that this approach can 
perform better than the approach in Section \ref{sec::GroupCenter}, in 
particular when the sample size is small.  
We now consider the expression $\widehat{\tau}_{\text{init}}$ in
\eqref{modSelThresh} as an upper bound on the threshold in the sense that 
it is chosen to tightly control false positives.  
%In practice, we use a plug-in estimator, in which the estimated $\hat{B}_{\init}$ and $\hat{B}_{\init}^{-1}$ from Algorithm 1 are plugged into \eqref{modSelThresh}.
% the theoretical value of $\tau_{\text{init}}$ in step 2 of Algorithm
% 2.  
In Section~\ref{inferenceGamma} we show in simulations 
that with this plug-in estimate $\widehat{\tau}_{\text{init}}$, Algorithm 2
can nearly reach the performance of GLS with the true $B$.  
Since this choice of $\widehat{\tau}_{\text{init}}$ acts as an order on the
threshold we need, the plug-in method can also be applied with a
multiplier between $0$ and $1$.  When we set $\widehat{\tau}_{\text{init}}$ at its lower
bound, namely,
\[
\sqrt{\log m } \lVert (D^T \hat{B}_{\init}^{-1}D)^{-1} \rVert_2^{1/2},
\; \; \text {where} \; \; \hat{B}_{\init}^{-1} \; \; \text{is obtained
  as in Step 3 from Algorithm 1,}
\]
we anticipate many false positives.  In
Figure~\ref{RankCorrplugin_decayExp_m1_2e3_v1}, we show that the
performance of Algorithm 2 is stable in the setting of small $n$ and
sparse $\gamma$ for different values of
$\widehat{\tau}_{\text{init}}$, demonstrating robustness of our methods
to the multiplier; there we observe that the performance can degrade
if the threshold is set to be too small, eventually reaching the
performance of Algorithm 1.

Second, if an upper bound on the number of differentially expressed
genes is known a priori, one can select a set of genes
$\widecheck{J}_0$ to group center such that the cardinality
$|\widecheck{J}_0|$ is understood to be chosen as an upper bound on
$d_0 =|\operatorname{supp}(\gamma)|$ based on prior knowledge.
We select the set $\widecheck{J}_0$ by ranking the components of the estimated
vector of mean differences $\widehat{\gamma}$.  In the data analysis
in Section~\ref{sec::UCData} we adopt this strategy in an iterative manner by
successively halving the number of selected genes, choosing at each
step the genes with largest estimated mean differences from the
previous step. We show in this data example and through simulation
that the proposed method is robust to the choice of $|\widecheck{J}_0|$.

Finally, it is worth noting that these algorithms readily generalize
to settings with more than two groups, in which case we simply group center within each 
group.  This is equivalent to applying the method with a different 
design matrix $D$.  In fact, we can move beyond group-wise mean 
comparisons to a regression analysis with a fixed design matrix $D$, 
which includes the $k$-group mean analysis as a special case. 
%We focus on the two-group setting in the present paper. 

\section{Theoretical results}

\label{sec::mainTheorem}

We first state Theorem~\ref{thm::GLSFixedB}, which provides the rate of convergence of the GLS estimator \eqref{GLSestimator} when we use a fixed approximation of the covariance matrix $B$.
We then provide in
Theorems~\ref{mainTheoremGroupCentering} and~\ref{mainTheoremModSel}
the convergence rates for estimating the group mean matrix $\beta \in \mathbb{R}^{2 \times m}$
for Algorithms 1 and 2 respectively. In Theorem~\ref{mainTheoremGroupCentering}
we state rates of convergence for the Gemini estimators of $B^{-1}$ and $A^{-1}$ when the input sample
covariance matrices use the group centering approach as defined in
Algorithm 1, while in Theorem~\ref{mainTheoremModSel}, 
we state only the rate of convergence for estimating $B^{-1}$,
anticipating that the rate for $A^{-1}$ can be similarly obtained, using the model selection
centering approach as defined in Algorithm 2.

\subsection{GLS under fixed covariance approximation}
\label{sec::GLSFixedBtheorem}

We now state a theorem on the rate of convergence of the GLS estimator
\eqref{GLSestimator}, 
where we use a fixed approximation $\BinvFixed$ to $B^{-1}$,
% to obtain $\widehat{\beta}(\BinvFixed)$, 
where the operator norm of $\DeltaFixed = \BinvFixed - B^{-1}$ is
small in the sense of \eqref{DeltaConditionTwoGroupD}.  
We will specialize Theorem~\ref{thm::GLSFixedB} to the case where
$B^{-1}$ is estimated using the baseline method in~\cite{Zhou14a}
when $X$ follows subgaussian matrix-variate distribution as in
\eqref{eq::matrix-normal-rep-intro}.
We prove Theorem \ref{thm::GLSFixedB} in 
Web Supplement Section~\ref{sec::proofsOfTheorems}. 
\begin{theorem}
\label{thm::GLSFixedB}
Let $Z$ be an $n \times m$ random matrix with independent entries $Z_{ij}$ satisfying
$\E Z_{ij} = 0$, $1 = \E Z_{ij}^2 \le \norm{Z_{ij}}_{\psi_2} \leq K$.
Let $Z_1, \ldots, Z_m \in \mathbb{R}^n$ be the columns of $Z$.  
Suppose the $j$th column of the data matrix satisfies $X_j \sim B^{1/2} Z_j$.  Suppose $\BFixed \in \mathbb{R}^{n \times n}$ is a positive definite symmetric matrix.  Let $\DeltaFixed := \BinvFixed - B^{-1}$.  Suppose
\begin{equation}
 \label{DeltaConditionTwoGroupD}
  \lVert \DeltaFixed \rVert_2 < \frac{1}{\left( n_{\max} / n_{\min}
    \right) \lVert B \rVert_2}, \text{ where } n_{\min} = n_1 \wedge
  n_2 \; \; \text{ and } \; \; n_{\max} =n_1 \vee n_2.
\end{equation}
Then with probability at least $1 - 8 / (m \vee n)^2$, for some
absolute constants $C$, $C'$,
\ben
\label{rateBetaHatFixedB1}
&&  \forall j, \quad \lVert \widehat{\beta}_j(\BinvFixed) - \beta_j^* \rVert_2  \; \leq \;  r_{n, m} :=  s_{n, m} + t_{n, m},
\quad \text{ where }  \\
\label{def:stFixedDelta}
&&  s_{n, m} = C \sqrt{{\log m \lVert B \rVert_2}/{n_{\min}}}
\quad \text{ and} \quad t_{n, m} = C' {\lVert \DeltaFixed
  \rVert_2}/{n_{\min}^{1/2}}; \\
\label{rateGammaHatFixedDelta}
\text{ and } 
&&  \lVert \widehat{\gamma}(\BFixed) - \gamma \rVert_\infty \leq \sqrt{2}
 \left( C \sqrt{\frac{\log m \lVert B \rVert_2}{n_{\min}}} + C' n_{\min}^{-1/2} \lVert \DeltaFixed \rVert_2 \right).
\een
\end{theorem}

% We prove Theorem \ref{thm::GLSFixedB} in Section \ref{sec::ProofThm1}.
% Additional technical lemmas are proved in Section~\ref{sec::proofsOfTheorems}.
\noindent{\textbf{Remarks.}}
 If the operator norm of $B$ is bounded, that is $\lVert B
  \rVert_2 < W$, 
then condition \eqref{DeltaConditionTwoGroupD} is equivalent to $\lVert \DeltaFixed \rVert_2 < 1 / (W n_{\operatorname{ratio}})$.
The term $t_{n, m}$ in \eqref{def:stFixedDelta} reflects the error due to approximating $B^{-1}$ with $B_{n,m}^{-1}$,
whereas $s_{n, m}$ reflects the error in estimating the mean matrix
\eqref{GLSestimator} using GLS with the true $B^{-1}$ for the random
design $X$.  The term $s_{n, m}$ is $O( \sqrt{\log m / n}
)$, whereas $t_{n, m}$ is $O(1/\sqrt{n})$. 
% If $\lVert B \rVert_2$ is
%bounded from below, then $s_{n, m}$ dominates $t_{n, m}$ because of the
%additional $\sqrt{\log m}$ factor. 
The dominating term $s_{n, m}$ in \eqref{def:stFixedDelta} can be replaced by the
tighter bound, namely, $s_{n, m}' = C' \log^{1/2} (m) \sqrt{\delta^T (D^T B^{-1}D)^{-1} \delta}$, with $\delta = (1, -1) \in \mathbb{R}^2$.  This bound correctly drops the factor of $\lVert B \rVert_2$
present in \eqref{def:stFixedDelta} and \eqref{rateGammaHatFixedDelta},
while revealing that variation aligned with the column space of $D$ is
especially important in mean estimation.

Note that the condition \eqref{DeltaConditionTwoGroupD} is not
  stringent, and that the $\hat{B}$ estimates used in Algorithms 1 and
  2 have much lower errors than this.
When $M = 0$ is known, $S_A$ and $S_B$ can be the usual Gram matrices, and the theory in~\cite{Zhou14a} guarantees 
that $t_{n, m}$ as defined in \eqref{def:stFixedDelta} has rate $C_A \sqrt{\log m / m}$, with $C_A = \sqrt{m} \lVert A \rVert_F / \operatorname{tr}(A)$. 
However in our setting, $M$ in general is nonzero.  In Sections 
\ref{sec::GroupCenter} and \ref{sec::ModelBasedCenter} we provide two 
constructions for $S_A$ and $S_B$, which differ in how the data are 
centered.  These constructions have a different bound $t_{n, m}$, as
we will discuss in Theorems \ref{mainTheoremGroupCentering} and \ref{mainTheoremModSel}.  

In Section \ref{sec::Simulations}, we present simulation results that
demonstrate the advantage of the oracle GLS and GLS with estimated
$\hat{B}$ \eqref{GLSestimator} over the sample mean based  (OLS) method
(c.f. \eqref{eq::sampleMean} and \eqref{unpairedTstat}) for mean estimation as well as the related variable selection problem
with respect to $\gamma$.  There, we scrutinize this quantity and its estimation procedure in detail.

\noindent{\textbf{Design effect.}}
The ``design effect'' is the variance of the ``oracle'' GLS estimator \eqref{GLSestimator} of $\gamma_j$ using the true $B$, that is,
\begin{equation} 
\label{designEffect}
\delta^T (D^T B^{-1}D)^{-1} \delta =
\operatorname{Var}(\widehat{\gamma}_j(B^{-1})),  \; \forall j = 1, \ldots, m.
\end{equation}

The design effect reflects the potential improvement of GLS over OLS.  
It appears as a factor above in $s_{n, m}'$, so it contributes to the
rate of mean parameter estimation as characterized in Theorem \ref{thm::GLSFixedB}.  Lower variance in the
GLS estimator of the mean difference contributes to greater power of the
test statistics relative to OLS.  The design effect also appears as a
scale factor in the test statistics for $\hat{\gamma}$
\eqref{glsTestStats}, and therefore it is particularly important that
the design effect is accurately estimated in order for the test
statistics to be properly calibrated.  
In a study focusing on mean differences, it may be desirable to assess the sample size needed to 
detect a given effect size using our methodology.  
Given the design effect, our tests for differential expression are
essentially Z-tests based on the GLS fits, 
followed by some form of multiple comparisons adjustment.
\begin{corollary} \label{theoremInference}
Let $\Omega = (D^T B^{-1} D)^{-1}$, $\widehat{\Omega} = (D^T \widehat{B}^{-1} D)^{-1}$, and $\Delta = \widehat{\Omega} - \Omega$.  Under the conditions of Theorem \ref{thm::GLSFixedB}, the relative error in estimating the design effect is bounded as 
\begin{equation} 
\label{eq::designEffectEstRateCorollaryThm1}
\frac{\abs{ \delta^T \widehat{\Omega} \delta -  \delta^T \Omega \delta}}{ \delta^T \Omega \delta } \leq  2 C' \frac{\kappa(B) \twonorm{B} \twonorm{\Delta} }{n_{\operatorname{ratio}}}, 
\end{equation}
with probability $1 - C / (m \vee n)^d$, for some absolute constants 
$C, C'$. 
\end{corollary}
We prove Corollary \ref{theoremInference} in 
Web Supplement Section~\ref{proofTheoremInference}.
Corollary \ref{theoremInference} implies that given an accurate
estimator of $B^{-1}$, the design effect is accurately estimated and
therefore suggests that traditional techniques can be used to gain an
approximate understanding of the power of our methods.  We show that
$B^{-1}$ can be accurately estimated under conditions in Theorems 3
and 4.  If pilot data are available that are believed to have similar
between-sample correlations to the data planned for collection in a
future study, Corollary~\ref{theoremInference} also justifies using
this pilot data to estimate the design effect. 
%The design effect also has implications for power analysis. 
 If no pilot data are available, it is possible to conduct power analyses based on various
plausible specifications for the $B$ matrix.

%shows that the relative error in estimating the design effect goes to zero.

\subsection{Rates of convergence for Algorithms 1 and 2}

 We state the following assumptions. \\
% on $A$ and $B$.
\noindent{\bf (A1)}
The number of nonzero off-diagonal entries of $A^{-1}$ and $B^{-1}$ satisfy
\bens
	\left| A^{-1} \right|_{0, \text{off}} &= & o(n / \log(m \vee n))
\qquad \qquad \qquad \quad (n, m \rightarrow \infty ) \quad \text{and} \\
	\left| B^{-1} \right|_{0, \text{off}} &= &
o\left(  \frac{m}{\log(m \vee n)} \wedge \frac{n_{\min}^2}{\lVert B \rVert_1^2}  \right) \qquad (n, m \rightarrow \infty ).
\eens
\noindent{\bf (A2)}
 The eigenvalues of $A$ and $B$ are bounded away from 0 and $+\infty$.
We assume that the stable ranks satisfy $r(A), r(B) \geq 4 \log(m \vee
n)$, where $r(A) = \FroNorm{A}^2 / \twonorm{A}^2$.  
\begin{theorem} 
\label{mainTheoremGroupCentering}
Suppose  that (A1) and (A2) hold.  Consider the data as
generated from model \eqref{modelMeanCov} with $\varepsilon = B^{1/2}
Z A^{1/2}$,  where $A \in \mathbb{R}^{m \times m}$ and $B \in
\mathbb{R}^{n \times n}$ are positive definite matrices, and $Z$ is an
$n \times m$ random matrix as defined in
Theorem~\ref{thm::GLSFixedB}.
Let  $C, C', C_1 C_2, C'', C'''$ be some absolute constants.
Let $C_A = \sqrt{m} \lVert A \rVert_F / \operatorname{tr}(A)$
and $C_B = \sqrt{n} \lVert B \rVert_F / \operatorname{tr}(B)$. 
\textbf{(I)} 
Let $\lambda_A$ and $\lambda_B$ denote the penalty parameters for \eqref{geminiObjectiveFnB}
and \eqref{geminiObjectiveFnA} respectively.  Suppose
\ben
 \label{GLassoPenaltyB}
\lambda_A \geq C\left( C_A K \frac{\log^{1/2}(m \vee n)}{\sqrt{m}}
  + \frac{\lVert B \rVert_1}{n_{\min}} \right)
\;\text{ and  } \; \lambda_B \geq C' \left( C_B K \frac{\log^{1/2}(m \vee n)}{\sqrt{n}} +
  \frac{\lVert B \rVert_1}{n_{\min}}\right).
%\label{GLassoPenaltyA}
\een
Then with probability at least $1 - C'' / (m \vee n)^2$,
for $\widehat{A \otimes B}$ as define in 
\eqref{estimatorAKroneckerB}, 
\bens
&&	\lVert \widehat{A \otimes B} - A \otimes B \rVert_2  \leq  \lVert A \rVert_2 \lVert B \rVert_2 \delta,\\
&&	\lVert \widehat{A \otimes B}^{-1} - A^{-1} \otimes B^{-1}
\rVert_2  \leq  \lVert A^{-1} \rVert_2 \lVert B^{-1} \rVert_2  \delta',\\
\text{ where }
\quad &&
\delta, \delta' = O\left(\lambda_A \sqrt{|B^{-1}|_{0, \text{off}} \vee 1} + \lambda_B \sqrt{|A^{-1}|_{0, \text{off}} \vee 1} \right).
\eens
Furthermore, 
 with probability at least $1 - C'''/(m \vee n)^2$,
\ben \
&&	\lVert \widehat{A \otimes B}
 - A \otimes B \rVert_F  \leq  \lVert A \rVert_F \lVert B \rVert_F
 \eta, \\
%&&	\lVert \widehat{A \otimes B}^{-1} - A^{-1} \otimes B^{-1}
%\rVert_F \leq  \lVert A^{-1} \rVert_F \lVert B^{-1} \rVert_F \eta',\\
\text{ where } &&
\label{eq::eta}
\eta = O\left(\lambda_A \sqrt{|B^{-1}|_{0, \text{off}} \vee n} / \sqrt{n} +
\lambda_B \sqrt{|A^{-1}|_{0, \text{off}} \vee m} / \sqrt{m} \right).
\een
The same conclusions hold for the inverse estimate, with $\eta$ being
bounded in the same order as in
\eqref{eq::eta}.
\textbf{(II)}
Let $\widehat{\beta}$ be defined as in (\ref{GLSestimator}) with
$\widehat{B}^{-1}$
being defined as in \eqref{BiHat} and $D$ as in (\ref{meanMatrixTwoGroups}).
Then, with probability at least $1 - C/m^d$ the following holds for all $j$,
\begin{equation}
\label{mainThmBoundTailCutpointGroup}
	\lVert \widehat{\beta}_j(\widehat{B}^{-1}) - \beta_j^*
        \rVert_2 \leq C_1 \lambda_A \sqrt{ \frac{n_{\operatorname{ratio}}  \left(|B^{-1}|_{0, \text{off}} \vee 1\right)}{n_{\min}}} +
        C_2 \sqrt{\log m} \lVert (D^T B^{-1} D)^{-1} \rVert_2^{1/2}.
\end{equation}
\end{theorem}
We prove Theorem \ref{mainTheoremGroupCentering} part I in Web Supplement Section \ref{sec::ProofMainThmPartI}; this relies on rates of convergence of $\widehat{B}^{-1}$ and $\widehat{A}^{-1}$ in the operator and  the Frobenius norm,
 which are established in Lemma \ref{boundCovOpFro}.
We prove part II in Web Supplement Section \ref{sec::ProofMainThmPartII}.

\noindent{\bf Remarks.}
We find that the additional complexity of estimating the mean matrix leads to an additional
additive term of order $1 / n$ appearing in the convergence rates for covariance estimation for $B$ and $A$.
In part I of Theorem~\ref{mainTheoremGroupCentering}, $\lambda_A$ is
decomposed into two terms, one term reflecting the variance of $S_B$, and one term reflecting the bias due to group centering.
The variance term goes to zero as $m$ increases, and the bias term goes to zero as $n$ increases.
To analyze the error in the GLS estimator based on $\widehat{B}^{-1}$,
we decompose $\lVert \widehat{\beta}_j(\widehat{B}^{-1}) - \beta_j^*
\rVert_2$ as \begin{equation*}
	 \lVert \widehat{\beta}_j(\widehat{B}^{-1}) - \beta_j^* \rVert_2 \leq\lVert \widehat{\beta}_j(\widehat{B}^{-1}) - \widehat{\beta}_j(B^{-1}) \rVert_2  + \lVert \widehat{\beta}_j(B^{-1})  - \beta_j^* \rVert_2,
\end{equation*}
where the first term is the error due to not knowing $B^{-1}$, and the second term is the error due to not knowing $\beta_j^*$.  The rate of convergence given in (\ref{mainThmBoundTailCutpointGroup}) reflects this decomposition.
For Algorithm 2, we have analogous rates of convergence for both mean and covariance estimation.
Simulations suggest that the constants in the rates for Algorithm 2
are smaller than those in \eqref{mainThmBoundTailCutpointGroup}.

We state the following assumptions for Theorem \ref{mainTheoremModSel} to hold on Algorithm 2.

\textbf{(A2')} Suppose (A2) holds, and $n = \Omega\left( (\log m) \Parens{
  {\twonorm{A} \twonorm{B} b_{\max}}/{C_A^2} } \right)$.

\textbf{(A3)}
Let $\text{supp}(\gamma) = \{ j : \gamma_j \neq 0 \}$.  Let $s = \abs{\text{supp}(\gamma)}$ denote the sparsity of $\gamma$.
Assume that $s = O\Parens{\frac{C_A}{\twonorm{B} } n \sqrt{\frac{m}{\log m}}}$.

\textbf{Remarks.}  Condition (A2') is mild, because the
condition on the stable rank of $B$ already implies that 
$n \geq \log m$.  

%Next, we state the main theorem bounding the error in
\begin{theorem} 
\label{mainTheoremModSel}
Suppose  that (A1), (A2'), and (A3) hold.  Consider the data as generated from model \eqref{meanMatrixTwoGroups}
with $\varepsilon = B^{1/2} Z A^{1/2}$,  where $A \in \mathbb{R}^{m
  \times m}$ and $B \in \mathbb{R}^{n \times n}$ are positive definite
matrices, and $Z$ is an $n \times m$ random matrix as defined in
Theorem \ref{mainTheoremGroupCentering}.
%satisfying $\E Z_{ij} = 0$, $1 = \E Z_{ij}^2 \le \norm{Z_{ij}}_{\psi_2} \leq K$.
%with independent entries $Z_{ij}$ satisfying
%$\E Z_{ij} = 0$, $1 = \E Z_{ij}^2 \le \norm{Z_{ij}}_{\psi_2} \leq K$.
 Let $\lambda_A$ denote the penalty parameter for estimating $B$.
 Suppose $\lambda_A$ is as defined in \eqref{GLassoPenaltyB}.  Let
 \begin{equation} 
\label{threshTheorem3}
	\tau_{\operatorname{init}} \asymp \sqrt{\log m} \lVert (D^T
        B^{-1}D)^{-1} \rVert_2^{1/2}.
\end{equation}
Then with probability at least $1 - C'' / (m \vee n)^2$, for  output
of Algorithm 2,
\begin{align} \label{mainThmErrorCov}
\twonorm{ \Trace{A} \Parens{\widehat{W}_2 \widehat{B}_{\rho} \widehat{W}_2}^{-1} - B^{-1} } \leq \frac{C' \lambda_A \sqrt{\abs{B^{-1}}_{0, \operatorname{off}} \vee 1 } }{b_{\min} \varphi_{\min}^2(\rho(B))}, \quad \text{and}
\end{align}
% Let $\widehat{\beta}$ be defined as in (\ref{GLSestimator}) with $\widehat{B}^{-1}$ being defined as in \eqref{BiHat} and $D$ as in (\ref{meanMatrixTwoGroups}).
\begin{equation}
\label{mainThmBoundTailCutpoint}
	\lVert \widehat{\beta}_j(\widehat{B}^{-1}) - \beta_j^*
        \rVert_2 \leq C_2 \sqrt{\log m} \lVert (D^T B^{-1} D)^{-1} \rVert_2^{1/2},
\end{equation}
for all $j$, for absolute constants $C$, $C_2$, $C'$, and $C''$.
\end{theorem}

We prove Theorem \ref{mainTheoremModSel} in Web Supplement Section~\ref{proofMainThmAlgTwo}. In
Web Supplement Section~\ref{MainTheoremFixedGenes} we also show a standalone result, namely Theorem~\ref{mainTheoremFixedGenes}, for the case of fixed sets of
group and globally centered genes.  This result shows how the
algorithm used in the preliminary step to choose which genes to group
center can be decoupled from the rest of the estimation procedure in
Algorithm 2, so long as certain conditions hold.  The proof of Theorem \ref{mainTheoremModSel} indeed validates that such conditions hold for the output of Algorithm 1.  It is worth
noting that a similar rate of convergence for estimating $A$ could also
be derived, but we focus on $B$ in our methodology and
applications, and therefore leave this as an exercise for interested readers.

% \textbf{Remarks:} \label{sec::designEffectRate}
We specialize Corollary \ref{theoremInference} to the case where $B^{-1}$ is estimated using Algorithm 2.  
\begin{corollary} \label{corDesignEffectAlg2}
Under the conditions of Theorem \ref{mainTheoremModSel}, we have with
probability $1-C/m^2$
\begin{equation} 
\label{eq::designEffectEstRate}
\frac{\abs{ \delta^T \widehat{\Omega} \delta -  \delta^T \Omega \delta}}{ \delta^T \Omega \delta } \leq  2 C' \frac{n_{\operatorname{ratio}}}{ \lambda_{\min}(B) } \kappa(B) \lambda_A \sqrt{\abs{B^{-1}}_{0, \operatorname{off}} \vee 1},
\end{equation}
for some absolute constants $C$ and $C'$.
\end{corollary}

\textbf{Remarks.}  \label{sec::designEffectRate}
The right-hand-side of \eqref{eq::designEffectEstRate} goes to zero because of the assumptions (A1), (A2'), and (A3), which ensure that the factor $\lambda_A \sqrt{\abs{B^{-1}}_{0, \operatorname{off}} \vee 1}$ goes to zero.  We conduct simulations to assess the accuracy of estimating the design effect in Section~\ref{inferenceGamma}.
% If the true design effect were known, then we could apply traditional GLS power analysis to the test statistics \eqref{glsTestStats}. This theorem implies that we can accurately estimate this design effect, and therefore can use the same techniques to gain an approximate understanding of the power of our methods.

%!TEX root = submit.arxiv.tex

\section{Simulations}
\label{sec::Simulations}
We present simulations to compare Algorithms 1 and 2 to both sample mean based
analysis and oracle algorithms that use knowledge of the true correlation
structures $A$ and $B$.
We show these results for a variety of population structures and sample sizes.
We construct covariance matrices for $A$ and $B$ from one of:
\begin{itemize}
\item AR1$(\rho)$ model.  The covariance matrix is of the form $B =\{\rho^{|i-j|}\}_{i,j}$, and the graph corresponding to $B^{-1}$ is a chain.
\item Star-Block model.
The  covariance matrix is block-diagonal with equal-sized
blocks whose inverses correspond to star structured graphs, where
$B_{ii} = 1$, for all $i$.  In each subgraph, a central hub node connects to all other nodes in the subgraph,
with no additional edges.  The covariance matrix for each block $S$ in $B$ is generated as in \cite{RWRY08}:
$S_{ ij} = \rho = 0.5$ if $(i,j) \in E$ and $S_{ij} = \rho^2$
otherwise.
\item Erd\H{o}s-R\'{e}nyi model.  We use the random concentration matrix model in
\cite{ZLW08}. The graph is generated according to a type of Erd\H{o}s-R\'{e}nyi random graph.  
Initially we set $B^{-1} = 0.25 I_{n \times n}$.
Then, we randomly select $d$ edges and update $B^{-1}$ as follows:
for each new edge $(i, j)$, a weight $w >0$ is chosen uniformly at random from $[w_{\min}, w_{\max}]$
where $w_{\min} = 0.6$ and $w_{\max} = 0.8$; we subtract $w$ from $B^{-1}_{ij}$ and $B^{-1}_{ji}$,
and increase $B^{-1}_{ii}$ and $B^{-1}_{jj}$ by $w$. This keeps $B^{-1}$ positive definite.
We then rescale so that $B^{-1}$ is an inverse correlation matrix.
\end{itemize}

\subsection{Accuracy of $\widehat{\gamma}$ and its implication for variable ranking}

Table~\ref{CovMeanMetrics} displays metrics that reflect how the choice of different population structures $B$ can affect the difficulty of the mean and covariance estimation problems.
Column 2  is a measure discussed by \citet{efron2007correlation}.  Column 3 appears directly in the theoretical analysis,
reflecting the entry-wise error in the sample correlation $\widehat{\Gamma}(B)$.
Columns 4 analogously reflects the entry-wise error for the Flip-Flop procedure in \citet{Zhou14a},
and is included here for completeness.  Column 5 displays the value of $\sqrt{\delta^T (D^T B^{-1} D)^{-1} \delta} $, where $\delta = (1, -1) \in \mathbb{R}^2$, which represents the standard deviation of the difference in means estimated using GLS with the true $B^{-1}$.  Column 6 displays what we call the standard deviation ratio, namely
\begin{equation} \label{sdRatio}
	\sqrt{\frac{u^TBu}{\delta^T (D^T B^{-1} D)^{-1} \delta}},
\end{equation}
where $u = ( \underbrace{1/n_1, \ldots, 1/n_1}_{n_1}, \underbrace{-1/n_2, \ldots, -1/n_2}_{n_2}) \in \mathbb{R}^n$
and $\delta = (1, -1) \in \mathbb{R}^2$, which reflects the potential efficiency gain for
GLS over sample mean based method~\eqref{eq::sampleMean}
for estimating $\gamma$.  Note that the standard deviation ratio depends on the relationship between the covariance matrix $B$ and the design matrix $D$.  In Table~\ref{CovMeanMetrics}, the first $n/2$ individuals are in group one, and the following $n/2$ are in group two.  The values in Column 6 show that substantial improvement is possible in mean estimation.
For an AR1 covariance matrix, the standard deviation ratio increases as the AR1 parameter increases;
as the correlations get stronger, the potential improvement in mean estimation due to GLS grows.
For the Star Block model with fixed block size, the standard deviation ratio is stable as $n$ increases.

\begin{table}[tb]
\centering
\begin{tabular}{rlrrrrr}
  \hline
 & $B$ & $\rho_B^2$ & $\lVert B \rVert_F / \text{tr}(B)$ & $\left| \rho(B)^{-1} \right |_{1, \text{off}}$ & sd GLS & sd ratio \\
  \hline
 \multicolumn{7}{l}{$n = 80$} \\ \hline
1 & AR1(0.2) & 0.00 & 0.12 & 32.92 & 0.27 & 1.00 \\
  2 & AR1(0.4) & 0.00 & 0.13 & 75.24 & 0.33 & 1.02 \\
  3 & AR1(0.6) & 0.01 & 0.16 & 148.12 & 0.40 & 1.07 \\
  4 & AR1(0.8) & 0.04 & 0.24 & 351.11 & 0.46 & 1.32 \\
  5 & StarBlock(4, 20) & 0.02 & 0.18 & 101.33 & 0.35 & 1.51 \\
  6 & ER(0.6, 0.8) & 0.01 & 0.14 & 92.75 & 0.17 & 1.21 \\
 \hline \multicolumn{7}{l}{$n = 40$} \\ \hline
 1 & AR1(0.2) & 0.00 & 0.16 & 16.25 & 0.38 & 1.01 \\
  2 & AR1(0.4) & 0.01 & 0.19 & 37.14 & 0.45 & 1.03 \\
  3 & AR1(0.6) & 0.03 & 0.23 & 73.12 & 0.53 & 1.12 \\
  4 & AR1(0.8) & 0.08 & 0.33 & 173.33 & 0.53 & 1.47 \\
  5 & StarBlock(2, 20) & 0.04 & 0.25 & 50.67 & 0.50 & 1.51 \\
  6 & ER(0.6, 0.8) & 0.02 & 0.21 & 47.24 & 0.25 & 1.23 \\
   \hline
\end{tabular}
\caption{Assessment of the difficulty of estimating $B^{-1}$ and the potential gain from GLS. The total correlation $\rho_B$ is the average squared off-diagonal value of the correlation matrix $\rho(B)$.  The fourth column is the design effect as defined in \eqref{designEffect}. The last column (sd ratio) presents the ratio of the standard deviation of the difference in sample means in~\eqref{eq::sampleMean} to the standard deviation of the GLS estimator of the difference in means.
The first three columns of the table reflect the difficulty of estimating $B$, whereas the last two columns reflect the potential improvement of GLS over the sample mean based method~\eqref{eq::sampleMean}. In the notation StarBlock$(a, b)$,  $a$ refers to the number of blocks, and $b$ refers to the block size.   } \label{CovMeanMetrics}
\end{table}

\begin{figure}[h!] \centering
\includegraphics[width=0.9 \textwidth]{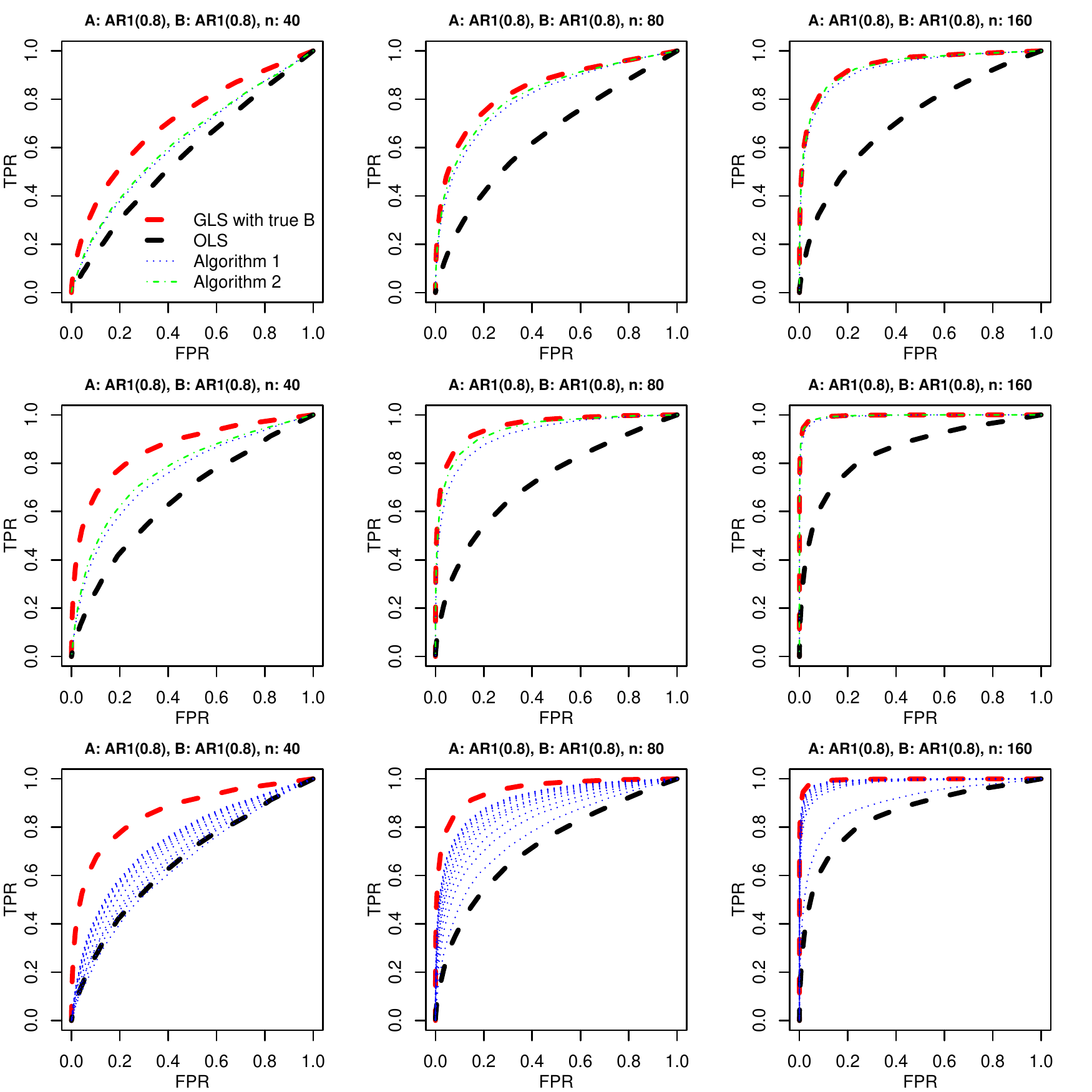}
\caption{ROC curves.  For each plot, the horizontal axis is false positive rate (FPR) and the vertical axis is true positive rate (TPR),
as we vary a threshold for classifying variables as null or non-null.
The covariance matrices $A$ and $B$ are both AR1 with parameter $0.8$,
with $m = 2000$ and $n = 40$, $80$, and $160$ in column one, two, and three, respectively.
Ten variables in $\gamma$ have nonzero entries.
On each trial, the group labels are randomly assigned, with equal sample sizes.
The marginal variance of each entry of the data matrix is equal to one.
For the first row of plots, the magnitude of each nonzero entry of $\gamma$ is $0.2$,
and for the second and third rows of plots, the magnitude of each nonzero entry of $\gamma$ is $0.3$.
In the first two rows we display ROC curves for Algorithms 1 and 2 with penalty parameters chosen to maximize area under the curve.
The third row displays an ROC curves for Algorithm 1, sweeping out penalty parameters.} \label{fig::ROC}
\end{figure}

In Figure \ref{fig::ROC},
we use ROC curves to illustrate the sensitivity and specificity for variable selection in the sense of how well we can identify the support for $\{ i: \gamma_i \neq 0\}$
when we threshold $\hat{\gamma}_i$ at various values. To evaluate and compare different methods, we let $\widehat{\gamma}$ be the output of Algorithm 1, Algorithm 2, the oracle GLS,
and the sample mean based method~\eqref{eq::sampleMean}.
% In our first experiment, we set the population covariance matrices $A$ and $B$ as AR1$(0.8)$, with $m = 2000$, and $n = 40$, $80$,
% and $160$.
These correspond to the four curves on each plot of the top two rows of plots.
% We compare the results for Algorithms 1 and 2 to the results of the oracle GLS and the
% sample mean based method~\eqref{eq::sampleMean}.
We find that Algorithm 1 and Algorithm 2 perform better than the sample mean based method~\eqref{eq::sampleMean}, and in some cases perform comparably to the oracle GLS.  Plots in the third row of Figure \ref{fig::ROC}
illustrate the sensitivity of Algorithm 1 to the choice of the graphical lasso (GLasso) penalty parameter~\eqref{GLassoPenaltyB}; the simulations are run using the \texttt{glasso} R package \citep{FHT07} to estimate $B$ via \eqref{geminiObjectiveFnB}.  The performance can degenerate to that of the sample mean based method~\eqref{eq::sampleMean}, if the penalty is too high.

\begin{figure}[h] \centering 
\includegraphics[width=.95\textwidth]{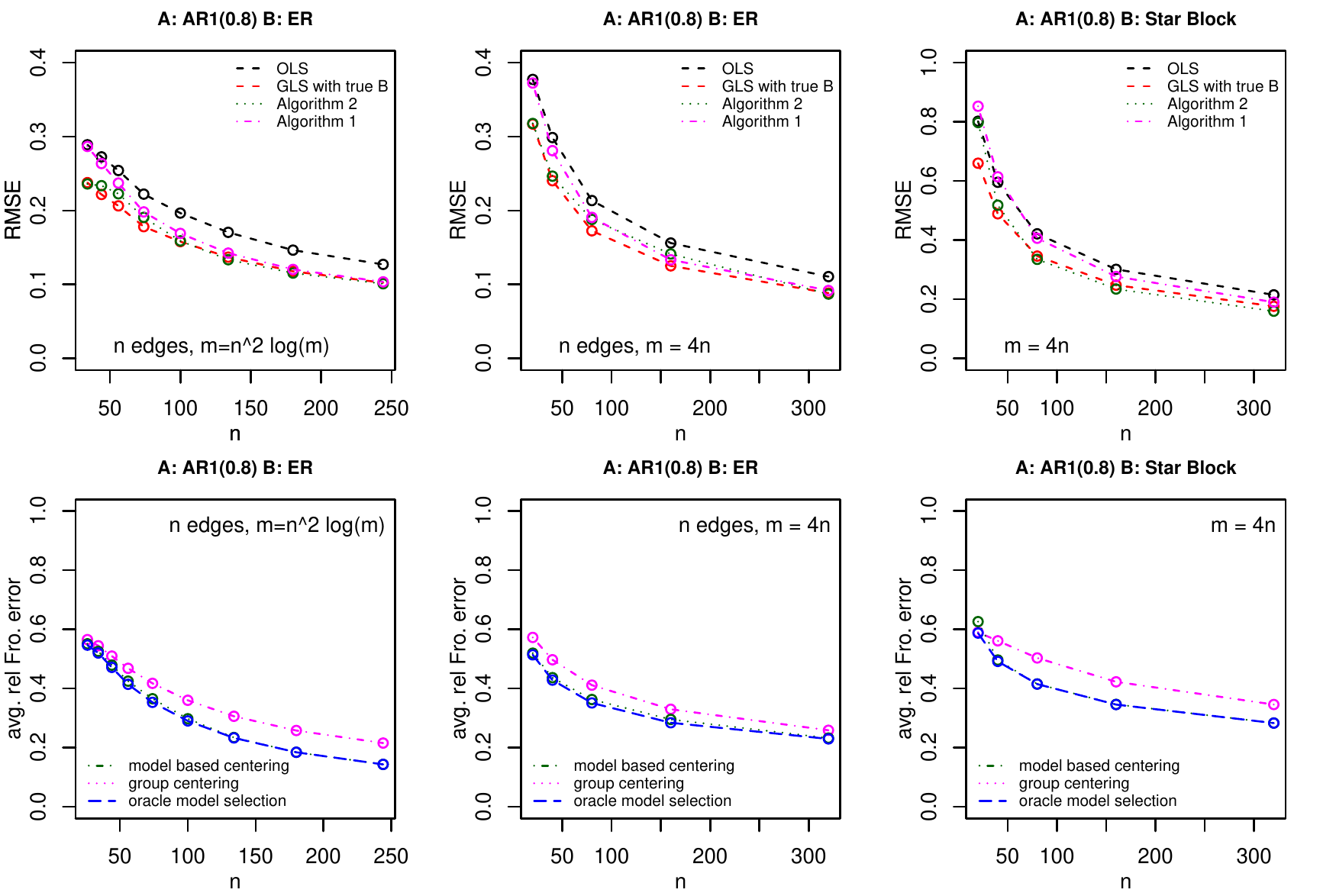}
\caption{Performance of centering methods as $n$ and $m$ are varied, with $n$ shown on the horizontal axis.  In the first column of plots, the number of edges is proportional to $\sqrt{m / \log(m)}$.  In the second and third columns of plots, the number of edges is proportional to $m$.  In the first two columns of plots, $B^{-1}$ is an Erd\H{o}s-R\'{e}nyi inverse covariance matrix.  In the third column, $B^{-1}$ is star block with blocks of size $10$.  The first row of plots shows RMSE for estimating $\gamma$, whereas the second row shows average relative Frobenius error in estimating $B^{-1}$. All panels are based on 250 simulation replications.}
\label{fig::sixPlotPanel}
\end{figure}

In the top row of Figure~\ref{fig::sixPlotPanel} we plot the root mean squared error (RMSE) when estimating the mean differences $\gamma$ for Algorithm 1, Algorithm 2, OLS (i.e. sample means) and the oracle GLS estimate. The population structures for $B$ are Erd\H{o}s-R\'{e}nyi and Star Block. Both Algorithms 1 and 2 consistently outperform the sample mean based method~\eqref{eq::sampleMean} for mean estimation, and Algorithm 2 even achieves comparable performance to the oracle GLS in some settings.
The bottom row displays the relative Frobenius error for estimating $B^{-1}$. Algorithm 2 outperforms Algorithm 1 in terms of covariance estimation and is comparable to oracle model selection, which only centers the columns with a true mean difference.

\begin{figure}[tb] \centering 
\includegraphics[width=0.9\textwidth]{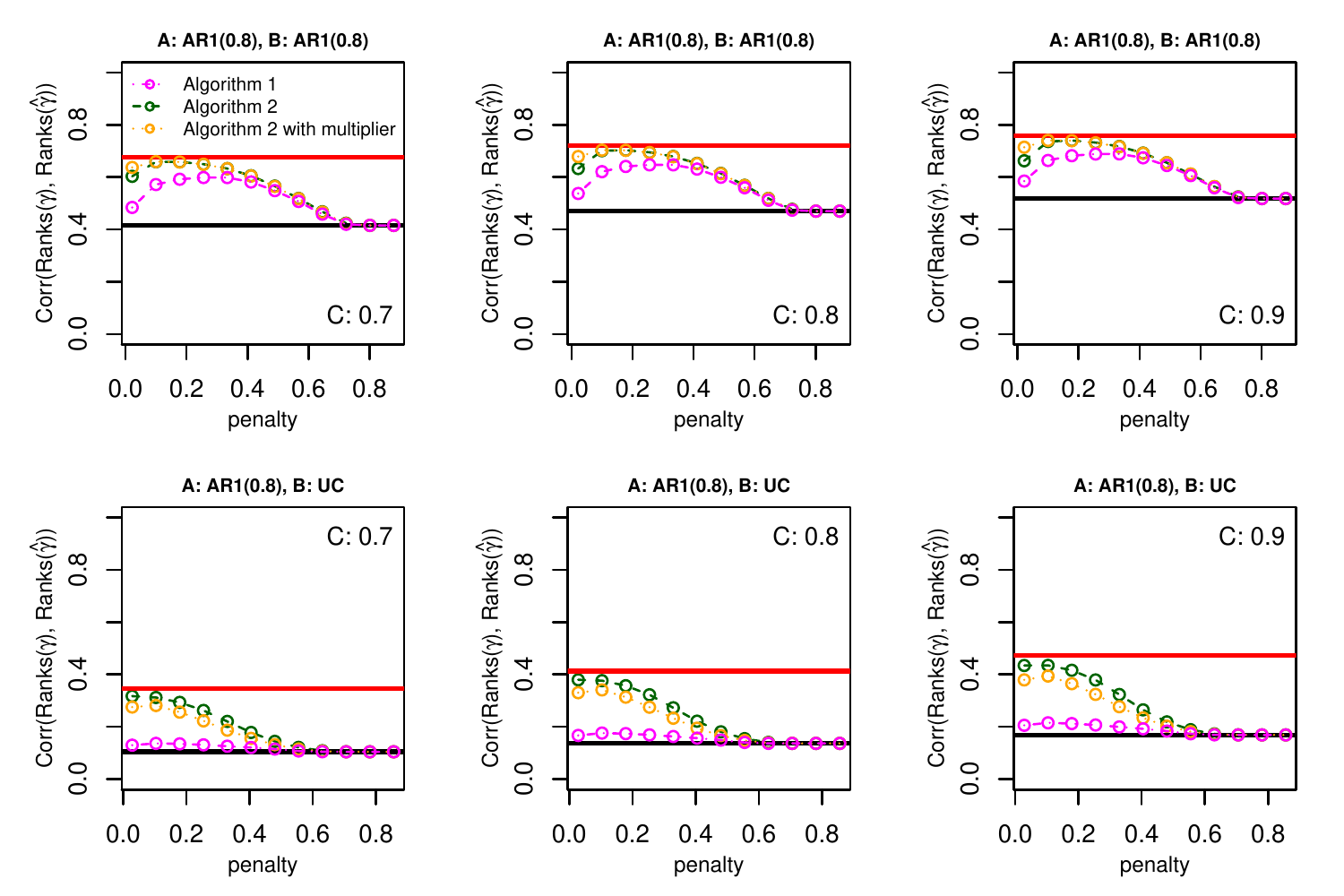}
\caption{This figure displays the correlation between the rankings of the components of $\gamma$ and $\widehat{\gamma}$, sorted by magnitude, denoted Corr(Ranks$(\gamma)$, Ranks$(\widehat{\gamma}))$ in the axis label.  The vector of mean differences is chosen as $\gamma_j = C \exp(- (3/2000) j)$, for $j = 1, \ldots, 2000$.  We also present the Algorithm 2 results with a multiplier on the threshold as described in Section \ref{sec::ModelBasedCenter}. In the top row, the true $B$ is AR1(0.8), with $n = 40$ and $m = 2000$.  In the bottom row, the true $B$ is chosen as an estimate from the UC data, with $n = 20$ and $m = 2000$.  For the top row, the group labels are randomly assigned;  for the bottom row, the first ten rows of the data are in group one, and the other ten are in group two.  The figure is averaged over $200$ replications.  The top and bottom horizontal lines represent GLS with true $B$ and OLS, respectively.  The vertical axis displays the correlation of ranks between $\widehat{\gamma}$ and $\gamma$, and the horizontal axis displays the GLasso penalty parameter.}
\label{RankCorrplugin_decayExp_m1_2e3_v1}
\end{figure}
In Figure \ref{RankCorrplugin_decayExp_m1_2e3_v1}, we illustrate that Algorithm 2 can perform well using a plug-in estimator $\widehat{\tau}_{\init}$ as in \eqref{modSelThresh}.  We compare the methods when the true mean structure is a decaying exponential; we display the correlation of the ranks of the entries of $\gamma$ to the ranks of the estimates of $\gamma$.  Algorithm 2 with a plugin estimator $\widehat{\tau}_{\init}$ can nearly reach the performance of GLS with the true $B$.  Furthermore, the plug-in version of Algorithm 2 also consistently outperforms Algorithm 1.  We also assess sensitivity to the choice of threshold: the curve labeled ``Algorithm 2'' uses the plug-in estimate $\widehat{\tau}_{\init}$, whereas ``Algorithm 2 with threshold multiplier'' uses a plug-in estimate of the lower bound given in \eqref{threshTheorem3} in Theorem \ref{mainTheoremModSel}.
% \begin{equation*} 	\ModSelThreshold \asymp \sqrt{\log(m)} \lVert (D^T B^{-1}D)^{-1} \rVert_2^{1/2}.
%\end{equation*}
These two-plug in estimators exhibit similar performance, showing robustness of Algorithm 2 to the choice of the threshold parameter.  In real data analysis, we validate this further.  For the top row (AR1), the ratio of thresholds \eqref{threshTheorem3}  to \eqref{modSelThresh} is $0.75$, and for the bottom row (UC), the ratio is $0.17$.

In Web Supplement Section~\ref{secComparison}, we perform additional simulations to 
compare Algorithm 2 to two similar methods using ROC curves, namely,
the  sphering method of \citet{allen2012inference}, which uses a matrix-variate model similar to ours, and the confounder adjustment method of \citet{wang2015confounder}, which uses a latent factor model.  Our simulations show that Algorithm 2 consistently outperforms these competing methods in a variety of simulation settings using matrix-variate data.

\subsection{Inference for the mean difference $\widehat{\gamma}$} \label{inferenceGamma}

Two basic approaches to conducting inference for mean differences are paired and unpaired t statistics.  The unpaired t statistic is defined as follows.  Let $X = (X_{ij})$.  Then the $j$th unpaired t statistic is
\begin{eqnarray} \label{unpairedTstat}
	T_j & = & \left( \widetilde{\beta}_j^{(1)} - \widetilde{\beta}_j^{(2)} \right) \widehat{\sigma}_j^{-1} (n_1^{-1} + n_2^{-1})^{-1/2},
\text{ where} \; \; \\
\nonumber
	\widehat{\sigma}_j^2 & = & (n_1 + n_2 - 2)^{-1} \sum_{k = 1}^{2} \sum_{i \in \mathcal{G}_k} \left(X_{ij} - \widetilde{\beta}_j^{(k)} \right)^2, 
\end{eqnarray}
where $\widetilde{\beta}_j^{(k)}$, $k = 1, 2$, and $j = 1, \ldots, m$, denotes the sample mean of group $k$ and variable $j$ as
defined in \eqref{eq::sampleMean}, and $\mathcal{G}_k$ is the set of indices corresponding to group $k$.  When there is a natural basis for pairing the observations, and paired units are anticipated to be positively correlated, we can calculate paired t statistics.  For the paired t statistic, suppose observations $i$ and $i' = i + n/2$ are paired, for $i \in \{1, \ldots, n/2\}$.  Note that samples can always be permuted so as to be paired in this way.  Define the paired differences $d_{ij} = X_{ij} - X_{i'j}$, for $i \in \{1, \ldots, n/2\}$.  Then the paired t statistic is $\overline{d}_j (n/2 - 1)^{1/2} / \left( \sum_{i = 1}^{n/2} (d_{ij} - \overline{d}_j)^2 \right)^{1/2}$, where $\overline{d}_j = (n/2)^{-1} \sum_{i = 1}^{n/2} d_{ij}$.

%We define paired and unpaired t statistics in the appendix, equation \eqref{unpairedTstat}.

\begin{figure}[tb] \centering 
\includegraphics[width=0.95\textwidth]{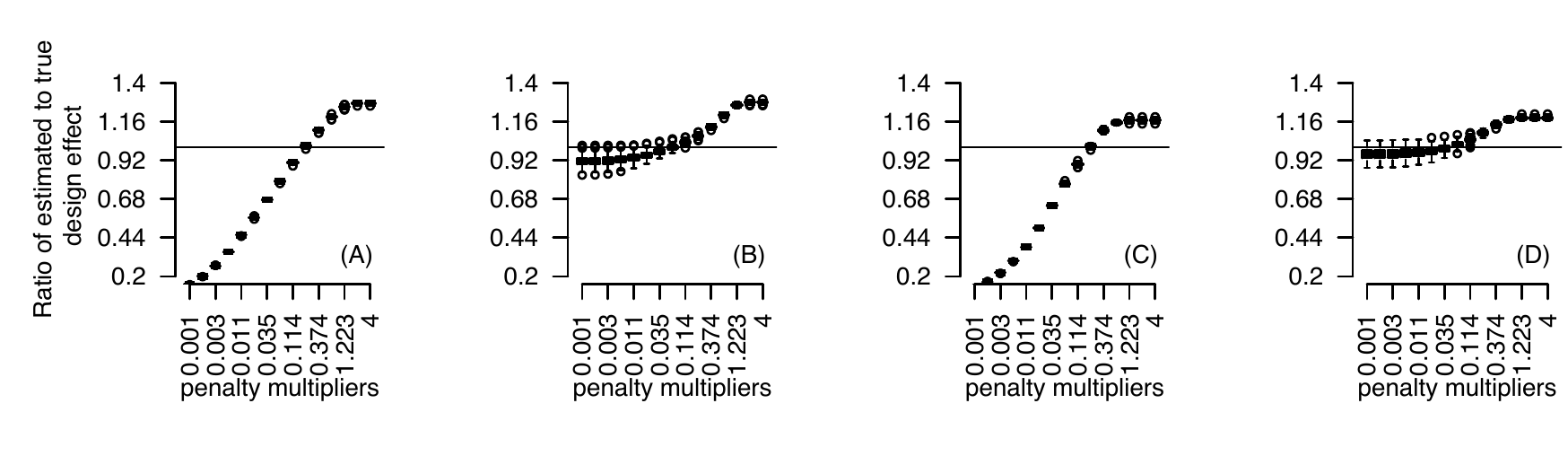}
\caption{Ratio of estimated design effect to true design effect when $B^{-1}$ is Erd\H{o}s-R\'{e}nyi, and $A$ is AR1$(0.8)$.  Figures (A) and (B) correspond to sample size $n = 80$; (C) and (D) correspond to $n = 40$.  Figures (A) and (C) correspond to Algorithm 1;  Figures (B) and (D) correspond to Algorithm 2, with ten columns group centered.  These results are based on dimension parameter $m=2000$ and $250$ simulation replications.}
\label{fig::boxplots}
\end{figure}
% previous version: boxplots4_designEffect_corrER68_2016_9_2 

Figure \ref{fig::boxplots}
considers estimation of the ``design effect''
$\delta^T (D^T B^{-1} D)^{-1} \delta$, as previously defined in \eqref{designEffect},
with $\delta = (1, -1)^T$.  The importance of this object is discussed in Sections~\ref{sec::GLSFixedBtheorem} and \ref{sec::designEffectRate}.  The design effect is estimated via $\delta^T (D^T \widehat{B}^{-1} D)^{-1} \delta$, with $\widehat{B}^{-1}$ from Algorithm 1 or 2.  The GLasso penalty parameters are chosen as
\begin{align}
\lambda_A &= f_A \left( C_A K \frac{\log^{1/2}(m \vee n)}{\sqrt{m}} + \frac{\lVert B \rVert_1}{n_{\min}} \right) \label{MultiplierGLassoPenaltyB}
\end{align}
where we sweep over the factor $f_A$, referred to as the penalty multiplier.
Figure \ref{fig::boxplots} displays boxplots of the ratio
$\delta^T (D^T \widehat{B}^{-1} D)^{-1} \delta / \delta^T (D^T B^{-1} D)^{-1} \delta$
over $250$ replications for each setting of the penalty multiplier $f_A$.  In Figure \ref{fig::boxplots},
$B^{-1}$ follows the Erd\H{o}s-R\'{e}nyi model, and $A$ is AR1$(0.8)$, with $m = 2000$, and $n = 40$ and $80$.  Figure \ref{fig::boxplots}
shows that Algorithm 2 (plots B and D) estimates the design effect to high accuracy and is quite
insensitive to the penalty multiplier as long as it is less than $1$, as predicted by the theoretical analysis.
Algorithm 1 also estimates the design effect with high accuracy, but with somewhat greater sensitivity to the tuning parameter.
The best penalty parameter for Algorithm 1 is around $0.1$, whereas reasonable penalty parameters for Algorithm 2 are
in the range $0.01$ to $0.1$.  This is consistent with smaller entrywise error in the sample covariance for model selection centering than for group centering.

\begin{figure}[h!] \centering 
\includegraphics[width=0.9 \textwidth]{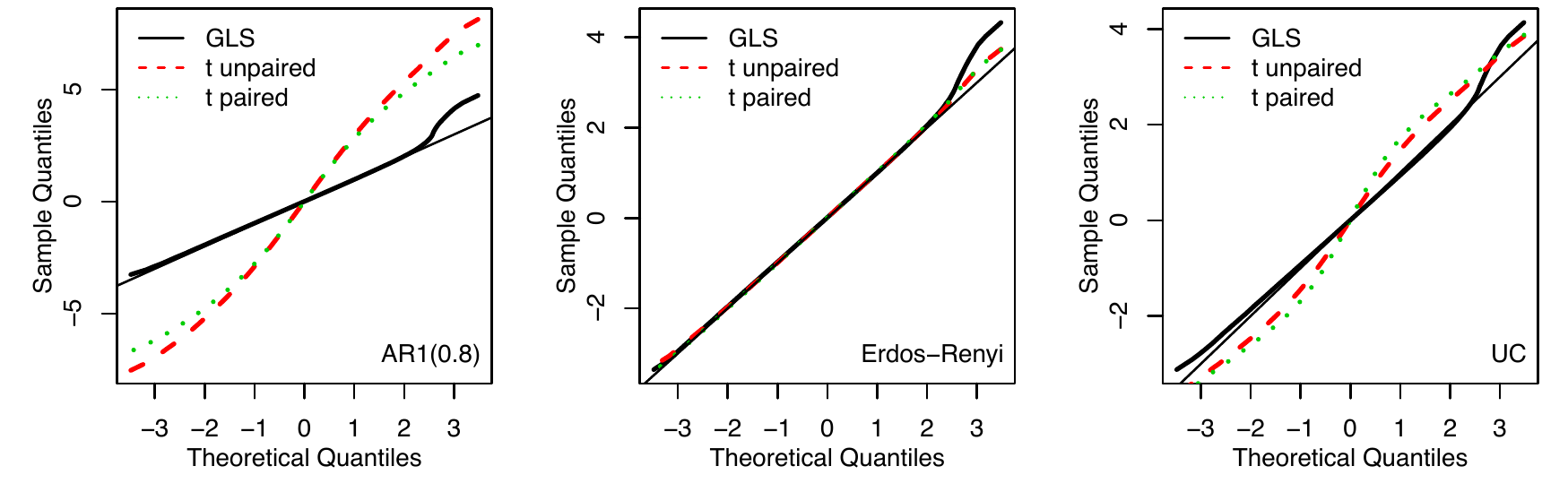}
\caption{Quantile plots of test statistics.  Ten genes have nonzero mean differences equal to $2$, $0.8$, and $1$ in the three plots, respectively.  In each plot $A$ is AR1$(0.8)$.  Covariance structures for $B$ are as indicated.  In the third plot, the true $B$ is set to $\widehat{B}$ for the ulcerative colitis data, described in Section \ref{sec::UCData}.  For the first two plots there are $n = 40$ samples and $m = 2000$ variables.  For the third plot there are $n=20$ samples and $m = 2000$ variables.  Each plot has $250$ simulation replications.}
\label{fig::QQAR}
\end{figure}

We next compare the results from Algorithm 2 to results obtained using paired and unpaired t statistics.
Figure \ref{fig::QQAR} illustrates the calibration and power of plug-in Z-scores, $\widehat{\gamma}_j / \widehat{\rm SE}(\hat{\gamma}_j)$ derived from Algorithm 2 for three population settings.  The standard error is calculated as $\sqrt{\delta^T (D^T \widehat{B}^{-1} D)^{-1} \delta}$, with $\delta = (1, -1$).  In the first and second plots, the data was simulated from AR1$(0.8)$ and Erd\H{o}s-R\'{e}nyi, respectively.  In the third plot, the data was simulated from $\widehat{B}$ for ulcerative colitis data described in Section \ref{sec::UCData}.  To obtain $\widehat{B}$, we apply Algorithm 2 to the ulcerative colitis data, using a Glasso penalty of $\lambda \approx 0.5 [(\log(m) / m) + 3/n]$ in step 1, followed by group centering the top ten genes in step 2, and using a Glasso penalty of $\lambda \approx 0.1 [(\log(m) / m) + 3/n]$ in step 4.  In all cases $A$ is AR1(0.8).  In each case, we introduce 10 variables with different population means in the two groups, by setting $\gamma=0.8$ for those variables, with the remaining $\gamma$ values equal to zero.
The ideal Q-Q plot would follow the diagonal except at the upper end of the range, as do our plug-in GLS test statistics.
The t statistics (ignoring dependence) are seen to be overly dispersed throughout the range, and are less sensitive to the real effects.

\subsection{Covariance estimation for $A$}
\begin{figure}
\includegraphics[width=0.95\textwidth]{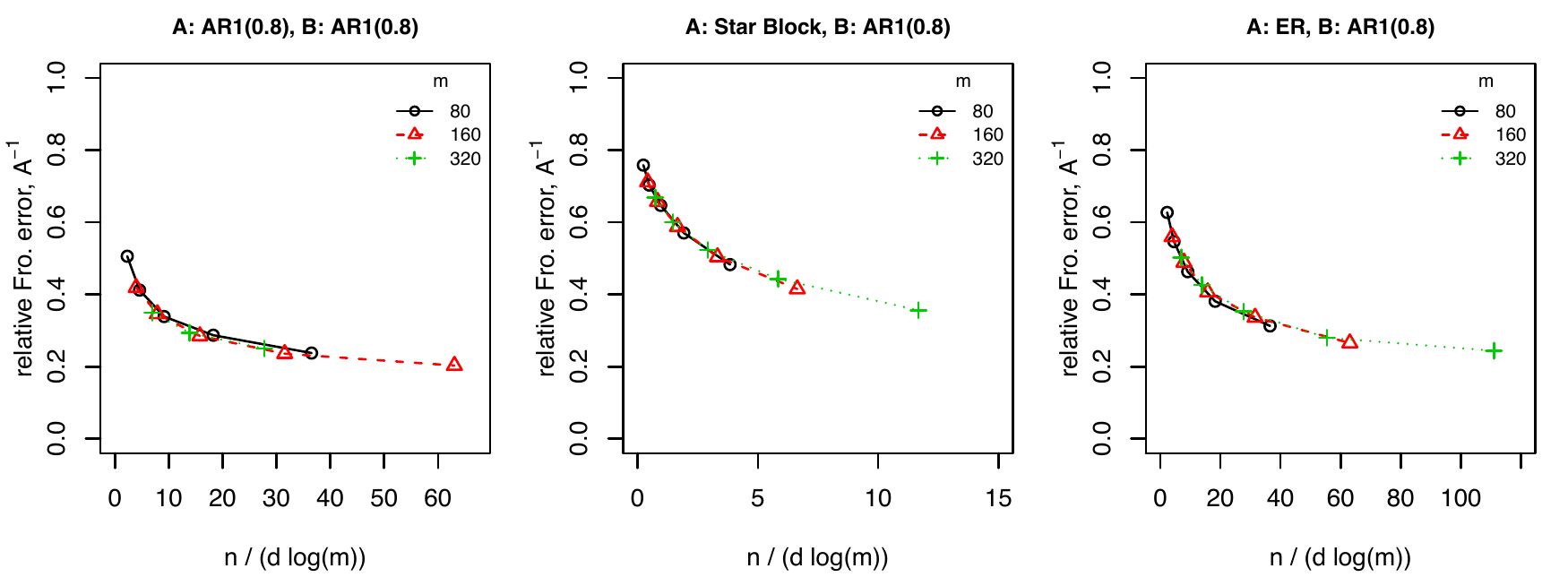}
\caption{Relative Frobenius error in estimating $A^{-1}$, as $n$ varies.  In each plot the matrix $B$ is AR1$(0.8)$ and $A$ is as indicated.  The vertical axis is relative Frobenius error, and the horizontal axis $n / (d \log(m))$, where $d$ is the maximum node degree.     The GLasso penalty is chosen to minimize the relative Frobenius error.  Each point is based on $250$ Monte Carlo replications.}
\label{fig::froberror}
\end{figure}

Figure \ref{fig::froberror} shows the relative Frobenius error in estimating $A^{-1}$ as $n$ grows, for fixed $m$.  The horizontal axis is $n / (d\log(m))$, scaled so that the curves align, where $d$ is the maximum node degree.  Because $\lVert A^{-1} \rVert_F$ is of order $\sqrt{m}$, the vertical axis essentially displays $\lVert \widehat{A}^{-1} - A^{-1} \rVert_F / \sqrt{m}$.  For estimating $A^{-1}$, the rate of convergence is of order $\sqrt{\log(m) / n}$.  For each of the three population structures, accuracy increases with respect to $n$.

% \begin{figure}
% \includegraphics[width=\textwidth]{RMSEsmallsampleSize}
% \caption{RMSE and relative Frobenius error in estimating $B^{-1}$, for smaller sample sizes $n$.  In each plot the matrix $A$ is AR1$(0.8)$.
% Each point is based on $250$ Monte Carlo replications.
% }
% \label{fig::smallSampleSize}
% \end{figure}

% \begin{figure}
% \includegraphics[width=\textwidth]{NinePlotsRankCorrpluginDecayExpm1is2e3}
% \caption{Correlation between ranking of estimated mean differences and true vector of mean differences, using a plug-in estimate of the theoretical threshold, rather than making a prior assumption about an upper bound on the number of genes with nonzero mean differences.
% Each point is based on $250$ Monte Carlo replications.
% }
 %\label{fig::smallSampleSize}
% \end{figure}

%!TEX root = submit.arxiv.tex

\section{Genomic study of ulcerative colitis}
\label{sec::UCData}

Ulcerative colitis (UC) is a chronic form of inflammatory bowel disease (IBD), resulting from inappropriate immune cell infiltration of the colon.
As part of an effort to better understand the molecular pathology of UC, \citet{lepage2011twin} reported on a study of mRNA expression in biopsy samples of the colon mucosal epithelium, with the aim of being able to identify gene transcripts that are differentially expressed between people with UC and healthy controls.  The study subjects were discordant identical twins, that is, monozygotic twins such that one twin has UC and the other does not.  This allows us to simultaneously explore dependences among samples (both within and between twins), dependences among genes, and mean differences between the UC and non-UC subjects.  The data set is available on the Gene Expression Omnibus, GEO accession GDS4519 \citep{edgar2002gene}.  

The data consist of 10 discordant twin pairs, for a total of 20 subjects.  Each subject's biopsy sample was assayed for mRNA expression, using the Affymetrix UG 133 Plus 2.0 array, which has 54,675 distinct transcripts.  Previous analyses of this data did not consider twin correlations or unanticipated non-twin correlations, and used very different methodology (e.g.\ Wilcoxon testing).
Roughly $70$ genes were found to be differentially expressed \citep{lepage2011twin}.

We applied our Algorithm 2 to the UC genomics data as follows.
First we selected the $2000$ most variable genes based on marginal variance and then rescaled each gene to have unit marginal variance.
We then applied step 1 of Algorithm 2, setting
$\lambda = 0.1 \approx 0.5 \left( \sqrt{\frac{\log(m)}{m}} + \frac{3}{n} \right)$,
with $m = 2000$ and $n = 20$.  For step 2 of the algorithm, we ranked the estimated mean differences, group centered the top ten, and globally centered the remaining genes.  We then re-calculated the Gram matrix $S_B$ using the centered data.  In step 3, following the Gemini approach, we applied the GLasso to $S_B$ using a regularization parameter $\lambda \approx 0.25(\sqrt{\log(m) / m} + 3/n)$.  We obtain estimated differences in means and test statistics via steps 4 through 6.
\begin{figure}[h!]
\includegraphics[width=0.95\textwidth]{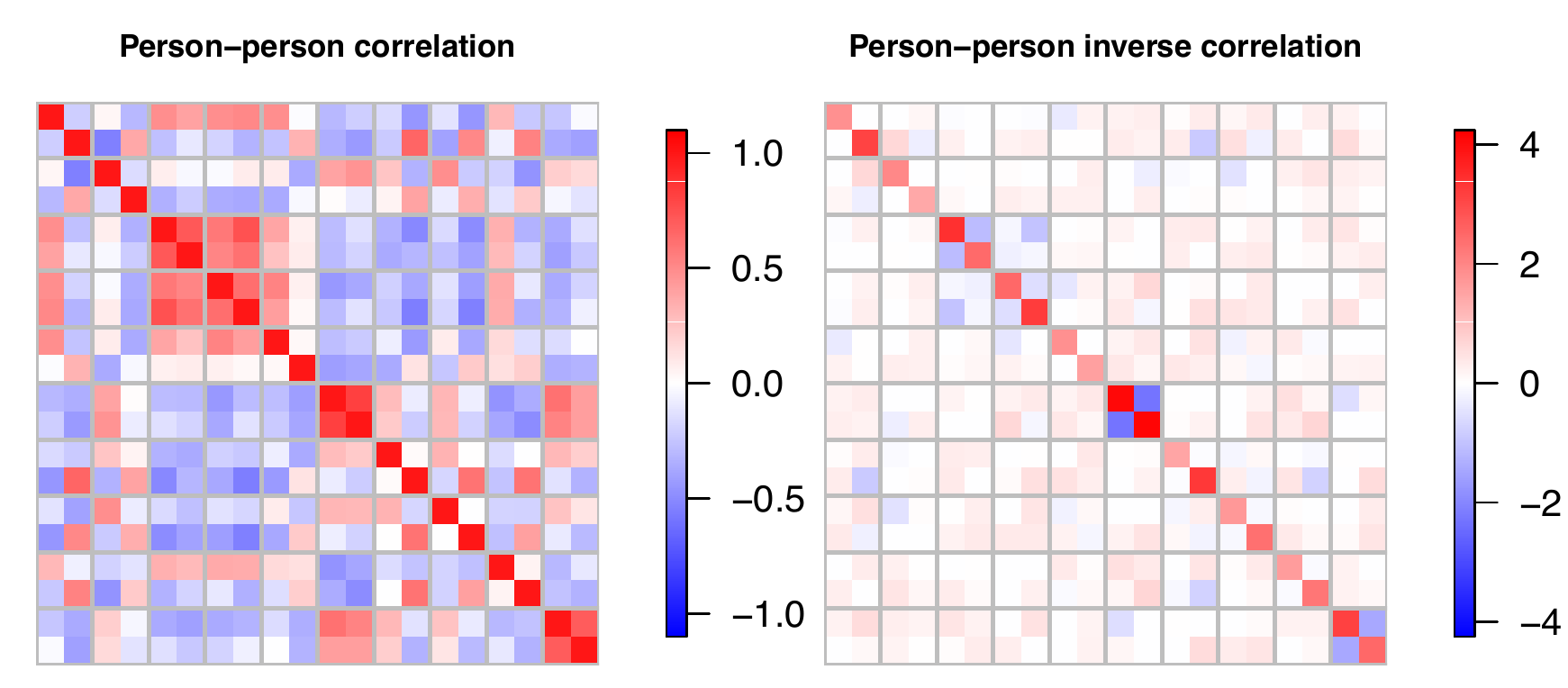}
\caption{Estimated person-person correlation matrix and its inverse,
estimated using the $2000$ genes with largest marginal variance.}
\label{fig::corrBHat}
\end{figure}
A natural analysis of these data using more standard methods would be a paired t-test for each mRNA transcript (paired by twin pair).  Such an approach is optimized for the situation where there is a constant level of correlation within all of the twin pairs, with no non-twin correlations.  However as in Efron (2008), we wish to accommodate unexpected correlations, which in this case would be correlations between non-twin subjects or a lack of correlation between twin subjects.  Our approach, developed in Section~\ref{sec::covEstimation}, does not require pre-specification or parameterization of the dependence structure, thus we were able to consider twin and non-twin correlations simultaneously.  Lepage et al. note that UC has lower heritability than other forms of IBD.  If UC has a relatively stronger environmental component, this could explain the pattern of correlations that we uncovered, as shown in Figure~\ref{fig::corrBHat}.  The samples are ordered so that twins are adjacent, corresponding to 2 by 2 diagonal blocks. The penalized inverse sample correlation matrix contains nonzero entries both within twin pairs and between twin pairs.

To also handle these unexpected non-twin correlations, we performed testing using Algorithm 2. We found only a small amount of evidence for differential gene expression between the UC and non-UC subjects.  Four of the adjusted p-values fell below a threshold of $0.1$, using the Benjamini-Hochberg adjustment; that is, four genes satisfied $2000 \widehat{p}_{(i)} / i < 0.1$, where $\widehat{p}_{(i)}$ is the $i^{th}$ order statistic of the p-values calculated using Algorithm 2, for $i = 1, \ldots, 2000$.  Based on our theoretical and simulation work showing that our procedure can successfully recover and accommodate dependence among samples, we argue that this is a more meaningful representation of the evidence in the data for differential expression compared to methods that do not adapt to dependence among samples.  Specifically, in Section~\ref{subsec::calib_test_stats} we demonstrate that our test statistics are properly calibrated and as a result have weaker (but more accurate) evidence for differential expression results. Below we argue that the sample-wise correlations detected by our approach would be expected to artificially inflate the evidence for differential expression.
% In Section~\ref{secComparisonUC} we show that CATE \citep[described in][]{wang2015confounder}, a method developed for differential expression analysis with confounded genomic data, has broadly similar results to our methods when applied to this data, but seems to have lower power as it produces weaker evidence for differential expression.

\subsection{Calibration of test statistics}
\label{subsec::calib_test_stats}

As noted above, based on the test statistics produced by Algorithm 2, we find evidence for only a small number of genes being differentially expressed.  This conclusion, however, depends on the test statistics conforming to the claimed null distribution whenever the group-wise means are equal.  In this section, we consider this issue in more detail.

\begin{figure}[tb] \centering
\includegraphics[width=0.9\textwidth]{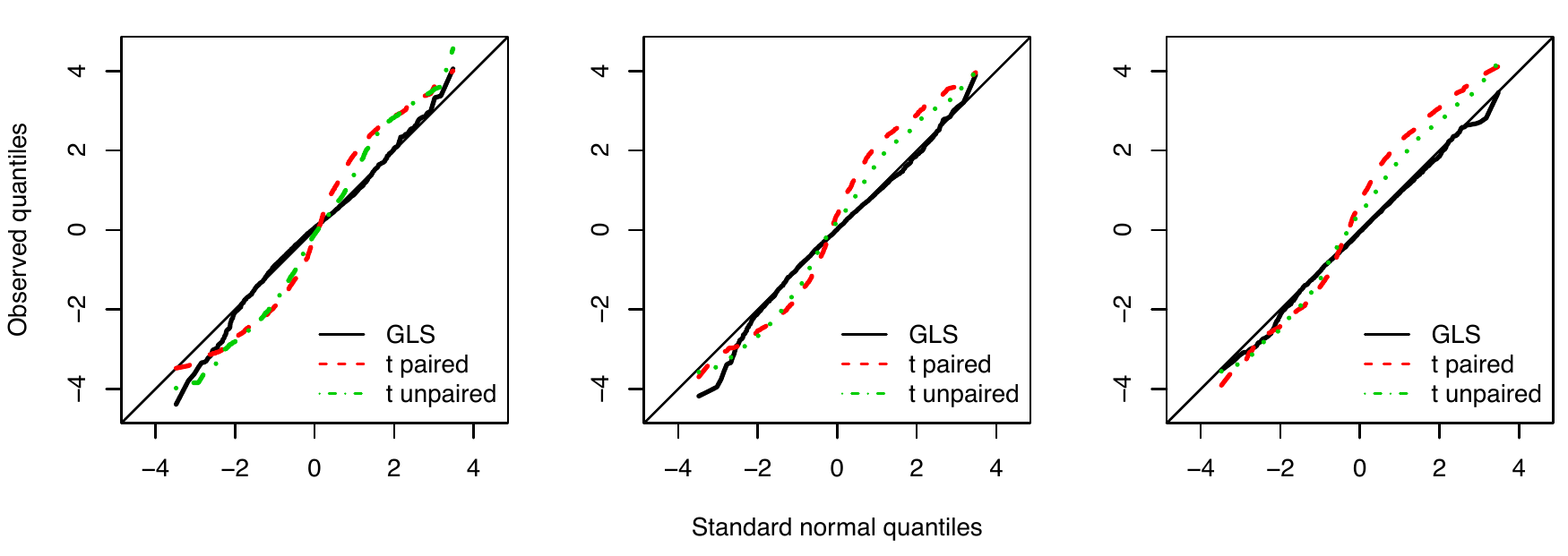}
\caption{Quantile plots of test statistics for three disjoint gene sets, each consisting of $2000$ genes.
The genes are partitioned based on marginal variance.  GLS statistics are taken from step 5 of Algorithm 2;
in step 2, the ten genes with greatest mean differences are selected for group centering.}
\label{fig::qqdata}
\end{figure}

The first plot of Figure \ref{fig::qqdata}
compares the empirical quantiles of $\Phi^{-1}(T_j)$
to the corresponding quantiles of a standard normal distribution, where $\Phi$ is the standard normal cdf and the $T_j$s are as defined
in \eqref{unpairedTstat}.
Plots 2 and 3 show the same information for successive non-overlapping blocks of two thousand genes sorted by marginal variance.
Since this is a discordant twins study, we also show results for the standard paired t statistics, pairing by twin.
In all cases, the paired and unpaired statistics are
more dispersed relative to the reference distribution.
By contrast, the central portion of the GLS test statistics coincide with the reference line.
Overdispersion of test statistics throughout their range is often taken to be evidence of miscalibration~\citep{devlin1999genomic}.
In this setting the GLS statistics are calibrated correctly under the null hypothesis, but the paired and unpaired t statistics are not.

\subsection{Stability of gene sets}
The motivation of our Algorithm 2 is that in many practical settings a relatively small fraction of variables may have differential means, and therefore it is advantageous to avoid centering variables presenting no evidence of a strong mean difference.  Here we assess the stability of the estimated mean differences as we vary the number of group centered genes in Algorithm 2.  To do so, we successively group center fewer genes, globally centering the remaining genes.

The iterative process is as follows.  Let $\hat{B}^{-1}_{(i)} \in \mathbb{R}^{n\times n}$ denote the estimate of $B^{-1}$ at iteration $i$, let $\widehat{\beta}_{(i)} \in \mathbb{R}^{2 \times m}$ denote the estimates of the group means $\beta$ on the $i$th iteration, let $\hat{\gamma}_{(i)}  \in {\mathbb R}^m$ denote the vector of differences in group means between the two groups, and let $\hat{\mu}_{(i)} \in \mathbb{R}^m$ denote vector of global mean estimates.  Let $\hat{\mu}(B^{-1}) \in \mathbb{R}^m$ denote the result of applying GLS with design matrix $D = 1_n$ to estimate the global means.

Initialize $\hat{\beta}_{(1)}$, $\hat{\mu}_{(1)}$ and $\hat{\gamma}_{(1)}$ using the sample means.  On the $i$th iteration,
\begin{enumerate}
\item Rank the genes according to $|\widehat{\gamma}_{(i-1)}|$.  Center the highest ranked $n_i'$ genes around $\widehat{\beta}_{(i-1)}$.  Center the remaining genes around $\hat{\mu}_{(i-1)}$.
\item Obtain $\widehat{B}^{-1}_{(i)}$ by applying GLasso to the centered data matrix from step 1.
\item Set $\hat{\beta}_{(i)} = \hat{\beta}(\hat{B}^{-1}_{(i)})$,   $\hat{\mu}_{(i)} = \hat{\mu}(\hat{B}^{-1}_{(i)})$, and $\hat{\gamma}_{(i)} = (1, -1) \hat{\beta}_{(i)}$.
\end{enumerate}

We assess the stability of the mean estimates by comparing the rankings of the genes across iterations of the algorithm.
Table \ref{topTenGenesInCommon} displays the number of genes in common out of the top
ten genes on each pair of iterations of the algorithm.
For example, three genes ranked in the top ten on the first iteration of the algorithm are also ranked in the top ten on the last iteration.
Iterations six through nine produce the same ranking of the top ten genes.
Three genes are ranked among the top ten on every iteration of the algorithm: DPP10-AS1, OLFM4, and PTN.  Web Supplement Table \ref{StabilitySimulationUC} shows simulations confirming these results.

\begin{table}[t]
%\begin{subfigure}{\textwidth}
\centering
\caption{Each iteration $k$ of the algorithm produces a ranking of all $2000$ genes.
For the top ten genes on each iteration, entry $(i, j)$ of the table shows the number
of genes in common in iterations $i$ and $j$ of the algorithm.
Note that the maximum possible value for any entry of the table is $10$;
if entry $(i, j)$ is $10$, then iterations $i$ and $j$ selected the same top ten genes.}
\label{topTenGenesInCommon}
\begin{tabular}{r|rrrrrrrrr}
  \hline
 & 1 & 2 & 3 & 4 & 5 & 6 & 7 & 8 & 9 \\
  \hline
1 &  10 &  10 &   7 &   5 &   5 &   3 &   3 &   3 &   3 \\
  2 &  10 &  10 &   7 &   5 &   5 &   3 &   3 &   3 &   3 \\
  3 &   7 &   7 &  10 &   6 &   5 &   3 &   3 &   3 &   3 \\
  4 &   5 &   5 &   6 &  10 &   8 &   5 &   5 &   5 &   5 \\
  5 &   5 &   5 &   5 &   8 &  10 &   7 &   7 &   7 &   7 \\
  6 &   3 &   3 &   3 &   5 &   7 &  10 &  10 &  10 &  10 \\
  7 &   3 &   3 &   3 &   5 &   7 &  10 &  10 &  10 &  10 \\
  8 &   3 &   3 &   3 &   5 &   7 &  10 &  10 &  10 &  10 \\
  9 &   3 &   3 &   3 &   5 &   7 &  10 &  10 &  10 &  10 \\
   \hline
\end{tabular}
\end{table}

\subsection{Stability analysis}
Table \ref{fdrStability} shows the number of genes that fall below an FDR threshold of $0.1$
on each iteration, for several values of the GLasso penalty $\lambda$.
The number of genes below the threshold is more sensitive to the number of group-centered genes than to the GLasso penalty parameter.
This is consistent with the first plot of Web Supplement Figure~\ref{fig::inference}
where the design effect (in the denominator of the test statistics) is likewise more sensitive to the
number of group centered genes than to the GLasso penalty.
When fewer than $128$ genes are group centered, the number of genes below an FDR threshold of $0.1$ is stable across the penalty parameters from $\lambda = 0.1$ to $\lambda = 0.8$.
\begin{table}[tb]
\centering
\caption{For the algorithm, this table shows the number of genes that are significant at an FDR level of $0.1$ on each iteration of the algorithm, for different values of the GLasso penalty $\lambda$.  The top row shows the number of genes group centered on each iteration.}
\label{fdrStability}
\begin{tabular}{|r|rrrrrrrrr|}
  \hline
n.group & 2000 & 1024 & 512 & 256 & 128 & 64 & 32 & 16 & 8 \\
  \hline
    $\lambda = 0.1$ & 1006 & 913 & 327 & 14 & 3 & 1 & 1 & 1 & 1 \\
    $\lambda = 0.2$ & 865 & 806 & 262 & 2 & 1 & 1 & 1 & 1 & 0 \\
    $\lambda = 0.3$ & 778 & 789 & 303 & 3 & 1 & 1 & 0 & 0 & 0 \\
    $\lambda = 0.4$ & 706 & 774 & 452 & 3 & 1 & 0 & 0 & 0 & 0 \\
    $\lambda = 0.6$ & 657 & 751 & 587 & 19 & 1 & 1 & 0 & 0 & 0 \\
    $\lambda = 0.8$ & 628 & 699 & 493 & 30 & 1 & 1 & 1 & 1 & 1 \\
   \hline
\end{tabular}
\end{table}

% \clearpage

% \input{proofmain}

%!TEX root = submit.arxiv.tex

\section{Conclusion}
\label{sec::conclude}
It has long been known that heteroscedasticity and dependence
between observations impacts the precision and degree of uncertainty for estimates of mean values and regression coefficients.
Further, data that are modeled for convenience as being independent observations may in fact show
unanticipated dependence ~\citep{kruskal1988miracles}.
This has motivated the development of numerous statistical methods, including generalized/weighted least squares (GLS/WLS),
mixed effect models, and generalized estimating equations (GEE).
Our approach utilizes recent advances in high dimensional statistics to permit estimation of
an inter-observation dependence structure (reflected in the matrix $B$ in our model).
Like GLS/GEE, we use an approach that alternates between mean and covariance estimation, but limit it in Algorithm 1 to a mean estimation step, followed by a covariance update, followed by a mean update, with an additional covariance and mean update if Algorithm 2 is used.  We provide convergence guarantees and rates for both algorithms.

Estimation of dependence or covariance structures usually requires some form of replication,
and/or strong models.  We require a relatively weak form of replication and a relatively weak model.
In our framework, the dependence among observations must be common (up to proportionality)
across a set of ``quasi-replicates'' (the columns of $X$, or the genes in our UC example).
These quasi-replicates may be statistically dependent, and may have different means.
We also require the precision matrices for the dependence structures to be sparse,
which is a commonly used condition in recent high-dimensional analyses.

In addition to providing theoretical guarantees, we also show through simulations
and a genomic data analysis that the approach improves estimation accuracy for the mean structure,
and appears to mitigate test statistic overdispersion, leading to test statistics that do not require post-hoc correction.
The latter observation suggests that undetected dependence among observations
may be one reason that genomic analyses are sometimes less reproducible than
traditional statistical methods would suggest, an observation made previously
by \citet{Efr09} and others.

Although our theoretical analysis guarantees the convergence of our procedure
even with a single observation of the random matrix X, there are reasons to expect this estimation problem to be fundamentally challenging.
One reason for this as pointed out by \citet{Efr09} and subsequently explored by ~\citet{Zhou14a},
is that the row-wise and column-wise dependence structures are somewhat non-orthogonal,
in that row-dependence can ``leak'' into the estimates of column-wise dependence, and vice-versa.
Our results suggest that while row-wise correlations make it more difficult to estimate column-wise correlations (and vice-versa),
when the emphasis is on mean structure estimation, even a somewhat rough estimate of the dependence structure ($B$)
can substantially improve estimation and inference.

\bibliographystyle{ims}
\bibliography{subgaussian}{}

\appendix

%\newpage

\clearpage
\setcounter{page}{1}
\setcounter{table}{0}
\setcounter{figure}{0}
\setcounter{equation}{0}

\setcounter{definition}{0}
\setcounter{theorem}{0}
\renewcommand\thetheorem{S\arabic{theorem}}
\renewcommand\thefigure{S\arabic{figure}}
\renewcommand\thetable{S\arabic{table}}
\renewcommand\theequation{S\arabic{equation}}

  \begin{center}
    {\Large\bf Supplement to ``Joint mean and covariance estimation with unreplicated
    matrix-variate data''} \\ 
    {\large Michael Hornstein, Roger Fan, Kerby Shedden, Shuheng Zhou
    
     Department of Statistics, University of Michigan} 
\end{center}

\section*{Outline}
%\etocdepthtag.toc{mtappendix}
%\etocsettagdepth{mtchapter}{none}
%\etocsettagdepth{mtappendix}{subsection}
%\tableofcontents??
We provide additional simulation and data analysis results in Section \ref{sec::simulationAppend} and \ref{sec::dataAppend}.  We state some preliminary results and notation in Section \ref{sec::apppreliminary}.  We prove Theorem \ref{thm::GLSFixedB} in Section \ref{sec::proofsOfTheorems} and Corollary \ref{theoremInference} in Section~\ref{proofTheoremInference}.  We prove Theorem \ref{mainTheoremGroupCentering} in Section \ref{sec::ProofMainThmPartI}, with additional lemmas proved in Section~\ref{sec::proofsforTheorem2}.  We prove entrywise convergence of the sample correlation matrices for Algorithm 1 in Section~\ref{app::entrywise_sample_corr}.  We prove Theorem \ref{mainTheoremModSel} in Section \ref{sec::proofTheorem3}, and we prove additional lemmas used in the proof of Theorem \ref{mainTheoremModSel} in Section \ref{sec::LemmasForTheorem3}.
In Section~\ref{secComparison} we provide additional comparisons between our method and some related methods on both simulated and real data.

%!TEX root = submit.arxiv.tex

\section{Additional simulation results}
\label{sec::simulationAppend}

\begin{figure}[tb] \centering 
\includegraphics[width=0.9\textwidth]{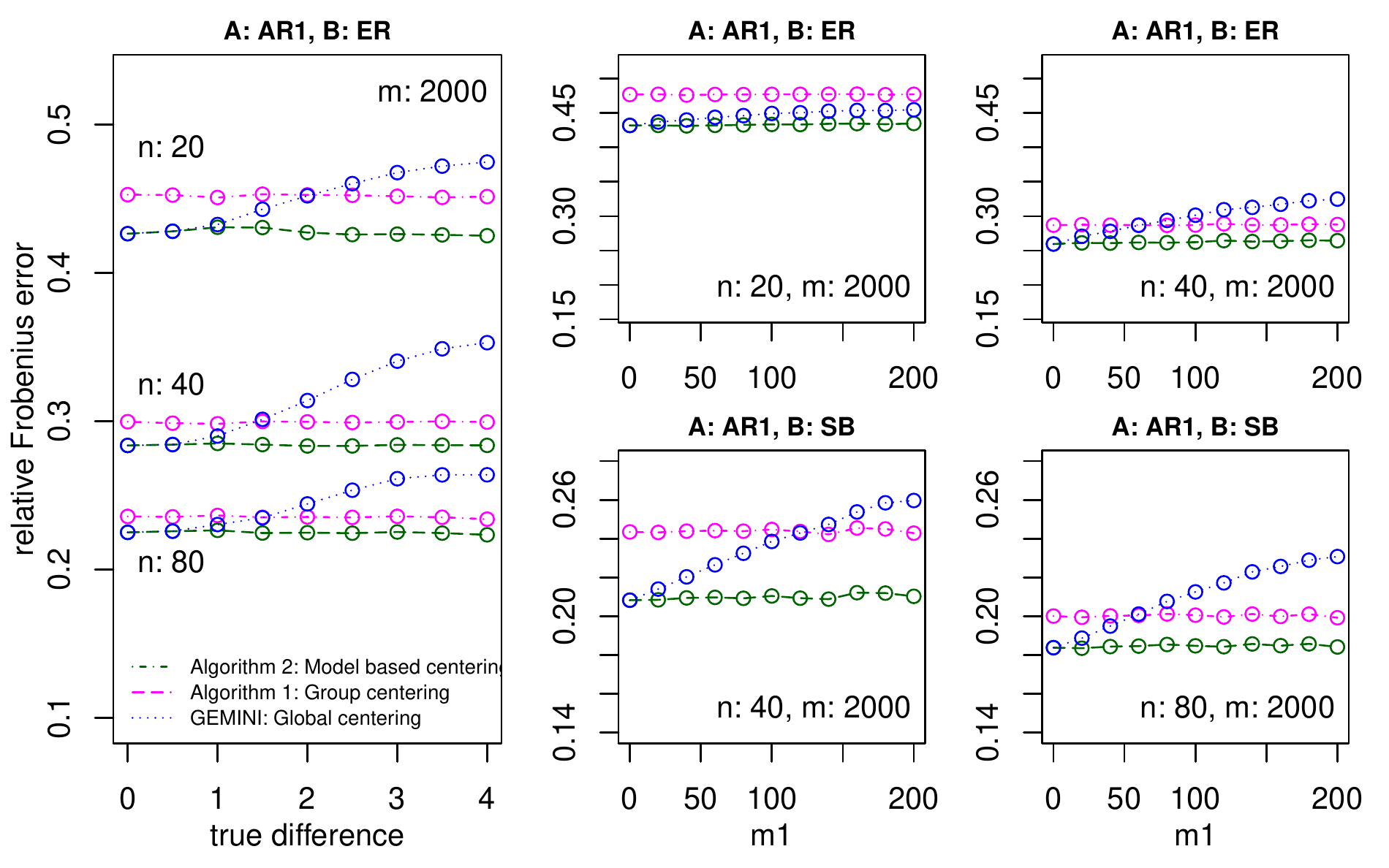}
\caption{Performance of Gemini, Algorithm 1, and Algorithm 2 for estimating $B$ under different mean and covariance structures. As the sample size increases, we can see that Algorithm 1 improves relative to Gemini and begins to catch up to Algorithm 2. Gemini's performance always degrades as the true differences grow or more differentially expressed genes  are added, while Algorithm 1 and 2 are stable. We set $B^{-1}$ as Erd\H{o}s-R\'{e}nyi (ER) or star-block with blocks of size 4 (SB). All plots use $A$ from an AR1$(0.8)$ model with $m=2000$ and are averaged over 200 replications. In the left plot the first $50$ genes are differentially expressed at the level specified on the $x$-axis. As indicated, the three groups of lines correspond to $n=20$, $40$, and $80$. In the right two columns there are \texttt{m1} number of genes with exponentially decaying true differences between groups, scaled so that the largest difference is 5 (resulting in an average difference of approximately 1). }
\label{fig::gemini_comp}
\end{figure}

Figure~\ref{fig::gemini_comp} demonstrates the effect of mean structure on covariance estimation. As expected, when there is no mean structure Gemini performs competitively. As more mean structure is added, however, its performance quickly decays to be worse than Algorithm 2. This also provides evidence that the plug-in estimator $\widehat{\tau}_{\text{init}}$ used in Algorithm 2 is appropriately selecting genes to group center, as when there are no or very few differentially expressed genes Algorithm 2 is still never worse than Gemini. Algorithm 1 does not perform as well as Algorithm 2 but still tends to eventually outperform Gemini as more mean structure is added. As the sample size increases, the difference between Algorithm 2 and Algorithm 1 decreases as the added noise from group centering becomes less of a factor. We still recommend using Algorithm 2 in most realistic scenarios, but this reinforces our theoretical finding that the two algorithms have the same error rates.

\section{Additional data analysis}
\label{sec::dataAppend}

%\begin{figure}[h] \centering
%\includegraphics[width=0.9\textwidth]{inferenceplot}
%\caption{The first plot displays the estimated design effect vs. the penalty multiplier for Algorithm 2.
%Each curve corresponds to a different number of columns being group centered.
%As the curves go from top to bottom, the number of group centered columns increases from $10$ to $2000$.
%The second plot shows a quantile plot of test statistics from the data vs. simulated test statistics;
% in the simulation, the population person-person covariance matrix is $\widehat{B}$, as estimated from the UC data.}
%\label{fig::inference}
%\end{figure}

%\begin{figure}[h]
%\centering
%\includegraphics[width=\textwidth]{qqcombined}
%\caption{Quantile plot and inverse covariance graphs.  The first two plots correspond to $\lambda =0.4$ and $128$ group centered genes.  The third plot corresponds to $\lambda = 0.5$ and $128$ group centered genes.  Green circles correspond to twins with UC, orange circles to twins without UC.  Twins are aligned vertically.} \label{fig::icovplot}
%\end{figure}

\begin{figure}
\centering
\begin{subfigure}[b]{\textwidth}
   \includegraphics[width=1\linewidth]{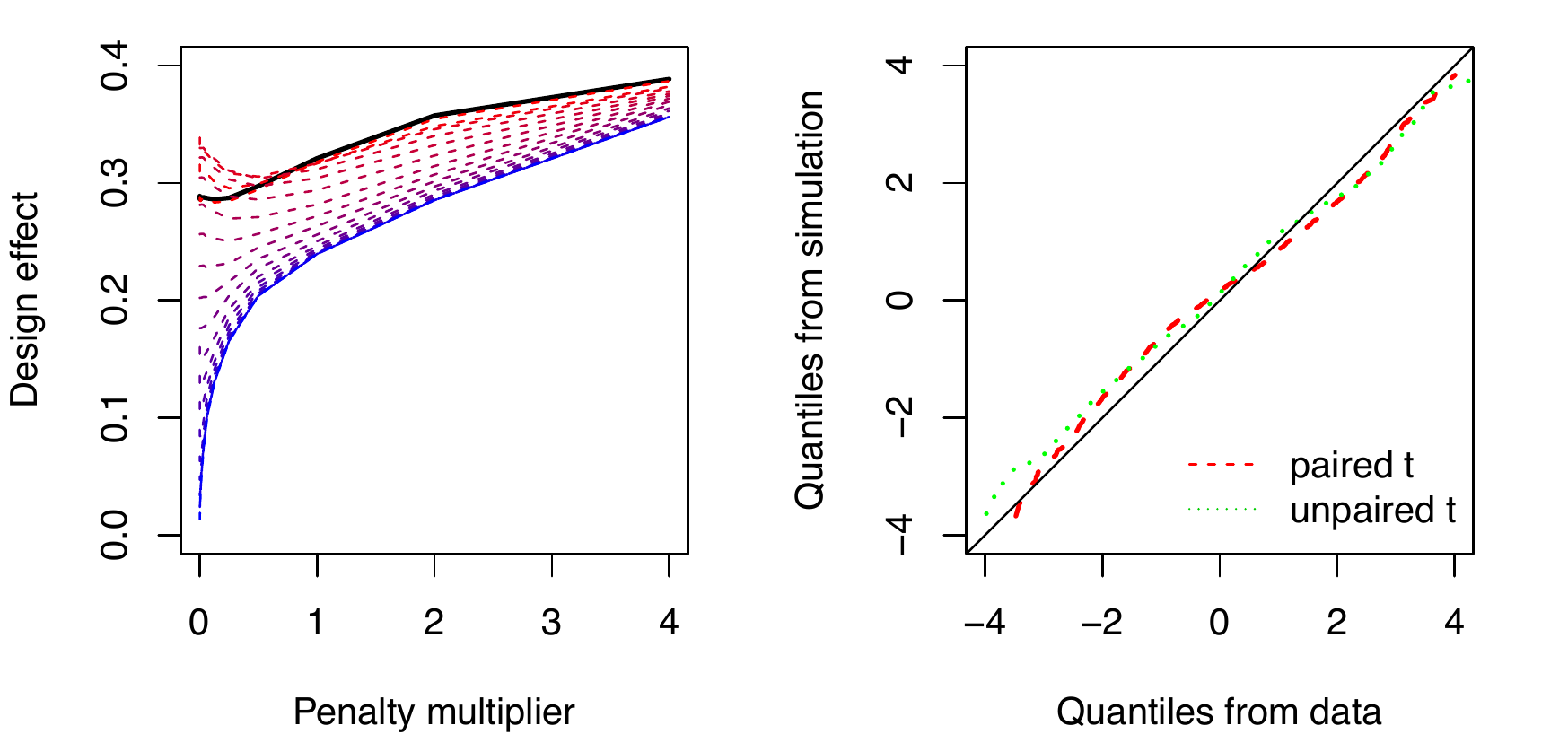}
   \caption{The first plot displays the estimated design effect vs. the penalty multiplier for Algorithm 2.
Each curve corresponds to a different number of columns being group centered.
As the curves go from top to bottom, the number of group centered columns increases from $10$ to $2000$.
The second plot shows a quantile plot of test statistics from the data vs. simulated test statistics;
 in the simulation, the population person-person covariance matrix is $\widehat{B}$, as estimated from the UC data.}
   \label{fig::inference} 
\end{subfigure}

\begin{subfigure}[b]{\textwidth}
   \includegraphics[width=1\linewidth]{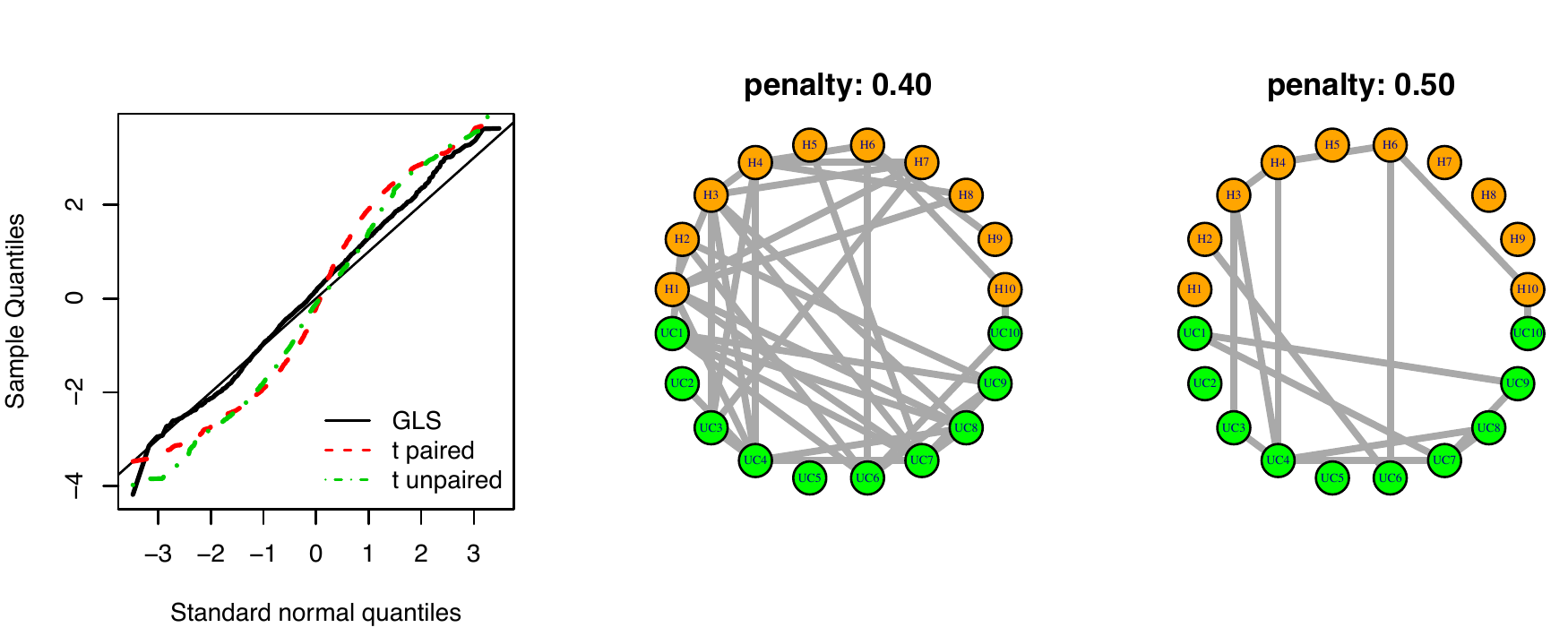}
   \caption{Quantile plot and inverse covariance graphs.  The first two plots correspond to $\lambda =0.4$ and $128$ group centered genes.  The third plot corresponds to $\lambda = 0.5$ and $128$ group centered genes.  Green circles correspond to twins with UC, orange circles to twins without UC.  Twins are aligned vertically.}
   \label{fig::icovplot}
\end{subfigure}
\caption[]{}
\end{figure}

As discussed in Section~\ref{sec::GLSFixedBtheorem}, it is particularly important that the design effect is accurately estimated in order for the test statistics to be properly calibrated.
The first plot of Figure~\ref{fig::inference} displays the sensitivity of the estimated design effect \eqref{designEffect} for Algorithm 2 to the GLasso penalty parameter and the number of group centered columns.  In the case that all columns are group centered, Algorithm 2 reduces to Algorithm 1.  If we group center all genes, the estimated design effect is sensitive to the penalty parameter, but if we group center a small proportion of genes, it is less sensitive to the penalty parameter.  This is further evidence that it may be advantageous to avoid over-centering the data when the true mean difference vector $\gamma$ may be sparse.
The second plot of Figure~\ref{fig::inference} shows a quantile plot comparing the distribution
of test statistics from the UC data to test statistics from a simulation whose population correlation structure is matched to the UC data.  The quantile plot reveals that we can reproduce the pattern of overdispersion in the test statistics using simulated data having person-person as well as gene-gene correlations.
Such correlations therefore provide a possible explanation for the overdispersion of the test statistics.

Figure \ref{fig::icovplot} displays a quantile plot and inverse covariance graph for $\lambda = 0.4$ and $128$ group centered genes.  Under these settings the test statistics appear correctly calibrated, coinciding with the central portion of the reference line.  Furthermore, the inverse covariance graph is sparse (38 edges).  In the inverse covariance graph, there are more edges between subjects with UC than between the healthy subjects, which could be explained by the existence of subtypes of UC inducing correlations between subsets of subjects. 
The third plot of Figure~\ref{fig::icovplot} displays a sparser inverse covariance graph, corresponding to a larger penalty $\lambda = 0.5$.  There are three edges between twin pairs, and there are more edges between subjects with UC than between those without UC. 

\subsection{Stability simulation} \label{sec::StabilitySim}

Table \ref{StabilitySimulationUC} shows the results from a simulation analogous to Table \ref{topTenGenesInCommon}, demonstrating stability across iterations of the procedure. Iteration 1 begins by group centering $1280$ genes and this number is halved in each successive iteration. We can see from the table that the gene rankings generated by Algorithm 2 are robust to misspecifying the number of differentially expressed genes. When the number of group centered genes is $160$ or below (iterations 4 through 8), the commonly selected genes among the top $20$ genes are stable.  Furthermore, the true positives remain stable as we decrease the amount of genes centered, while the false positives decrease.
\begin{table}[tb]
\centering
\caption{Number of genes in common among genes ranked in the top $20$ when different numbers of genes are group centered.  This simulation is analogous to Table \ref{topTenGenesInCommon}.
Note that the maximum possible value for any entry of the table is $20$; if entry $(i, j)$ is $20$, then iterations $i$ and $j$ selected the same top twenty genes.
The first $10$ genes have a difference of 1.5 and the second $10$ have a difference of $1$. All remaining genes have a true mean difference of zero. We use $B$ as estimated from the UC data, and $A$ is from an $\text{AR1}(0.8)$ model. These simulations have $n  = 20$ individuals and 2000 genes and are averaged over $200$ replications. The last two rows display the average number of true and false positives among the genes ranked in the top 20 of each iteration.}
\label{StabilitySimulationUC}
\begin{tabular}{r|rrrrrrrr}
  \hline
  & 1 & 2 & 3 & 4 & 5 & 6 & 7 & 8 \\
  \hline
1 & 20.0 & 17.6 & 15.8 & 14.8 & 14.3 & 14.0 & 14.0 & 13.9 \\
  2 & 17.6 & 20.0 & 17.9 & 16.8 & 16.2 & 15.9 & 15.8 & 15.8 \\
  3 & 15.8 & 17.9 & 20.0 & 18.7 & 18.1 & 17.8 & 17.7 & 17.6  \\
  4 & 14.8 & 16.8 & 18.7 & 20.0 & 19.3 & 19.0 & 18.9 & 18.8 \\
  5 & 14.3 & 16.2 & 18.1 & 19.3 & 20.0 & 19.6 & 19.5 & 19.4 \\
  6 & 14.0 & 15.9 & 17.8 & 19.0 & 19.6 & 20.0 & 19.8 & 19.7 \\
  7 & 14.0 & 15.8 & 17.7 & 18.9 & 19.5 & 19.8 & 20.0 & 19.8 \\
  8 & 13.9 & 15.8 & 17.6 & 18.8 & 19.4 & 19.7 & 19.8 & 20.0 \\
  \hline
  TP & 12.7 & 14.3 & 15.6 & 16.4 & 16.7 & 16.8 & 16.8 & 16.8 \\
  FP & 7.3 & 5.7 & 4.4 & 3.6 & 3.3 & 3.2 & 3.2 & 3.2 \\
   \hline
\end{tabular}
\end{table}

%!TEX root = submit.arxiv.tex

\section{Preliminary results}
\label{sec::apppreliminary}

In this section, we refresh notation and introduce propositions that are shared in the proofs of the theorems.  For convenience, we first restate some notation.
\begin{align}
&D = \begin{bmatrix}
  \ones_{n_1} & 0 \\
  0 & \ones_{n_2}
  \end{bmatrix} \in \mathbb{R}^{n \times 2}
  \label{def:D} \\
&\Omega = (D^T B^{-1} D)^{-1} \text{ and } \OmegaFixed = (D^T \BFixed^{-1} D)^{-1}
  \label{def:Omega_app}\\
&\Delta = \BFixed^{-1} - B^{-1} \\
&\widehat{\beta}(\widehat{B}^{-1}) = (D^T \widehat{B}^{-1} D)^{-1} D^T \widehat{B}^{-1} X \in \mathbb{R}^{2 \times m}
\end{align}
When $D$ has the form (\ref{def:D}), the singular values are
$\sigma_{\max}(D) = \sqrt{n_{\max}}$ and $\sigma_{\min}(D) =
\sqrt{n_{\min}}$.
The condition number is $\kappa(D) = \sigma_{\max}(D) /
\sigma_{\min}(D) = \sqrt{n_{\text{ratio}}}$
where $n_{\text{ratio}} = \max(n_1, n_2) / \min(n_1, n_2)$.

We first state some convenient notation and bounds.
\begin{align}
& r_a := a_{\max} / a_{\min} \text{ and } r_b := b_{\max} / b_{\min}; \notag \\
& 1/ \varphi_{\min}(A) = \lVert A^{-1} \rVert_2 \leq \lVert \rho(A)^{-1} \rVert_2 / a_{\min} = \frac{1}{a_{\min} \varphi_{\min}(\rho(A))}, \label{ReciprocalMinEvalA} \\
& 1/ \varphi_{\min}(B) = \lVert B^{-1} \rVert_2 \leq \lVert \rho(B)^{-1} \rVert_2 / b_{\min} = \frac{1}{b_{\min} \varphi_{\min}(\rho(B))}, \label{ReciprocalMinEvalB} \\
&1/ \varphi_{\min}(\rho(A)) = \lVert \rho(A)^{-1} \rVert_2 \leq a_{\max} \lVert A^{-1} \rVert_2, \label{ReciprocalMinEvalCorrA} \\
&1/ \varphi_{\min}(\rho(B)) = \lVert \rho(B)^{-1} \rVert_2 \leq b_{\max} \lVert B^{-1} \rVert_2 \label{ReciprocalMinEvalCorrB} \\
& \lVert A \rVert_2 \leq a_{\max} \lVert \rho(A) \rVert_2, \quad
\lVert B \rVert_2 \leq b_{\max} \lVert \rho(B) \rVert_2,
\label{boundOpNormAandB} \\
& \lVert \rho(A) \rVert_2 \leq \lVert A \rVert_2 / a_{\min}, \quad
\text{ and } \quad \lVert \rho(B) \rVert_2 \leq \lVert B \rVert_2 /
b_{\min}.\label{boundOpNormAandBII}
\end{align}

The eigenvalues of the correlation matrices satisfy
\ben
\label{evalsInequalityCorrA}
0 < \varphi_{\min}(\rho(A)) \leq 1 \leq \varphi_{\max}(\rho(A)) \; \text{
  and } \;  0 < \varphi_{\min}(\rho(B)) \leq 1 \leq \varphi_{\max}(\rho(B)).
\een

In the remainder of this section, we state preliminary results and highlight important intermediate steps that are used in the proofs of Theorems \ref{thm::GLSFixedB} and \ref{mainTheoremGroupCentering}.  First we state propositions used in mean estimation for Theorems \ref{thm::GLSFixedB} and \ref{mainTheoremGroupCentering}.
% Then, we state Theorem \ref{thm::frob}, which is used in the proof of correlation estimation for Algorithm 1.

\subsection{Propositions}
We now state propositions used in the proofs of Lemmas \ref{HansonWrightBetaBinvBetaStar} and \ref{BetaBtildeBetaBStarHansonWright}.  We defer the proof of Proposition~\ref{L2boundInverseOfDtransBinvD}  to Section~\ref{sec::proofofBinvD}.
\begin{proposition}
\label{L2boundInverseOfDtransBinvD}
For $\Omega$ as defined in (\ref{def:Omega_app}) and some design matrix $D$,
\[
  \lVert \Omega \rVert_2 \leq \lVert B \rVert_2 / \sigma_{\min}^2(D)
\]
In the case that $D$ is defined as in (\ref{def:D}), we have $\lVert \Omega \rVert_2 \leq \lVert B \rVert_2 / n_{\min}$.

Furthermore,
\begin{equation} \label{minEvalOmega}
\lambda_{\min}(\Omega) \geq \frac{\lambda_{\min}(B) }{n_{\max}}.
\end{equation}
\end{proposition}
We state the following perturbation bound.
\begin{theorem}[Golub \& Van Loan, Theorem 2.3.4] \label{GolubInversePerturbation}
If $A$ is invertible and $\lVert A^{-1} E \rVert_p < 1$, then $A + E$ is invertible and
\[
  \lVert (A + E)^{-1} - A^{-1} \rVert_p \leq \frac{\lVert E \rVert_p \lVert A^{-1} \rVert_p^2}{1 - \lVert A^{-1} E \rVert_p} \leq \frac{\lVert E \rVert_p \lVert A^{-1} \rVert_p^2}{1 - \lVert A^{-1} \rVert_p \lVert E \rVert_p}.
\]
\end{theorem}

In Proposition \ref{L2BoundDTBiHatDinvMinusDTBiDInv},
we provide auxiliary upper bounds that depend on $\lVert \Delta \rVert_2$, $\lVert B \rVert_2$, $\kappa(D)$, and $\sigma_{\min}(D)$. We defer the proof of
Proposition \ref{L2BoundDTBiHatDinvMinusDTBiDInv} to the end of this section, for clarity of presentation.

\begin{proposition}
\label{L2BoundDTBiHatDinvMinusDTBiDInv}
Let $\Delta = \BFixed^{-1} - B^{-1}$.
\begin{align}
    \delta_0(\Delta) &:= \lVert \OmegaFixed - \Omega \rVert_2 \leq \frac{1}{\sigma_{\min}^2(D)} \frac{\lVert B \rVert_2^2 \lVert \Delta \rVert_2}{1/ \kappa^2(D) -  \lVert B \rVert_2 \lVert \Delta \rVert_2}
    \label{def:delta0} \\
  \delta_1(\Delta) &:= \left \lVert \Omega D^T \Delta  \right \rVert_2 \leq \sigma_{\max}(D) \lVert B \rVert_2 \lVert \Delta \rVert_2 / \sigma_{\min}^2(D) = \frac{\sqrt{n_{\max}}}{n_{\min}} \lVert B \rVert_2 \lVert \Delta \rVert_2. \label{def:delta1}
\end{align}
If $\lVert (D^T B^{-1}D)^{-1} D^T \Delta D \rVert_2 < 1$, then
\begin{align}
 \delta_2(\Delta) &:= \left \lVert  \left(\OmegaFixed - \Omega \right) D^T \Delta \right \rVert_2 \leq \frac{\kappa(D)}{\sigma_{\min}(D)} \frac{\lVert B \rVert_2^2 \lVert \Delta \rVert_2^2}{1/ \kappa^2(D) -  \lVert B \rVert_2 \lVert \Delta \rVert_2} \label{L2BoundOmegaTildeMinusOmegaDTDelta} \\
 \delta_3(\Delta) &:= \left \lVert \left(\OmegaFixed - \Omega \right) D^T B^{-1} \right \rVert_2 \leq \frac{\kappa(D)}{\sigma_{\min}(D)} \frac{\lVert B \rVert_2^2 \lVert B^{-1} \rVert_2 \lVert \Delta \rVert_2}{1/ \kappa^2(D) -  \lVert B \rVert_2 \rVert_2 \lVert \Delta \rVert_2}  \label{L2BoundOmegaTildeMinusOmegaDTBinv}
\end{align}
\end{proposition}

The following proposition is a corollary of Proposition \ref{L2BoundDTBiHatDinvMinusDTBiDInv}.
\begin{proposition} \label{L2BoundsOmegaTwoGroup}
When $D$ has the form (\ref{def:D}), and $\Omega$ is as defined in \eqref{def:Omega_app},
\begin{align*}
\delta_0(\Delta)
  &= \lVert \OmegaFixed - \Omega \rVert_2 \leq \frac{1}{n_{\min}} \frac{\lVert B \rVert_2^2 \lVert \Delta \rVert_2}{1/n_{\text{ratio}} -  \lVert B \rVert_2 \lVert \Delta \rVert_2} \\
\delta_1(\Delta)
  &= \left \lVert \Omega D^T \Delta  \right \rVert_2 \leq \frac{\sqrt{n_\text{ratio}}}{\sqrt{n_{\min}}} \lVert B \rVert_2 \lVert \Delta \rVert_2 \\
\delta_2(\Delta)
  &= \left \lVert  \left(\OmegaFixed - \Omega \right) D^T \Delta \right \rVert_2 \leq \frac{\sqrt{n_\text{ratio}}}{\sqrt{n_{\min}}} \frac{\lVert B \rVert_2^2 \lVert \Delta \rVert_2^2}{1/n_{\text{ratio}} -  \lVert B \rVert_2 \lVert \Delta \rVert_2}
% \delta_3(\Delta)
%   &= \left \lVert \left(\Omega_1 - \Omega \right) D^T B^{-1} \right \rVert_2 \leq \frac{\sqrt{n_\text{ratio}}}{\sqrt{n_{\min}}} \frac{\lVert B \rVert_2^2 \lVert B^{-1} \rVert_2 \lVert \Delta \rVert_2}{1/n_{\text{ratio}} -  \lVert B \rVert_2 \rVert_2 \lVert \Delta \rVert_2}
\end{align*}
\end{proposition}

Let $K$ be defined as in Theorem \ref{thm::GLSFixedB}.
We express the entrywise rates of convergence of the sample correlation
matrices $\widehat{\Gamma}(B)$
and $\widehat{\Gamma}(A)$, respectively, in terms of the following quantities:
\ben
\label{entrywiseRateBcorr}
\widetilde{\alpha} = C_A K \frac{\log^{1/2}(m)}{\sqrt{m}}
\left(1 +  \frac{\norm{B}_1}{n} \right) + \frac{\lVert B
  \rVert_1}{n_{\min}} \; \text{ and } \;
\widetilde{\eta} = C_B K \frac{\log^{1/2}(m \vee n)}{\sqrt{n}} +
\frac{\lVert B \rVert_1}{n}.
\een

\section{Proof of Theorem \ref{thm::GLSFixedB} and Corollary \ref{theoremInference}}
\label{sec::proofsOfTheorems}
\subsection{Proof of Theorem \ref{thm::GLSFixedB}}
\label{sec::ProofThm1}
\begin{proofof2}
Let $\BFixed \in \mathbb{R}^{n \times n}$ denote a fixed positive
definite matrix. Let $D$ be as defined as in \eqref{meanMatrixTwoGroups}.  Define $\DeltaFixed = \BFixed^{-1} - B^{-1}$ and
\begin{equation} \label{def:Omega}
\Omega = (D^T B^{-1} D)^{-1} \text{ and } \OmegaFixed = (D^T \BFixed^{-1} D)^{-1}.
\end{equation}
Note that we can decompose the error for all $j$ as
\begin{equation} \label{GLSDecompTriangleIneq}
  \lVert \widehat{\beta}_j(\BFixed^{-1}) - \beta_j^* \rVert_2
    \leq \lVert \widehat{\beta}_j(B^{-1}) - \beta_j^* \rVert_2 + \lVert \widehat{\beta}_j(\BFixed^{-1}) - \widehat{\beta}_j(B^{-1}) \rVert_2 =: \text{I} + \text{II}.
\end{equation}
We will use the following lemmas, which are proved in subsections
\ref{sec::rateBetaHatBHatinvBetaHatBinv}
and \ref{sec::rateBetaHatBinvBetaStar}, to bound these two terms on the right-hand side, respectively.
\begin{lemma} \label{HansonWrightBetaBinvBetaStar}
Let $\mathcal{E}_2$ denote the event
\begin{equation} \label{eventE2}
  \mathcal{E}_2 = \left\{ \lVert \widehat{\beta}_j(B^{-1}) - \beta_j^* \rVert_2 \leq s_{n, m} \right \},  \quad \text{with} \quad s_{n, m} = C_3 d^{1/2} \sqrt{\frac{\log(m) \lVert B \rVert_2}{n_{\min}}}.
\end{equation}
Then $P(\mathcal{E}_2) \geq 1 - 2 / m^d$.
\end{lemma}
\begin{lemma} \label{BetaBtildeBetaBStarHansonWright}
Let $\BFixed \in \mathbb{R}^{n \times n}$ denote a fixed matrix such that $\BFixed \succ 0$.  Let $X_j \in \mathbb{R}^n$ denote the $j$th column of $X$, where $X$ is a realization of model (\ref{modelMeanCov}).  Let $\mathcal{E}_3$ denote the event
\begin{equation}
\mathcal{E}_3 = \left \{  \lVert \widehat{\beta}_j(\BFixed^{-1}) - \widehat{\beta}_j(B^{-1}) \rVert_2 \leq t_{n, m} \right\}, \quad \text{ with} \quad t_{n, m} = \widetilde{C} n_{\min}^{-1/2} \lVert \DeltaFixed \rVert_2.
\end{equation}
for some absolute constant $\widetilde{C}$.  Then $P(\mathcal{E}_3) \geq 1 - 2 / m^d$.
\end{lemma}
The proof of \eqref{rateBetaHatFixedB1} follows from the union bound $P(\mathcal{E}_2 \cap \mathcal{E}_3) \geq 1 - P(\mathcal{E}_2) - P(\mathcal{E}_3) \geq 1 - 4 / m^d$. Next we prove \eqref{rateGammaHatFixedDelta}. Let $r_{n, m} = s_{n, m} + t_{n, m}$, as defined in \eqref{rateBetaHatFixedB1}.  Let $\delta = (1, -1) \in \mathbb{R}^2$.
Then
\[
	| \widehat{\gamma}_j(\BinvFixed) - \gamma_j | = \left | \delta^T \left( \widehat{\beta}_j(\BinvFixed) - \beta_j^* \right) \right | \leq \lVert \delta \rVert_2 \lVert \widehat{\beta}_j(\BinvFixed) - \beta_j^* \rVert_2 = \sqrt{2} \lVert \widehat{\beta}_j(\BinvFixed) - \beta_j^* \rVert_2,
\]
where we used the Cauchy-Schwarz inequality.  Hence if $\lVert \widehat{\beta}_j(\BinvFixed) - \beta_j \rVert_2 \leq r_{n, m}$, it follows that $| \widehat{\gamma}_j(\BinvFixed) - \gamma_j | \leq \sqrt{2} r_{n, m}$.  The result holds by applying a union bound over the variables $j = 1, \ldots, m$.
\end{proofof2}
This completes the proof of Theorem \ref{thm::GLSFixedB}.

\subsection{Proof of Corollary \ref{theoremInference} and Corollary \ref{corDesignEffectAlg2}}
\label{proofTheoremInference}

First note that by Proposition~\ref{L2BoundsOmegaTwoGroup},
\begin{align}
\abs{\delta^T (D^T \widehat{B}^{-1} D)^{-1} \delta  - \delta^T (D^T B^{-1} D)^{-1} \delta} &= \abs{\delta^T \Parens{(D^T \widehat{B}^{-1} D)^{-1} - (D^T B^{-1} D)^{-1}} \delta} \notag \\
&\leq \twonorm{\delta}^2 \twonorm{(D^T \widehat{B}^{-1} D)^{-1} - (D^T B^{-1} D)^{-1}} \notag \\
&= 2 \twonorm{(D^T \widehat{B}^{-1} D)^{-1} - (D^T B^{-1} D)^{-1}} \notag \\
&\leq 2 \frac{\twonorm{B}^2 \twonorm{\Delta} }{n_{\min}}. \label{numeratorCor2}
\end{align}
Note that by Proposition~\ref{L2boundInverseOfDtransBinvD},
\begin{equation} \label{denomCor2}
|\delta^T \Omega \delta| \geq \frac{\lambda_{\min}(B)}{n_{\max}}.
\end{equation}
Corollary \ref{theoremInference} follows from \eqref{numeratorCor2} and \eqref{denomCor2}, which provide an upper bound on the numerator and lower bound on the denominator, respectively.  

Corollary \ref{corDesignEffectAlg2} holds because by \eqref{mainThmErrorCov} of Theorem \ref{mainTheoremModSel},
\begin{equation}
\abs{\delta^T \Parens{\widehat{\Omega} - \Omega} \delta} \leq 2 \frac{\twonorm{B}^2 }{n_{\min}} \Parens{ \frac{C' \lambda_A \sqrt{\abs{B^{-1}}_{0, \operatorname{off}} \vee 1 } }{b_{\min} \varphi_{\min}^2(\rho(B))} } \leq 2 C' \frac{\kappa(B)}{n_{\min}} \lambda_A \sqrt{\abs{B^{-1}}_{0, \operatorname{off}} \vee 1}
\end{equation}

\subsection{Proof of Lemma~\ref{HansonWrightBetaBinvBetaStar}}
\label{sec::rateBetaHatBinvBetaStar}
\begin{proofof2}
First, we show that
\begin{equation} \label{snmBound}
  \lVert \Omega^{1/2} \rVert_F + d^{1/2} K^2 \sqrt{\log(m)} \lVert \Omega \rVert_2^{1/2} / \sqrt{c} \leq s_{n, m},
\end{equation}
with $s_{n, m}$ as defined in \eqref{def:stFixedDelta}.
Because $\lVert \Omega^{1/2} \rVert_F \leq \sqrt{2} \lVert \Omega^{1/2} \rVert_2$, it follows that
\bens
\lVert \Omega^{1/2} \rVert_F + d^{1/2} K^2 \sqrt{\log(m)} \lVert \Omega \rVert_2^{1/2} / \sqrt{c}
&\leq & \left(\sqrt{2} + d^{1/2} K^2 \sqrt{\log(m)} / \sqrt{c} \right) \lVert \Omega \rVert_2^{1/2} \\
&\leq & C_3 d^{1/2} \sqrt{\log(m)} \lVert \Omega \rVert_2^{1/2} \leq C_3 d^{1/2} \sqrt{ \frac{\log(m) \lVert B \rVert_2}{n_{\min}}},
\eens
where the last step follows from Proposition
\ref{L2boundInverseOfDtransBinvD}. Next, we express $\widehat{\beta}_j(B^{-1})  - \beta_j^*$ as
\bens
  \widehat{\beta}_j(B^{-1})  - \beta_j^* = \Omega^{1/2} \eta_j,
  \quad\text{where } \quad
\eta_j = \Omega^{-1/2} \left(\widehat{\beta}_j(B^{-1}) - \beta_j^*
\right).
\eens
By the bound (\ref{snmBound}), event $\mathcal{E}_2^c$ implies $\{  \lVert \Omega^{1/2} \eta_j \rVert_2 > \lVert \Omega^{1/2} \rVert_F + d^{1/2} K^2 \sqrt{\log(m)} \lVert \Omega \rVert_2^{1/2} / \sqrt{c} \}$.
Therefore,
\bens
P\left( \lVert \Omega \eta_j \rVert_2 \geq s_{n, m} \right)
&\leq & P\left(  \lVert \Omega \eta_j \rVert_2 >  \lVert \Omega^{1/2} \rVert_F + d^{1/2} K^2 \sqrt{\log(m)} \lVert \Omega \rVert_2^{1/2} / \sqrt{c} \right) \\
&\leq & P\left( \left | \lVert \Omega^{1/2} \eta_j \rVert_2 -  \lVert \Omega^{1/2} \rVert_F \right|  > d^{1/2} K^2 \sqrt{\log(m)} \lVert \Omega \rVert_2^{1/2} / \sqrt{c} \right) \\
&\leq  & 2 \exp\left( \frac{-c \left( d^{1/2} K^2 \sqrt{\log(m)} \lVert \Omega \rVert_2^{1/2} / \sqrt{c} \right)^2}{K^4 \lVert \Omega^{1/2} \rVert_2^2 } \right) \\
&= & 2\exp\left( \frac{- d \log(m) \lVert \Omega \rVert_2}{ \lVert \Omega^{1/2} \rVert_2^2} \right)
=  2\exp\left( - d \log(m) \right) = 2/m^d.
\eens
\end{proofof2}

\subsection{Proof of Lemma~\ref{BetaBtildeBetaBStarHansonWright}}
\label{sec::rateBetaHatBHatinvBetaHatBinv}

\begin{proofof2}
The proof will proceed in the following steps.  First, we show that $\widehat{\beta}_j(\BinvFixed) - \widehat{\beta}_j(B^{-1})$ can be expressed as $V Z_j$, where
\[
    V = \left( \OmegaFixed D^T \BFixed^{-1} - \Omega D^T B^{-1} \right)B^{1/2} \in \mathbb{R}^{2 \times m}
\]
is a fixed matrix, and $Z_j = B^{-1/2}X_j$.  Second, we show that
\[
  \lVert V \rVert_F + d^{1/2} K^2 \log^{1/2}(m) \lVert V \rVert_2 / \sqrt{c} \leq \widetilde{C} n_{\min}^{-1/2} \lVert \Delta \rVert_2.
\]
Third, we use the first and second steps combined with the Hanson-Wright inequality to show that with high probability, $\lVert VZ_j \rVert_2$ is at most $\widetilde{C} n_{\min}^{-1/2} \lVert \Delta \rVert_2$.

For the first step of the proof, let $Z_j = B^{-1/2} X_j$, and note that $\widehat{\beta}_j(\BinvFixed) - \widehat{\beta}_j(B^{-1}) = VZ_j$, where $V \in \mathbb{R}^{2 \times m}$ is a fixed matrix, because
\begin{align*}
  \widehat{\beta}_j(\BFixed^{-1}) - \widehat{\beta}_j(B^{-1}) &= \left[ (D^T \BFixed^{-1} D)^{-1} D^T \BFixed^{-1} - \Omega D^T B^{-1} \right] B^{1/2} (B^{-1/2} X_j) \\
  &= \left[ (D^T \BFixed^{-1} D)^{-1} D^T \BFixed^{-1} - \Omega D^T B^{-1} \right] B^{1/2} Z_j.
\end{align*}

For the second step of the proof, we show that $\lVert V \rVert_F + d^{1/2} K^2 \log^{1/2}(m) \lVert V \rVert_2 / \sqrt{c} \leq \widetilde{C} n_{\min}^{-1/2} \lVert \Delta \rVert_2$.  First we obtain an upper bound on $V$.  By the triangle inequality,
\bens
\lefteqn{
\lVert \OmegaFixed D^T \BFixed^{-1} - \Omega D^T B^{-1} \rVert_2
= \left \lVert  \OmegaFixed D^T \BFixed^{-1} - \Omega D^T B^{-1}
\right \rVert_2 } \\
&\leq & \left \lVert \left(\OmegaFixed - \Omega \right) D^T
  (\BFixed^{-1} - B^{-1}) \right \rVert_2
+ \left \lVert \left(\OmegaFixed - \Omega \right) D^T B^{-1} \right \rVert_2 + \left \lVert \Omega D^T \Delta \right \rVert_2 \\
&= & \delta_2(\Delta) + \delta_3(\Delta) + \delta_1(\Delta).
\eens
We bound each of the three terms using Proposition \ref{L2BoundDTBiHatDinvMinusDTBiDInv},
\begin{align*}
\delta_2(\Delta)
  &= \left \lVert  \left(\OmegaFixed - \Omega \right) D^T \Delta \right \rVert_2 \leq \frac{\sqrt{n_\text{ratio}}}{\sqrt{n_{\min}}} \frac{\lVert B \rVert_2^2 \lVert \Delta \rVert_2^2}{1/n_{\text{ratio}} -  \lVert B \rVert_2 \lVert \Delta \rVert_2} \\
\delta_3(\Delta)
  &= \left \lVert \left(\OmegaFixed - \Omega \right) D^T B^{-1} \right \rVert_2 \leq \frac{\sqrt{n_\text{ratio}}}{\sqrt{n_{\min}}} \frac{\lVert B \rVert_2^2 \lVert B^{-1} \rVert_2 \lVert \Delta \rVert_2}{1/n_{\text{ratio}} -  \lVert B \rVert_2 \rVert_2 \lVert \Delta \rVert_2} \\
\delta_1(\Delta)
  &= \left \lVert \Omega D^T \Delta  \right \rVert_2 \leq \frac{\sqrt{n_\text{ratio}}}{\sqrt{n_{\min}}} \lVert B \rVert_2 \lVert \Delta \rVert_2.
\end{align*}
Applying the above bounds yields
\bens
 \lVert V \rVert_2 &\leq &
\frac{\sqrt{n_\text{ratio}}}{\sqrt{n_{\min}}} \lVert \Delta \rVert_2 \lVert B \rVert_2^{1/2}
\left( \frac{\lVert B \rVert_2^2 \lVert \Delta \rVert_2}{1/ \kappa^2(D) -  \lVert B \rVert_2 \lVert \Delta \rVert_2}
+ \frac{\lVert B \rVert_2^2 \lVert B^{-1} \rVert_2}{1/ \kappa^2(D) -  \lVert B \rVert_2 \rVert_2 \lVert \Delta \rVert_2}
+  \lVert B \rVert_2 \right) \\
& \leq & \widetilde{C} n_{\min}^{-1/2} \lVert \Delta \rVert_2.
\eens
For the third step of the proof, we use the Hanson-Wright inequality
to bound $\lVert V Z_j \rVert_2$:
\bens
\lefteqn{
 P\left( \lVert V Z_j \rVert_2 > \widetilde{C} n_{\min}^{-1/2} \lVert \Delta \rVert_2 \right)
\leq  P\left( \lVert V Z_j \rVert_2 > \lVert V \rVert_F + d^{1/2} K^2
  \log^{1/2}(m) \lVert V \rVert_2 / \sqrt{c}  \right) }
\\ &= & P\left( \lVert V Z_j \rVert_2 - \lVert V \rVert_F > d^{1/2} K^2 \log^{1/2}(m) \lVert V \rVert_2 / \sqrt{c} \right) \\
  &\leq & P\left( \left | \lVert V Z_j \rVert_2 - \lVert V \rVert_F \right | > d^{1/2} K^2 \log^{1/2}(m) \lVert V \rVert_2 / \sqrt{c} \right) \\
  &\leq & 2\exp\left(- \frac{c\left( d^{1/2} K^2 \log^{1/2}(m) \lVert V \rVert_2 / \sqrt{c}  \right)^2}{K^4 \lVert V \rVert_2^2} \right) \qquad \text{(Hanson-Wright inequality)} \\
  &= & 2 \exp\left( -d \log(m) \right) = 2 / m^d.
\eens
\end{proofof2}

\subsection{Proof of Proposition~\ref{L2boundInverseOfDtransBinvD}}
\label{sec::proofofBinvD}
\begin{proofof2}
Let $D = U \Psi V^T$ be the singular value decomposition of $D$, with
$U \in \mathbb{R}^{n \times 2}$, $\Psi \in \mathbb{R}^{2 \times 2}$,
and $V \in \mathbb{R}^{2 \times 2}$.  Then
 $(D^T B^{-1} D)^{-1} = (V \Psi U^T B^{-1} U \Psi V^T)^{-1} = V \Psi^{-1} (U^T B^{-1} U)^{-1} \Psi^{-1} V^T.$
Thus
\begin{align*}
  \lVert (D^T B^{-1} D)^{-1} \rVert_2 &= \lVert \Psi^{-1} (U^T B^{-1} U)^{-1} \Psi^{-1} \rVert_2 \qquad \text{(because $V$ is square, orthonormal)} \\
  &\leq \lVert \Psi^{-1} \rVert_2^2 \lVert (U^T B^{-1} U)^{-1} \rVert_2 \qquad \text{(sub-multiplicative property)} \\
  &= \sigma_{\max}^2(\Psi^{-1}) \lVert (U^T B^{-1} U)^{-1} \rVert_2 \\
  &= \lVert (U^T B^{-1} U)^{-1} \rVert_2 / \sigma_{\min}^2(\Psi) = \lVert (U^T B^{-1} U)^{-1} \rVert_2 / \sigma_{\min}^2(D),
\end{align*}
where $\sigma_{\min}(D) = \sigma_{\min}(\Psi)$, because $\Psi$ is the
diagonal matrix of singular values of $D$.
Next, note that $  \lVert (U^T B^{-1} U)^{-1} \rVert_2  = 1 /
\varphi_{\min}(U^T B^{-1} U)$ and
\bens
\varphi_{\min}(U^T B^{-1} U)& = &\min_{\eta \in \mathbb{R}^2} \eta^T U^T B^{-1} U \eta / \eta^T \eta.
\eens
We perform the change of variables $\gamma = U \eta$, under which $\eta^T \eta = \gamma^T U^T U \gamma = \gamma^T \gamma$ (that is, $U$ preserves the length of $\eta$ because the columns of $U$ are orthonormal).  Hence
\begin{align*}
  \varphi_{\min}(U^T B^{-1} U) &= \min_{\gamma \in \text{col}(U), \gamma \neq 0} \gamma^T B^{-1} \gamma / \gamma^T \gamma \\
  &\geq \min_{\gamma \neq 0} \gamma^T B^{-1} \gamma / \gamma^T \gamma \\
  &= \varphi_{\min}(B^{-1}) = 1 / \lVert B \rVert_2.
\end{align*}
We have shown that $1 / \varphi_{\min}(U^T B^{-1} U) \leq \lVert B \rVert_2$,
which implies that
\[
  \lVert (U^T B^{-1} U)^{-1} \rVert_2 \leq \lVert B \rVert_2.
\]
Therefore
\[
  \lVert (D^T B^{-1} D)^{-1} \rVert_2 \leq \lVert B \rVert_2 / \sigma_{\min}^2(D).
\]
In the special case of the two-group design matrix,
$\sigma_{\min}^2(D) = n_{\min}$, so \\
$  \lVert (D^T B^{-1} D)^{-1} \rVert_2 \leq \lVert B \rVert_2 / n_{\min}.$

The proof of \eqref{minEvalOmega} is as follows:
\begin{align*}
\lambda_{\min}(\Omega) &= \frac{1}{\lambda_{\max}\Parens{\Omega^{-1}} } = \frac{1}{\lambda_{\max}\Parens{D^T B^{-1} D} } \geq \frac{1}{\twonorm{D}^2 \lambda_{\max}(B^{-1}) } = \frac{\lambda_{\min}(B) }{\twonorm{D}^2} = \frac{\lambda_{\min}(B) }{n_{\max}}.
\end{align*}
\end{proofof2}

\subsection{Proof of Proposition~\ref{L2BoundDTBiHatDinvMinusDTBiDInv}}
\begin{proofof2}
%{Proposition~\ref{L2BoundDTBiHatDinvMinusDTBiDInv}}
By the definitions of $\OmegaFixed$ in \eqref{def:Omega_app}
and $\Delta = \BFixed^{-1} - B^{-1}$, we have by Theorem \ref{GolubInversePerturbation}
\begin{align*}
\lVert \OmegaFixed - \Omega \rVert_2 &= \lVert (D^T \BFixed D)^{-1} - \Omega \rVert_2 \\
&= \left \lVert \left(D^T \BFixed^{-1} D- D^T B^{-1} D + D^T B^{-1} D \right)^{-1} -  \Omega \right \rVert_2 \\
&= \left \lVert \left( D^T B^{-1} D + D^T \Delta D \right)^{-1} -  \Omega \right \rVert_2 \\
&\leq \frac{\lVert D^T \Delta D \rVert_2 \lVert \Omega \rVert_2^2}{1 - \lVert  \Omega \rVert_2 \lVert D^T \Delta D \rVert_2} \qquad \text{(by Theorem \ref{GolubInversePerturbation})} \\
% &\leq \frac{\left( \sigma_{\max}^2(D) \lVert \Delta \rVert_2 \right) \left( \lVert B \rVert_2^2 / \sigma_{\min}^4(D) \right) }{1 - \left(\lVert B \rVert_2 / \sigma_{\min}^2(D) \right) \left( \sigma_{\max}^2(D) \lVert \Delta \rVert_2 \right) } \\
&\leq
% numerator
\frac{\left( \sigma_{\max}^2(D) / \sigma_{\min}^4(D) \right) \lVert B \rVert_2^2 \lVert \Delta \rVert_2  ) }
% denom
{1 - \kappa^2(D) \lVert B \rVert_2  \lVert \Delta \rVert_2 }.
\end{align*}
In the last step we apply Proposition \ref{L2boundInverseOfDtransBinvD}. Thus
\bens
  \lVert \OmegaFixed - \Omega \rVert_2 &\leq &
 \frac{1}{\sigma_{\min}^2(D)} \frac{\kappa^2(D) \lVert B \rVert_2^2
   \lVert \Delta \rVert_2}
{1 - \kappa^2(D) \lVert B \rVert_2 \lVert \Delta \rVert_2} \\
& = & \frac{1}{\sigma_{\min}^2(D)} \frac{\lVert B \rVert_2^2 \lVert \Delta \rVert_2}{(1/ \kappa^2(D)) -  \lVert B \rVert_2 \lVert \Delta \rVert_2}.
\eens
We prove \eqref{def:delta1} using the submultiplicative property of the operator norm and Proposition \ref{L2boundInverseOfDtransBinvD}:
\bens
\left \lVert \Omega D^T \Delta  \right \rVert_2
\leq \frac{\lVert B \rVert_2 }{\sigma_{\min}^2(D)} \sigma_{\max}(D)
\lVert \Delta \rVert_2
= \frac{\kappa(D)}{\sigma_{\min}(D)} \lVert B \rVert_2 \lVert \Delta \rVert_2.
\eens

We prove (\ref{L2BoundOmegaTildeMinusOmegaDTDelta}), as follows:
\begin{align*}
\left \lVert  \left(\OmegaFixed - \Omega \right) D^T \Delta \right \rVert_2 &\leq \left \lVert \OmegaFixed - \Omega \right \rVert_2 \left \lVert D^T \right \rVert_2 \left \lVert  \Delta \right \rVert_2 \\
&\leq  \left[ \frac{1}{\sigma_{\min}^2(D)} \frac{\lVert B \rVert_2^2 \lVert \Delta \rVert_2}{\left( 1/ \kappa^2(D) \right) -  \lVert B \rVert_2 \lVert \Delta \rVert_2} \right] \sigma_{\max}(D) \lVert \Delta \rVert_2 \qquad \text{(by Proposition \ref{L2BoundDTBiHatDinvMinusDTBiDInv})} \\
&= \frac{\kappa(D)}{\sigma_{\min}(D)} \frac{\lVert B \rVert_2^2 \lVert \Delta \rVert_2^2}{\left( 1/\kappa^2(D) \right) -  \lVert B \rVert_2 \lVert \Delta \rVert_2}.
\end{align*}
The proof of (\ref{L2BoundOmegaTildeMinusOmegaDTBinv}) is analogous.
\end{proofof2}

%!TEX root = submit.arxiv.tex

\section{Proof of Theorem \ref{mainTheoremGroupCentering}}
\label{sec::ProofMainThmPartI}
Note that the proof in the current Section follows exactly the same steps as the proof of Theorems 3.1 and 3.2 in \citet{Zhou14a}.
Theorem \ref{mainTheoremGroupCentering} {\bf Part II} is proved in Section~\ref{sec::ProofMainThmPartII}.
To prove Theorem \ref{mainTheoremGroupCentering} {\bf Part I}, we first
state Lemma~\ref{boundCovOpFro}, which establishes rates of convergence for estimating $A^{-1}$ and $B^{-1}$
in the operator and the Frobenius norm. We then state the auxiliary
Lemma~\ref{lemma:absoluteErrorKroneckerProduct}, which is
identical to that for Theorems 11.1 and 11.2 of~\cite{Zhou14a},
except that we plug in $\widetilde{\alpha}$ and $\widetilde{\eta}$ as
defined in~\eqref{entrywiseRateBcorr}.
Putting these results together proves Theorem~\ref{mainTheoremGroupCentering}, {\bf Part I.}
We prove these auxiliary results in Section~\ref{sec::proofsforTheorem2}.

Let $\mathcal{X}_0$ denote the event
\begin{align}
  &\forall i, j \qquad \left| \frac{(e_i - p_i)^T XX^T (e_j - p_j)}{\text{tr}(A^*) \sqrt{b_{ii}^* b_{jj}^*}} - \rho_{ij}(B)  \right| \leq \widetilde{\alpha}
    \label{eq:X_0_B} \\
  &\forall i, j \qquad \left| \frac{X_i^T (I - P_2)X_j}{\text{tr}(B^*) \sqrt{a_{ii}^* a_{jj}^*}} - \rho_{ij}(A)  \right| \leq \widetilde{\eta},
    \label{eq:X_0_A}
\end{align}
with $\mathcal{X}_0(B)$ and $\mathcal{X}_0(A)$ denoting the events defined by equations~\eqref{eq:X_0_B} and~\eqref{eq:X_0_A}, respectively.

Let $\widetilde{\alpha}$ and $\widetilde{\eta}$ be as defined in \eqref{entrywiseRateBcorr}.
On event $\mathcal{X}_0(A)$,  for all $j$,
$\hat{\Gamma}_{jj}(A)  = \rho_{jj}(A) = 1$ and
\ben
\label{eq::delta-nf}
\max_{j,k, j\not=k}|\hat{\Gamma}_{jk}(A) - \rho_{jk}(A)| \leq \frac{2\widetilde{\eta}}{1 - \widetilde{\eta}}
\een
On event $\mathcal{X}_0(B)$, for all $j$, $\hat{\Gamma}_{jj}(B)  = \rho_{jj}(B) = 1$ and
\ben
\label{eq::delta-nm}
%\text{and } \quad
\max_{j,k, j\not=k}|\hat{\Gamma}_{jk}(B) - \rho_{jk}(B)| \leq  \frac{2\widetilde{\alpha}}{1 - \widetilde{\alpha}}.
\een
\begin{lemma} \label{boundCovOpFro}
Suppose (A1) and (A2) hold.  Let $\widehat{W}_1$
and $\widehat{W}_2$ be as defined in \eqref{W1hatW2hat}.  Let
$\widehat{A}_{\rho}$ and $\widehat{B}_{\rho}$ be as defined in
(\ref{geminiObjectiveFnA}) and (\ref{geminiObjectiveFnB}).  For some
absolute constants $18 < C, C' < 36$,
the following events hold with probability at least $1 - 2/(n \vee m)^2$,
\begin{align}
\delta_{A, 2} &:= \lVert \widehat{W}_1 \widehat{A}_\rho \widehat{W}_1
/ \operatorname{tr}(B) -
A \rVert_2 \leq C a_{\max} \kappa(\rho(A))^2 \lambda_B \sqrt{|A^{-1}|_{0, \text{off}} \vee 1} \\
\delta_{B, 2} &:= \lVert \widehat{W}_2 \widehat{B}_\rho \widehat{W}_2
/ \operatorname{tr}(A) -
B \rVert_2 \leq C' b_{\max} \kappa(\rho(B))^2 \lambda_A \sqrt{|B^{-1}|_{0, \text{off}} \vee 1} \\
\delta_{A, F} &:= \lVert \widehat{W}_1 \widehat{A}_\rho \widehat{W}_1
/ \operatorname{tr}(B) -
A \rVert_F \leq C a_{\max} \kappa(\rho(A))^2 \lambda_B \sqrt{|A^{-1}|_{0, \text{off}} \vee m} \\
\delta_{B, F} &:= \lVert \widehat{W}_2 \widehat{B}_\rho \widehat{W}_2
/ \operatorname{tr}(A) -
B \rVert_F \leq C' b_{\max} \kappa(\rho(B))^2 \lambda_A \sqrt{\offzero{B^{-1}} \vee n};
\end{align}
and for some $10 < C, C' < 19$,
\bens
\delta_{A, 2}^- &:= & \left \lVert \operatorname{tr}(B) \left( \widehat{W}_1 \widehat{A}_\rho \widehat{W}_1 \right)^{-1} - A^{-1} \right \rVert_2 \leq \frac{C \lambda_B \sqrt{\offzero{A^{-1}} \vee 1}}{a_{\min} \varphi^2_{\min}(\rho(A))} \label{rateCovAInvOp} \\
\delta_{B, 2}^- &:= & \left \lVert \operatorname{tr}(A) \left( \widehat{W}_2 \widehat{B}_\rho \widehat{W}_2\right)^{-1}  - B^{-1} \right \rVert_2 \leq \frac{C' \lambda_A \sqrt{|B^{-1}|_{0, \text{off}} \vee 1}}{b_{\min} \varphi^2_{\min}(\rho(B))} \label{rateCovBInvOp} \\
\delta_{A, F}^- &:= & \left \lVert \operatorname{tr}(B) \left( \widehat{W}_1 \widehat{A}_\rho \widehat{W}_1 \right)^{-1}  - A^{-1} \right \rVert_F \leq \frac{C \lambda_B \sqrt{|A^{-1}|_{0, \text{off}} \vee m}}{a_{\min} \varphi^2_{\min}(\rho(A))} \\
\delta_{B, F}^- &:= & \left \lVert \operatorname{tr}(A)
\left( \widehat{W}_2 \widehat{B}_\rho \widehat{W}_2 \right)^{-1} -
B^{-1} \right \rVert_F \leq \frac{C' \lambda_A \sqrt{|B^{-1}|_{0, \text{off}} \vee n}}{b_{\min} \varphi^2_{\min}(\rho(B))}.
\eens
\end{lemma}
Lemma~\ref{lemma:absoluteErrorKroneckerProduct} follows from
Theorems 11.1 and 11.2 of~\cite{Zhou14a,Zhou14supp},
where we now plug in $\widetilde{\alpha}$ and $\widetilde{\eta}$ as
defined in~\eqref{entrywiseRateBcorr}. For completeness, we provide a
sketch in Section~\ref{sec::proof-intermediate-Kron-prod}.

\begin{lemma} \label{lemma:absoluteErrorKroneckerProduct}
Suppose (A1) and (A2) hold.  For $\varepsilon_1, \varepsilon_2 \in (0, 1)$, let
\[
	\lambda_A = \widetilde{\eta} / \varepsilon_1, \quad \lambda_B = \widetilde{\alpha} / \varepsilon_2,
\]
for $\widetilde{\alpha}$, $\widetilde{\eta}$ as defined in
(\ref{entrywiseRateBcorr}),
% and (\ref{entrywiseRateAcorr}),
and suppose $\lambda_A, \lambda_B < 1$.
Then on event $\mathcal{X}_0$, for $18 < C, C' < 36$,
\bens
&& \lVert \widehat{A \otimes B} - A \otimes B \rVert_2 \leq
\frac{\lambda_A \wedge \lambda_B}{2}
\lVert A \rVert_2 \lVert B \rVert_2 + C \lambda_B a_{\max} \lVert B \rVert_2 \kappa(\rho(A))^2  \sqrt{|A^{-1}|_{0, \text{off}} \vee 1} \\
	& &  + C' \lambda_A b_{\max} \lVert A \rVert_2
        \kappa(\rho(B))^2 \sqrt{|B^{-1}|_{0, \text{off}} \vee 1} \\
&& + 2\left[ C' \lambda_A b_{\max} \kappa(\rho(B))^2 \sqrt{|B^{-1}|_{0, \text{off}} \vee 1} \right] \left[ C \lambda_B a_{\max} \kappa(\rho(A))^2 \sqrt{|A^{-1}|_{0, \text{off}} \vee 1} \right],
\eens
and for $10 < C, C' < 19$,
\begin{align*}
	&\lVert \widehat{A \otimes B}^{-1} - A^{-1} \otimes B^{-1} \rVert_2 \leq \frac{\lambda_A \wedge \lambda_B}{3} \lVert A^{-1} \rVert_2 \lVert B^{-1} \rVert_2 + C \lambda_B \lVert B^{-1} \rVert_2 \frac{\sqrt{|A^{-1}|_{0, \text{off}} \vee 1}}{a_{\min} \varphi^2_{\min}(\rho(A))} \\
	&\quad + C' \lambda_A \lVert A^{-1} \rVert_2 \frac{\sqrt{|B^{-1}|_{0, \text{off}} \vee 1}}{b_{\min} \varphi^2_{\min}(\rho(B))} + \frac{3}{2} \left[C \lambda_B \frac{\sqrt{|A^{-1}|_{0, \text{off}} \vee 1}}{a_{\min} \varphi^2_{\min}(\rho(A))} \right] \left[ C' \lambda_A \frac{\sqrt{|B^{-1}|_{0, \text{off}} \vee 1}}{b_{\min} \varphi^2_{\min}(\rho(B))} \right];
\end{align*}
For $18 < C, C' < 36$,
\bens
& &\lVert \widehat{A \otimes B} - A \otimes B \rVert_F \leq \frac{\lambda_A \wedge \lambda_B}{2} \lVert A \rVert_F \lVert B \rVert_F + C \lambda_B a_{\max} \lVert B \rVert_F \kappa(\rho(A))^2  \sqrt{|A^{-1}|_{0, \text{off}} \vee m} \\
& &  + C' \lambda_A b_{\max} \lVert A \rVert_F \kappa(\rho(B))^2 \sqrt{|B^{-1}|_{0, \text{off}} \vee n} \\
&& + 2\left[ C' \lambda_A b_{\max} \kappa(\rho(B))^2 \sqrt{|B^{-1}|_{0, \text{off}} \vee n} \right] \left[ C \lambda_B a_{\max} \kappa(\rho(A))^2 \sqrt{|A^{-1}|_{0, \text{off}} \vee m} \right],
\eens
and for $10 < C, C' < 19$,
\begin{align*}
	&\lVert \widehat{A \otimes B}^{-1} - A^{-1} \otimes B^{-1} \rVert_F \leq \frac{\lambda_A \wedge \lambda_B}{3} \lVert A^{-1} \rVert_2 \lVert B^{-1} \rVert_F + C \lambda_B \lVert B^{-1} \rVert_F \frac{\sqrt{|A^{-1}|_{0, \text{off}} \vee m}}{a_{\min} \varphi^2_{\min}(\rho(A))} \\
	&\quad + C' \lambda_A \lVert A^{-1} \rVert_F \frac{\sqrt{|B^{-1}|_{0, \text{off}} \vee n}}{b_{\min} \varphi^2_{\min}(\rho(B))} + \frac{7}{5} \left[C \lambda_B \frac{\sqrt{|A^{-1}|_{0, \text{off}} \vee m}}{a_{\min} \varphi^2_{\min}(\rho(A))} \right] \left[ C' \lambda_A \frac{\sqrt{|B^{-1}|_{0, \text{off}} \vee n}}{b_{\min} \varphi^2_{\min}(\rho(B))} \right].
\end{align*}
\end{lemma}

\subsection{Proof of Theorem \ref{mainTheoremGroupCentering}, Part I}
\label{app::proofKroneckerProduct_details}

We state additional helpful bounds:
\begin{align}
	&(a_{\min} \vee \varphi_{\min}(A)) \sqrt{m} \leq \lVert A
        \rVert_F = \left( \sum_{i = 1}^m \varphi_i^2(A) \right)^{1/2}
        \leq \sqrt{m} \lVert A \rVert_2,
\label{boundsAFro} \\
	&(b_{\min} \vee \varphi_{\min}(B)) \sqrt{n} \leq \lVert B
        \rVert_F = \left( \sum_{i = 1}^m \varphi_i^2(B) \right)^{1/2}
        \leq \sqrt{n} \lVert B \rVert_2,
\label{boundsBFro} \\
	& \sqrt{m} / a_{\max} = \left( \frac{1}{a_{\max}} \vee
          \frac{1}{\varphi_{\max}(A)} \right) \sqrt{m} \leq \lVert
        A^{-1} \rVert_F \leq \sqrt{m} \lVert A^{-1} \rVert_2,
\label{boundsAinvFro}
\end{align}
and
\begin{align}
	& \sqrt{n} / b_{\max} = \left( \frac{1}{b_{\max}} \vee
          \frac{1}{\varphi_{\max}(B)} \right) \sqrt{n} \leq \lVert
        B^{-1} \rVert_F \leq \sqrt{n} \lVert B^{-1} \rVert_2.
\label{boundsBinvFro}
\end{align}

\begin{proofof}{Theorem \ref{mainTheoremGroupCentering}, {\bf Part I}}
We plug in bounds as in \eqref{boundOpNormAandB}  and \eqref{boundOpNormAandBII}
into Lemma \ref{lemma:absoluteErrorKroneckerProduct} to obtain under (A1) and (A2), $\left \lVert \widehat{A \otimes B} - A \otimes B \right \rVert_2 \leq \lVert A \rVert_2 \lVert B \rVert_2 \delta$, where
\begin{align*}
\delta &= \frac{\lambda_A \wedge \lambda_B}{2} + \frac{C r_a
  \kappa(\rho(A))}{\varphi_{\min}(\rho(A))} \lambda_B \sqrt{
  |A^{-1}|_{0, \text{off}} \vee 1}
+ \frac{C' r_b \kappa(\rho(B))}{\varphi_{\min}(\rho(B))} \lambda_A \sqrt{  |B^{-1}|_{0, \text{off}} \vee 1} \\
&+ 2\left[ \frac{C r_a \kappa(\rho(A))}{\varphi_{\min}(\rho(A))}
  \lambda_B \sqrt{  |A^{-1}|_{0, \text{off}} \vee 1} \right]
\left[ \frac{C' r_b \kappa(\rho(B))}{\varphi_{\min}(\rho(B))} \lambda_A \sqrt{  |B^{-1}|_{0, \text{off}} \vee 1} \right] \\
&= \frac{\lambda_A \wedge \lambda_B}{2} + \log^{1/2}(m \vee n)\left(
  \sqrt{\frac{ |A^{-1}|_{0, \text{off}} \vee 1}{m}}+ \sqrt{\frac{ |B^{-1}|_{0, \text{off}} \vee 1}{n}} \right) + o(1).
\end{align*}
For the inverse, we plug in bounds as in \eqref{ReciprocalMinEvalCorrA} and \eqref{ReciprocalMinEvalCorrB} into Lemma \ref{lemma:absoluteErrorKroneckerProduct} to obtain under (A1) and (A2), $\left \lVert \widehat{A \otimes B}^{-1} - A^{-1} \otimes B^{-1} \right \rVert_2 \leq \lVert A^{-1} \rVert_2 \lVert B^{-1} \rVert_2 \delta'$, where
\begin{align*}
\delta' &= \frac{\lambda_A \wedge \lambda_B}{3} + \frac{C r_a
  \lambda_B \sqrt{  |A^{-1}|_{0, \text{off}} \vee
    1}}{\varphi_{\min}(\rho(A))}
+ \frac{C' r_b \lambda_A \sqrt{  |B^{-1}|_{0, \text{off}} \vee 1}}{\varphi_{\min}(\rho(B))} \\
&+ \frac{3}{2} \left[ \frac{C r_a \lambda_B \sqrt{  |A^{-1}|_{0,
        \text{off}} \vee 1}}{\varphi_{\min}(\rho(A))} \right]
\left[ \frac{C' r_b \lambda_A \sqrt{  |B^{-1}|_{0, \text{off}} \vee 1}}{\varphi_{\min}(\rho(B))} \right] \\
&\asymp \frac{\lambda_A \wedge \lambda_B}{3} +
\log^{1/2}(m \vee n) \left( \sqrt{\frac{ |A^{-1}|_{0, \text{off}} \vee 1}{m}} + \sqrt{\frac{ |B^{-1}|_{0, \text{off}} \vee 1}{n}} \right) + o(1).
\end{align*}
The bounds in the Frobenius norm are proved in a similar manner; see~\cite{Zhou14a} to finish.
\end{proofof}

\subsection{Proof of Theorem \ref{mainTheoremGroupCentering}, Part II}
\label{sec::ProofMainThmPartII}

\begin{proofof2}

Let $\widehat{B}^{-1} = \widehat{W}_2 \widehat{B}_{\rho} \widehat{W}_2$.  Let $\widehat{\Delta} = \widehat{B}^{-1} - B^{-1}$. Let $\mathcal{E}_0(B)$ denote the event given by equations (\ref{rateCovAInvOp}) and (\ref{rateCovBInvOp}), which we know has probability at least $1 - 2/(n \vee m)^2$ from Lemma~\ref{boundCovOpFro}, and define the event
\begin{equation}
\mathcal{E}_4 = \left\{ \lVert \widehat{\beta}_j(\widehat{B}^{-1}) - \beta_j^* \rVert_2 \leq s_{n, m} + t_{n, m}' \right\},
\end{equation}
where $s_{n, m}$ is as defined in (\ref{def:stFixedDelta}) and
\begin{equation}
t_{n, m}' := C \lambda_A \sqrt{ \frac{n_{\text{ratio}} \left(|B_0^{-1}|_{0, \text{off}} \vee 1\right)}{n_{\min}}}.
\end{equation}

Under $\mathcal{E}_0(B)$, we see that
\begin{equation} \label{eq:Delta_o_const}
\lVert \widehat{\Delta} \rVert_2 \leq \frac{C' \lambda_A \sqrt{|B^{-1}|_{0, \text{off}} \vee 1}}{b_{\min} \varphi^2_{\min}(\rho(B))} = o(1).
\end{equation}
Using Proposition \ref{L2boundInverseOfDtransBinvD} and the fact that $\lVert D \rVert_2 = \sqrt{n_{\max}}$, we get that
\begin{equation} \label{eq:D_delta_D_intermediate}
  \lVert \Omega D^T \widehat{\Delta} D \rVert_2 \leq n_{\text{ratio}} \lVert B \rVert_2 \lVert \widehat{\Delta} \rVert_2,
\end{equation}
From \eqref{eq:Delta_o_const} we know that $\lVert \widehat{\Delta}
\rVert_2 \leq 1 / (n_{\text{ratio}} \lVert B \rVert_2)$,
which we can plug into \eqref{eq:D_delta_D_intermediate} to show that
$\lVert \Omega D^T \widehat{\Delta} D \rVert_2 < 1$.
This implies that $\widetilde{C} n_{\min}^{-1/2} \lVert
\widehat{\Delta} \rVert_2 \leq t_{n, m}'$.
Therefore, we can apply Theorem~\ref{thm::GLSFixedB} to get that the
conditional probability of $\mathcal{E}_4$ given $\mathcal{E}_0(B)$ is at least $1-4/(n \vee m)^2$.

We can then bound the unconditional probability,
\bens
P(\mathcal{E}_4^c)
  &\leq & P\left( \mathcal{E}_4^c \mid \mathcal{E}_0(B) \right) P\left( \mathcal{E}_0(B) \right) + P\left( \mathcal{E}_0(B)^c \right) \\
  &\leq &  P\left( \mathcal{E}_4^c \mid \mathcal{E}_0(B) \right)  + P\left( \mathcal{E}_0(B)^c \right) \\
  &\leq & \frac{4}{(n \vee m)^2}  + \frac{2}{(n \vee m)^2}.
\eens

\end{proofof2}

\section{More proofs for Theorem~\ref{mainTheoremGroupCentering}}
\label{sec::proofsforTheorem2}
The proof of Lemma~\ref{boundCovOpFro} appears in
Section~\ref{sec::proof-lemma-cov-opFro}.
The proofs of auxiliary lemmas appear in
Section~\ref{sec::proof-intermediate-Kron-prod}.      

\subsection{Proof of Lemma \ref{boundCovOpFro}} 
\label{sec::proof-lemma-cov-opFro}
In order to prove Lemma \ref{boundCovOpFro}, we need Theorem~\ref{thm::frob}, which shows explicit non-asymptotic convergence rates in the Frobenius norm for estimating $\rho(A)$, $\rho(B)$, and their
inverses.  Theorem~\ref{thm::frob} follows from the standard proof; see~\cite{RBLZ08,ZRXB11}
%is stated below and proved in Section \ref{sec:::Fro-norm-consistency-A}.  
We also need Proposition~\ref{sandwichBound} and
Lemma~\ref{boundW1W2errorOpFro}, 
which are stated below and proved in Section~\ref{sec::proof-intermediate-Kron-prod}.  

\begin{theorem}
\label{thm::frob}
Suppose that (A2) holds. 
Let $\hat{A}_{\rho}$ and $\hat{B}_{\rho}$ be the unique minimizers defined by
\eqref{geminiObjectiveFnA} and~\eqref{geminiObjectiveFnB} with
 sample correlation matrices $\hat\Gamma(A)$ and $\hat\Gamma(B)$
as their input.  

Suppose that event $\mathcal{X}_0$ holds, with  
\ben
\nonumber
& & \widetilde{\eta} \sqrt{\offzero{A^{-1}} \vee 1} = o(1)
 \quad \text{and } \quad \widetilde{\alpha} \sqrt{\offzero{B^{-1}} \vee
   1}= o(1). \\
\label{eq::lambda-choir}
& & \text{Set for some } \; 0< \e, \ve < 1, \; \;
 \lambda_{B} =
{\widetilde{\alpha} }/{\ve} \; \mbox{ and } \;
\lambda_{A}   =  {\widetilde{\eta} }/{\e}.
\een
Then on event $\mathcal{X}_0$, we have for $9< C <  18$
\begin{eqnarray}
& &
\nonumber
\twonorm{\hat{A}_{\rho} -  \rho(A)} \le
\fnorm{\hat{A}_{\rho} -  \rho(A)}
\leq C \Cond(\rho(A))^2 \lambda_{B} \sqrt{\offzero{A^{-1}}  \vee 1}, \\
\nonumber
& &
\twonorm{\hat{B}_{\rho} -  \rho(B)} \le \fnorm{\hat{B}_{\rho} -  \rho(B)} \leq
C \Cond(\rho(B))^2 \lambda_{A}\sqrt{\offzero{B^{-1}}  \vee 1},
\end{eqnarray}
and
\begin{eqnarray}
& &
\label{eq::eventAop}
\twonorm{\hat{A}_{\rho}^{-1} -  \rho(A)^{-1}}
\le \fnorm{\hat{A}_{\rho}^{-1} -  \rho(A)^{-1}}
<  \frac{C \lambda_{B} \sqrt{\offzero{A^{-1}}  \vee 1}}{2\vp^2_{\min}(\rho(A))}, \\
\label{eq::eventBop}
&  &
\twonorm{\hat{B}_{\rho}^{-1} - \rho(B)^{-1}} \le
\fnorm{\hat{B}_{\rho}^{-1} - \rho(B)^{-1}} \leq
\frac{C \lambda_{A} \sqrt{\offzero{B^{-1}}  \vee 1}}{2\vp^2_{\min}(\rho(B))}.
\end{eqnarray}
\end{theorem}

We now state an auxiliary result, Lemma \ref{boundW1W2errorOpFro}, where we prove a bound on the
error in the diagonal entries of the covariance matrices, and on their
reciprocals.  
The following Lemma provides bounds analogous to those in
Claim 15.1~\cite{Zhou14a,Zhou14supp}.
%Proposition 15.2 from~\citet{Zhou14a,Zhou14supp}.  
\begin{lemma} \label{boundW1W2errorOpFro}
Let $\widehat{W}_1$ and $\widehat{W}_2$ be as defined in
\eqref{W1hatW2hat}.  Let $W_1 = \sqrt{\operatorname{tr}(B)}
\operatorname{diag}(A)^{1/2}$ and $W_2 = \sqrt{\operatorname{tr}(A)}
\operatorname{diag}(B)^{1/2}$.  Suppose event $\mathcal{X}_0$ holds,
as defined in \eqref{eq:X_0_B}, \eqref{eq:X_0_A}.  For $\eta' :=
\frac{\widetilde{\eta}}{\sqrt{1 - \widetilde{\eta}}} \leq
\frac{\lambda_B}{6}$ and $\alpha' := \frac{\widetilde{\alpha}}{\sqrt{1
    - \widetilde{\alpha}}} \leq \frac{\lambda_A}{6}$, 
\begin{align*}
	\left \lVert \widehat{W}_1 - W_1 \right \rVert_2 &\leq \widetilde{\eta} \sqrt{\tr{B}} \sqrt{a_{\max}}, \qquad \left \lVert \widehat{W}_1^{-1} - W_1^{-1} \right \rVert_2 \leq \frac{\widetilde{\eta}}{1 - \widetilde{\eta}} / \sqrt{\tr{B}} \sqrt{a_{\min}}, \\
	\left \lVert \widehat{W}_2 - W_2 \right \rVert_2 &\leq
        \widetilde{\alpha} \sqrt{\tr{A}} 
\sqrt{b_{\max}}, \text{ and } \left \lVert \widehat{W}_2^{-1} -
  W_2^{-1} \right \rVert_2 
\leq \frac{\widetilde{\alpha}}{1 - \widetilde{\alpha}} / \sqrt{\tr{A}} \sqrt{b_{\min}}.
\end{align*}
\end{lemma}

\begin{proposition} \label{sandwichBound} \citep{Zhou14a}. 
Let $\widehat{W}$ and $W$ be diagonal positive definite matrices.
Let $\widehat{\Psi}$ and $\Psi$ be symmetric positive definite matrices.  Then
\begin{align*}
\left \lVert \widehat{W} \widehat{\Psi} \widehat{W} - W\Psi W \right \rVert_2 &\leq \left( \left \lVert \widehat{W} - W \right \rVert_2 + \lVert W \rVert_2 \right)^2 \left \lVert \widehat{\Psi} - \Psi \right \rVert_2 \\
&\qquad + \left \lVert \widehat{W} - W \right \rVert_2  \left( \left \lVert \widehat{W} - W \right \rVert_2 + 2 \right) \lVert \Psi \rVert_2 \\
\left \lVert \widehat{W} \widehat{\Psi} \widehat{W} - W\Psi W \right \rVert_F &\leq \left( \left \lVert \widehat{W} - W \right \rVert_2 + \lVert W \rVert_2 \right)^2 \left \lVert \widehat{\Psi} - \Psi \right \rVert_F \\
&\qquad + \left \lVert \widehat{W} - W \right \rVert_2  \left( \left \lVert \widehat{W} - W \right \rVert_2 + 2 \right) \lVert \Psi \rVert_F.
\end{align*}
\end{proposition}

\begin{proofof}{ Lemma \ref{boundCovOpFro}}
Assume that event $\mathcal{X}_0$ holds.  The proof follows exactly
that of Lemma 15.3 in~\cite{Zhou14a, Zhou14supp}, in view of
Theorem~\ref{thm::frob}, Lemma \ref{boundW1W2errorOpFro} and Proposition 15.2
from~\cite{Zhou14a,Zhou14supp}, which is restated immediately above in
Proposition~\ref{sandwichBound}.
\end{proofof}

It remains to prove Lemma \ref{boundW1W2errorOpFro}.

\begin{proofof}{Lemma \ref{boundW1W2errorOpFro}}
%[Proof of Lemma \ref{boundW1W2errorOpFro}]
Suppose that event $\mathcal{X}_0$ holds.  Then
\begin{equation*}
\max_{i = 1, \ldots, m} \left| \frac{\sqrt{X_i^T (I -
      P_2)X_i}}{\sqrt{a_{ii} \operatorname{tr}(B)}} - 1 \right| \leq
\left(1 - \sqrt{1 - \widetilde{\eta}} \right) 
\bigvee \left( \sqrt{1 + \widetilde{\eta}} - 1 \right) \leq \widetilde{\eta}.
\end{equation*}
Thus for all $i$,
\begin{equation*}
\frac{1}{\sqrt{1 + \widetilde{\eta}}} \leq \frac{ \sqrt{a_{ii} \operatorname{tr}(B)}}{\sqrt{X_i^T (I - P_2)X_i}} \leq \frac{1}{\sqrt{1 - \widetilde{\eta}}},
\end{equation*}
so
\begin{equation*}
	\left| \frac{ \sqrt{a_{ii} \operatorname{tr}(B)}}{\sqrt{X_i^T
              (I - P_2)X_i}} - 1 \right| \leq \left(\frac{1 - \sqrt{1
              - \widetilde{\eta}}}{\sqrt{1 - \widetilde{\eta}}}
        \right) \bigvee \left( \frac{\sqrt{1 + \widetilde{\eta}} - 1}{\sqrt{1 + \widetilde{\eta}}} \right) \leq \frac{\widetilde{\eta}}{\sqrt{1 - \widetilde{\eta}}}.
\end{equation*}
\end{proofof}

\subsection{Proof of Lemma \ref{lemma:absoluteErrorKroneckerProduct}}
\label{sec::proof-intermediate-Kron-prod}

In order to prove Lemma~\ref{lemma:absoluteErrorKroneckerProduct}, we
state Lemma \ref{intermediateKroneckerProdInTermsDelta},
Lemma~\ref{A1KronB1minusAKronB}, and
Proposition~\ref{trAtrB_estimator}.  
Let $\lVert \cdot \rVert$ denote a matrix norm such that $\lVert A
\otimes B \rVert = \lVert A \rVert \lVert B \rVert$.
  Let
\begin{align}
\Delta &:= \widehat{W}_1 \widehat{A}_{\rho} \widehat{W}_1 \otimes \widehat{W}_2 \widehat{B}_{\rho} \widehat{W}_2 / \operatorname{tr}(A) \operatorname{tr}(B) - A \otimes B, \label{DeltaKronProd} \\
\Delta' &:= \operatorname{tr}(A) \operatorname{tr}(B) \left( \widehat{W}_1 \widehat{A}_{\rho} \widehat{W}_1\right)^{-1} \otimes \left( \widehat{W}_2 \widehat{B}_{\rho} \widehat{W}_2 \right)^{-1}  - A^{-1} \otimes B^{-1}. \label{DeltaPrimeKronProd}
\end{align}
Lemma~\ref{intermediateKroneckerProdInTermsDelta} is identical to Lemma 15.5
of ~\citet{Zhou14a}, 
except that we now plug in quantities $\widetilde{\alpha}$ and
$\widetilde{\eta}$ as defined in \eqref{entrywiseRateBcorr}.
% and \eqref{entrywiseRateBcorr}.  
%The proof is identical as well, except it uses the centered data
%matrix $(I - P_2)X$.  
Likewise, Proposition \ref{trAtrB_estimator} is analogous to (20) in
Theorem 4.1 of \citet{Zhou14a}, 
except that we now  use the centered data matrix $(I - P_2) X$, together with the rates $\widetilde{\alpha}$, $\widetilde{\eta}$. 
\begin{lemma} \label{intermediateKroneckerProdInTermsDelta}
Let $\widehat{A \otimes B}$ be as in (\ref{estimatorAKroneckerB}).  Then for $\Sigma = A \otimes B$,
\begin{align}
&\left \lVert \widehat{A \otimes B}^{-1} - \Sigma^{-1} \right \rVert \leq (\widetilde{\alpha} \wedge \widetilde{\eta}) \lVert A^{-1} \rVert \lVert B^{-1} \rVert + (1 + \widetilde{\alpha} \wedge \widetilde{\eta}) \lVert \Delta' \rVert \label{intermediateBoundKronInvKnownTrace} \\
&\left \lVert \widehat{A \otimes B} - \Sigma \right \rVert \leq \frac{\lambda_A \wedge \lambda_B}{2} \lVert A \rVert \lVert B \rVert + (1 + \frac{\lambda_A \wedge \lambda_B}{2}) \lVert \Delta \rVert. \label{intermediateBoundKronKnownTrace}
\end{align}
\end{lemma}

Lemma \ref{A1KronB1minusAKronB} is a helpful bound on the difference of Kronecker products.  
\begin{lemma} \label{A1KronB1minusAKronB} ~\citep{Zhou14a}.
For matrices $A_1$ and $B_1$, let $\Delta_A := A_1 - A$ and $\Delta_B := B_1 - B$.  Then
\[
	\lVert A_1 \otimes B_1  - A \otimes B \rVert \leq \lVert \Delta_A \rVert \lVert B \rVert + \lVert \Delta_B \rVert \lVert A \rVert + \lVert \Delta_A \rVert \lVert \Delta_B \rVert.
\]
\end{lemma}

\begin{proposition} \label{trAtrB_estimator}
Under the event $\mathcal{X}_0$, as defined in as defined in \eqref{eq:X_0_B}, \eqref{eq:X_0_A}, 
\begin{equation*}
\left| \lVert (I - P_2)X \rVert_F^2 - \text{\normalfont tr}(A) \text{\normalfont tr}(B) \right| \leq (\widetilde{\alpha} \wedge \widetilde{\eta}) \text{\normalfont tr}(A) \text{\normalfont tr}(B).
\end{equation*}
\end{proposition}

\begin{proofof}{Lemma \ref{lemma:absoluteErrorKroneckerProduct}}
%[Proof of Lemma \ref{lemma:absoluteErrorKroneckerProduct}.]
Assume that event $\mathcal{X}_0$ as defined in (\ref{eq:X_0_B}),
(\ref{eq:X_0_A}) holds.  
The proof follows exactly the steps in Theorems 11.1 and 11.2 in Supplementary
Material of~\cite{Zhou14a, Zhou14supp}.
\end{proofof}

%, which is stated and proved in Section \ref{sec::proof-intermediate-Kron-prod}.  }

\begin{proofof}{Lemma \ref{intermediateKroneckerProdInTermsDelta}}
By the triangle inequality and the sub-multiplicativity of the 
norm $\lVert \cdot \rVert$, with $\Delta$ and $\Delta'$ as defined in (\ref{DeltaKronProd}) and (\ref{DeltaPrimeKronProd}),
\begin{align}
&\operatorname{tr}(A) \operatorname{tr}(B) \left \lVert \left( \widehat{W}_1^{-1} \widehat{A}_{\rho}^{-1} \widehat{W}_1^{-1} \right) \otimes \left( \widehat{W}_2^{-1} \widehat{B}_{\rho}^{-1} \widehat{W}_2^{-1} \right) \right \rVert \leq \lVert A^{-1} \rVert \lVert B^{-1} \rVert + \lVert \Delta' \rVert \label{BoundEstKroneckerInvDenomTrATrB} \\
&\left \lVert \left( \widehat{W}_1 \widehat{A}_{\rho} \widehat{W}_1 \right) \otimes \left( \widehat{W}_2 \widehat{B}_{\rho} \widehat{W}_2 \right) / \operatorname{tr}(A) \operatorname{tr}(B) \right \rVert \leq \lVert A \rVert \lVert B \rVert + \lVert \Delta \rVert.   \label{BoundEstKroneckerDenomTrATrB}
\end{align}
Following proof of Lemma 15.5~\cite{Zhou14a,Zhou14supp}, we have 
by definition of $\Delta'$, and Proposition \ref{trAtrB_estimator},  and (\ref{BoundEstKroneckerInvDenomTrATrB}),
\bens
\norm{\widehat{A \otimes B}^{-1} - A^{-1} \otimes B^{-1} }
&\leq & (\widetilde{\alpha} \wedge \widetilde{\eta}) 
\left( \lVert A^{-1} \rVert \lVert B^{-1} \rVert + \lVert \Delta' \rVert \right) + \lVert \Delta' \rVert.
\eens
By Proposition \ref{trAtrB_estimator}, we have for $\lambda_A \geq 3 \widetilde{\alpha}$, $\lambda_B \geq 3 \widetilde{\eta}$, where $\widetilde{\alpha} \wedge \widetilde{\eta} \leq \frac{\lambda_A \wedge \lambda_B}{3}$,
\begin{align}
&\left| \frac{1}{\lVert (I - P_2) X \rVert_F^2} - \frac{1}{\operatorname{tr}(A) \operatorname{tr}(B)} \right| = \left| \frac{\lVert (I - P_2) X \rVert_F^2 - \operatorname{tr}(A) \operatorname{tr}(B)}{ \lVert (I - P_2) X \rVert_F^2 \operatorname{tr}(A) \operatorname{tr}(B) }  \right| \notag\\
&\leq \left| \frac{\widetilde{\alpha} \wedge \widetilde{\eta} }{\lVert (I - P_2) X \rVert_F^2} \right| \leq \frac{\widetilde{\alpha} \wedge \widetilde{\eta}}{\operatorname{tr}(A) \operatorname{tr}(B) (1 - \widetilde{\alpha} \wedge \widetilde{\eta})} \notag \\
&\text{ thus } \left| \frac{\operatorname{tr}(A) \operatorname{tr}(B)}{\lVert (I - P_2) X \rVert_F^2} - 1 \right| \leq \frac{\widetilde{\alpha} \wedge \widetilde{\eta}}{1 - \widetilde{\alpha} \wedge \widetilde{\eta}} \leq \frac{\lambda_A \wedge \lambda_B}{2}. \label{BoundTrATrBMinusFroOverFro}
\end{align}
By the triangle inequality, the definition of $\Delta$ in
(\ref{DeltaKronProd}), 
and (\ref{BoundEstKroneckerDenomTrATrB}) and
(\ref{BoundTrATrBMinusFroOverFro}),
\bens
\left \lVert \widehat{A \otimes B} - A \otimes B \right \rVert 
& \le &  \frac{\lambda_A + \lambda_B}{2} \lVert A \rVert \lVert B
\rVert + (1 + \frac{\lambda_A + \lambda_B}{2}) \lVert \Delta \rVert;
\eens
See the proof of Lemma 15.5~\cite{Zhou14a,Zhou14supp}.
\end{proofof}

\begin{proofof}{Proposition \ref{trAtrB_estimator}}
Suppose event $\mathcal{X}_0$ holds.  Note that
\[
  E[ \lVert (I - P_2)X \rVert_F^2 ] = \text{tr}\left((I - P_2)E[XX^T](I - P_2) \right) = \text{tr}(A) \text{tr}(\widetilde{B})
\]
Decomposing by columns, we obtain the inequality,
\begin{align*}
 & \left| \lVert (I - P_2)X \rVert_F^2 - \text{tr}(A) \text{tr}(B) \right|  = \left| \sum_{j = 1}^m \lVert (I - P_2) X_j \rVert_2^2 - a_{jj} \text{tr}(B) \right| \\
  &\leq \sum_{j = 1}^m \left| X_j^T (I - P_2) X_j - a_{jj} \text{tr}(B) \right| \leq \sum_{j = 1}^m \widetilde{\eta}_{jj} a_{jj} \text{tr}(B) \leq \widetilde{\eta} \text{tr}(A) \text{tr}(B).
\end{align*}
Decomposing by rows, we obtain the inequality,
\begin{align*}
 & \left| \lVert (I - P_2)X \rVert_F^2 - \text{tr}(A) \text{tr}(B) \right| = \left| \sum_{i = 1}^n  \lVert (e_i - p_i)^T X \rVert_2^2 - b_{ii} \text{tr}(A) \right| \\
  &\leq \sum_{i = 1}^n \left| (e_i - p_i)^T XX^T (e_i - p_i) - b_{ii} \text{tr}(A) \right| \leq \sum_{i = 1}^n \widetilde{\alpha}_{ii} b_{ii} \text{tr}(A) \leq \widetilde{\alpha} \text{tr}(A) \text{tr}(B).
\end{align*}
Therefore $\left| \lVert (I - P_2)X \rVert_F^2 - \text{tr}(A) \text{tr}(B) \right| \leq (\widetilde{\alpha} \wedge \widetilde{\eta}) \text{tr}(A) \text{tr}(B)$.
\end{proofof}

\section{Entrywise convergence of sample correlations}
\label{app::entrywise_sample_corr}

In this section we prove entrywise rates of convergence for the sample
correlation matrices in Theorem \ref{thm::large-devi-cor-multiple}.  
The theorem applies to the Kronecker product model,
$\operatorname{Cov}(\operatorname{vec}(X)) = A^* \otimes B^*$, 
where for identifiability we define the sample covariance matrices as 
\begin{equation*} 
	A^* = \frac{m}{\operatorname{tr}(A)}A \quad \text{ and }  \quad B^* = \frac{\operatorname{tr}(A)}{m} B,
\end{equation*} 
with the scaling chosen so that $A^*$ has trace $m$.  Let $\rho(A) \in
\mathbb{R}^{m \times m}$ and $\rho(B) \in \mathbb{R}^{n \times n}$
denote the correlation matrices corresponding to covariance matrices
$A^*$ and $B^*$, respectively.  Assume that that the mean of $X$
satisfies the two-group model \eqref{meanMatrixTwoGroups}.  Let $P_2$
be as defined in \eqref{def:withinGroupProjection}.  
The matrix $I - P_2$ is a projection matrix of rank $n - 2$ that
performs within-group centering.  
The sample covariance matrices are defined as 
\begin{align} 
  S(B^*) &= \frac{1}{m} \sum_{j = 1}^m (I -  P_2) X_j X_j^T (I - P_2), \label{sampleCovGroupCenter} \\
  S(A^*) &= X^T(I - P_2)X / n, \label{colSampleCov}
\end{align}
where $S(B^*)$ has null space of dimension two.  
\begin{theorem}
\label{thm::large-devi-cor-multiple}
Consider a data generating random matrix as in (\ref{modelMeanCov}).
Let $C$ be some absolute constant.  Let $\widetilde{\alpha}$ and
$\widetilde{\eta}$ be as defined in \eqref{entrywiseRateBcorr}.
Let  $m \vee n \geq 2$.  Then with probability at least $1- \frac{3}{(m \vee n)^2}$,
for $\widetilde{\alpha}, \widetilde{\eta} < 1/3$, and $\hat\Gamma(A)$  and $\hat\Gamma(B)$
as in (\ref{defGammaAGammaB}),
\begin{eqnarray*}
\nonumber
& &  \forall i \not=j, \; \abs{\hat\Gamma_{ij}(B) -\rho_{ij}(B)}
\leq \frac{\widetilde{\alpha}}{1-\widetilde{\alpha}} +  \abs{\rho_{ij}(B)}
\frac{\widetilde{\alpha}}{1-\widetilde{\alpha}}  \leq 3 \widetilde{\alpha}, \\
\nonumber
& &
\forall i \not=j, \;
\abs{\hat\Gamma_{ij}(A) -\rho_{ij}(A)}
\leq  \frac{\widetilde{\eta}}{1-\widetilde{\eta}} +\abs{
 \rho_{ij}(A)}\frac{\widetilde{\eta}}{1-\widetilde{\eta}} \leq 3 \widetilde{\eta}.
\label{eq::zero-wei-bound}
\end{eqnarray*}
\end{theorem}

We state three results used in the proof of Theorem
\ref{thm::large-devi-cor-multiple}:  
Proposition \ref{rateConvergenceSampleCovToB} provides an entrywise
rate of convergence of $S(B^*)$, Proposition
\ref{rateConvergenceSampleCovToA} provides an 
entrywise rate of convergence of $S(A^*)$, and Lemma
\ref{lemma::large-devi-rep} 
states that these entrywise rates imply $\mathcal{X}_0$.  Let 
\begin{equation} 
\widetilde{B} := (I - P_2)B^*(I - P_2) = \operatorname{Cov}( (I - P_2) X_j),
\end{equation} 
where $X_j$ is the $j$th column of $X$.  Let $\widetilde{b}_{ij}$ denote the $(i, j)$th entry of $\widetilde{B}$.  

\begin{proposition} \label{rateConvergenceSampleCovToB}
Let $d > 2$.  Then with probability at least $1 - 2/m^{d - 2}$, 
\begin{equation} \label{eventB_cov}
    \forall i, j \,  \left| S_{ij}(B^*) -  b_{ij}^* \right| \leq \phi_{B, ij},
\end{equation} 
with 
\begin{equation}  \label{tailCutpointSampleCovGroupCenToB}
	\phi_{B, ij} = C \frac{\log^{1/2}(m)}{\sqrt{m}} \frac{\lVert
          A^* \rVert_F}{\sqrt{m}} 
\sqrt{\widetilde{b}_{ii} \widetilde{b}_{jj}} + \frac{3 \lVert B^* \rVert_1}{n_{\min}}.
\end{equation}

\end{proposition}

\begin{proposition}
\label{rateConvergenceSampleCovToA}
Let $d > 2$.  Then with probability at least $1 - 2/n^{d - 2}$, 
\begin{equation} \label{eventA_cov}
	\forall i, j \, \left| S_{ij}(A^*) - a_{ij}^* \tr{B^*} / n \right| > \phi_{A, ij},
\end{equation}
with
\begin{equation} \label{rateA}
\phi_{A, ij} = (a_{ij}^* / n) \left| \tr{\widetilde{B}} - \tr{B^*}
\right| +  d^{1/2} K \log^{1/2}(n \vee m) (1/n) \sqrt{a_{ij}^{*2} + a_{ii}^* a_{jj}^*} \lVert \widetilde{B} \rVert_F.
\end{equation}
\end{proposition}

\begin{lemma}
\label{lemma::large-devi-rep}
Suppose that (A2) holds and that $m \vee n \geq 2$.  
The event \eqref{eventA_cov} defined in
Proposition~\ref{rateConvergenceSampleCovToA} 
implies that $\mathcal{X}_0(A)$ holds. 
Similarly, the event \eqref{eventB_cov} defined in
Proposition~\ref{rateConvergenceSampleCovToB} 
implies $\mathcal{X}_0(B)$.  
Hence $\prob{\X_0} \geq 1- \frac{3}{(m \vee n)^2}$.
\end{lemma}

Proposition \ref{rateConvergenceSampleCovToB} is proved in section
\ref{rateConvergenceGroupCenterGemini:Bhat:proof}.
Proposition \ref{rateConvergenceSampleCovToA} is proved in section
\ref{rateConvergenceGroupCenterGemini:Ahat:proof}.
Lemma \ref{lemma::large-devi-rep} is proved in section \ref{proofSampleCorrRate}.  
Note that  Lemma \ref{lemma::large-devi-rep} follows from Propositions
\ref{rateConvergenceSampleCovToB} and
\ref{rateConvergenceSampleCovToA}.  
We now prove Theorem \ref{thm::large-devi-cor-multiple}, 
which follows from Lemma \ref{lemma::large-devi-rep}.

\begin{proofof}{Theorem \ref{thm::large-devi-cor-multiple}}
%[Proof of Theorem \ref{thm::large-devi-cor-multiple}
Let $q_i$ denote the $i$th column of $I - P_2$, so that $q_i^T XX^T q_j$ is the $(i, j)$th entry of $(I - P_2)XX^T(I - P_2)$.  Under $\mathcal{X}_0(B)$, the sample correlation $\widehat{\Gamma}(B)$ satisfies the following bound:
\bens
  \left| \widehat{\Gamma}_{ij}(B) - \rho_{ij}(B) \right| &= &
\left | \frac{q_i^T XX^T q_j}{\sqrt{q_i^T XX^T q_i} \sqrt{q_j^T XX^T q_j}} - \rho_{ij}(B) \right | \\
  &=  &\left | \frac{q_i^T XX^T q_j / \left( \text{tr}(A^*)
        \sqrt{b_{ii}^* b_{jj}^* } \right)}{\sqrt{q_i^T XX^T q_i /
        \left( b_{ii}^* \text{tr}(A^*) \right)} \sqrt{q_j^T XX^T q_j /
        \left( b_{jj}^* \text{tr}(A^*) \right)} } - \rho_{ij}(B)
  \right |  \\
&\le  &
\left | \frac{q_i^T XX^T q_j / \left( \text{tr}(A^*) \sqrt{b_{ii}^* b_{jj}^* } \right) - \rho_{ij}(B)}{\sqrt{q_i^T XX^T q_i / \left( b_{ii}^* \text{tr}(A^*) \right)} \sqrt{q_j^T XX^T q_j / \left( b_{jj}^* \text{tr}(A^*) \right)} } \right | \\
  &+ &  \left | \frac{ \rho_{ij}(B)}{\sqrt{q_i^T XX^T q_i / \left( b_{ii}^* \text{tr}(A^*) \right)} \sqrt{q_j^T XX^T q_j / \left( b_{jj}^* \text{tr}(A^*) \right)} } - \rho_{ij}(B) \right | \\
  &\leq &  \frac{\widetilde{\alpha} }{1 - \widetilde{\alpha} } + |\rho_{ij}(B)| \left| \frac{1}{1 - \widetilde{\alpha}} - 1 \right| \\
  &\leq & 3 \widetilde{\alpha},
\eens
where the first inequality holds by $\mathcal{X}_0(B)$ and the second inequality holds for $\widetilde{\alpha} \leq 1/3$.
Similarly, under $\mathcal{X}_0(A)$ we obtain an entrywise bound on the sample correlation $\widehat{\Gamma}(A)$:
\bens
  \left| \widehat{\Gamma}_{ij}(A) - \rho_{ij}(A) \right| 
&= & \left | \frac{X_i^T (I - P_2) X_j}{\sqrt{X_i^T (I - P_2) X_i} \sqrt{X_j^T (I - P_2) X_j}} - \rho_{ij}(A) \right | \\
&=  & \left | \frac{X_i^T (I - P_2) X_j / \left( \text{tr}(B^*) \sqrt{a_{ii}^* a_{jj}^* } \right)}{\sqrt{X_i^T (I - P_2) X_i / \left( a_{ii}^* \text{tr}(B^*) \right)} \sqrt{X_j^T (I - P_2) X_j / \left( a_{jj}^* \text{tr}(B^*) \right)} } - \rho_{ij}(A) \right | \\
 &\le &
\left | \frac{X_i^T (I - P_2) X_j / \left( \text{tr}(B^*) \sqrt{a_{ii}^* a_{jj}^* } \right) - \rho_{ij}(A)}{\sqrt{X_i^T (I - P_2) X_i / \left( a_{ii}^* \text{tr}(B^*) \right)} \sqrt{X_j^T (I - P_2) X_j / \left( a_{jj}^* \text{tr}(B^*) \right)} } \right | \\
&+  & \left | \frac{ \rho_{ij}(A)}{\sqrt{X_i^T (I - P_2) X_i / \left( a_{ii}^* \text{tr}(B^*) \right)} \sqrt{X_j^T (I - P_2) X_j / \left( a_{jj}^* \text{tr}(B^*) \right)} } - \rho_{ij}(A) \right | \\
 &\leq & 
\frac{\widetilde{\eta} }{1 - \widetilde{\eta} } +
  |\rho_{ij}(A)| \left| \frac{1}{1 - \widetilde{\eta}} - 1 \right|  \leq 3 \widetilde{\eta},
\eens
where the first inequality holds by $\mathcal{X}_0(A)$, and the second inequality holds for $\widetilde{\eta} < 1/3$.  

By Lemma~\ref{lemma::large-devi-rep}, the event $\mathcal{X}_0 = \mathcal{X}_0(B) \cap \mathcal{X}_0(A)$ holds with probability at least $1 - 3 / (n \vee m)^2$, which completes the proof.  
\end{proofof}

\subsection{Proof of Proposition \ref{rateConvergenceSampleCovToB}}
\label{rateConvergenceGroupCenterGemini:Bhat:proof}

We first present Lemma~\ref{lemma::BiasGroupCentering} 
and Lemma~\ref{SampleCovProjectionToB}, which decompose the rate of
convergence into a bias term and a variance term, respectively.  
We then combine the rates for the bias and variance terms to prove the entrywise rate of convergence for the sample covariance.  Define 
\begin{align} 
\mathcal{B}(B^*) &:= E[S(B^*)] - B^* \quad \text{ and } \label{BiasMatrixB} \\
\sigma(B^*) &:= S(B^*) - E[S(B^*)]. \label{VarianceMatrixB}
\end{align} 
We state maximum entrywise bounds on $\mathcal{B}(B^*)$ and
$\sigma(B^*)$ in Lemma \ref{lemma::BiasGroupCentering} and Lemma
\ref{SampleCovProjectionToB}, respectively.  
Proofs for these lemmas are provided in
Section~\ref{sec::proofofBiasGroupCentering} 
and~\ref{sec::proofofSampleCovProjectionToB} respectively.

\begin{lemma} 
\label{lemma::BiasGroupCentering}
For $\mathcal{B}(B^*)$ as defined in \eqref{BiasMatrixB}, 
\begin{equation}
\lVert \mathcal{B}(B^*) \rVert_{\max} \leq \frac{3 \lVert B^* \rVert_1}{n_{\min}}.
\end{equation}
\end{lemma}

\begin{lemma} \label{SampleCovProjectionToB}
Let $\sigma(B^*)$ be as defined in \eqref{VarianceMatrixB}.  With probability at least $1 - 2 / m^d$, 
\[
	|\sigma_{ij}(B^*)| = \left| S_{ij}(B^*) - b_{ij}^* \right| < C \log^{1/2}(m) \frac{\lVert A^* \rVert_F}{\operatorname{tr}(A^*)} \sqrt{\widetilde{b}_{ii} \widetilde{b}_{jj}}.
\] 
\end{lemma}

We now prove the entrywise rate of convergence for the sample covariance $S(B^*)$.  

\begin{proofof}{Proposition \ref{rateConvergenceSampleCovToB}}
%[Proof of Proposition \ref{rateConvergenceSampleCovToB}]
By the triangle inequality, 
\begin{align*} 
	 \left| S_{ij}(B^*) -  b_{ij}^* \right| &\leq \left| S_{ij}(B^*) - E[S_{ij}(B^*)] \right| + \left|E[S_{ij}(B^*)] -  b_{ij}^* \right| \\
	 &= |\mathcal{B}_{ij}(B^*)| + |\sigma_{ij}(B^*)| \\
	 &\leq \phi_{B, ij},
\end{align*} 
where the last step follows from Lemmas \ref{lemma::BiasGroupCentering} and \ref{SampleCovProjectionToB}.  
\end{proofof}

\textbf{Remark.}  Note that the first term of (\ref{tailCutpointSampleCovGroupCenToB}) is of order $\log^{1/2}(m) / \sqrt{m}$, and the second term is of order $\lVert B^* \rVert_1 / n_{\min}$.

\subsection{Proof of Proposition \ref{rateConvergenceSampleCovToA}}
\label{rateConvergenceGroupCenterGemini:Ahat:proof}
\begin{proofof2}
We express the $(i, j)$th entry of $S(A^*)$ as a quadratic form in
order to apply the Hanson-Wright inequality to obtain an entrywise
large deviation bound.  
Without loss of generality, let $i = 1$, $j = 2$.  The $(1, 2)$ entry of $S(A^*)$ can be expressed as a quadratic form, as follows,
\begin{align*}
  S_{12}(A^*) &= X_1^T (I - P_2) X_2 / n \\
  &= (1/2) \begin{bmatrix} X_1^T & X_2^T \end{bmatrix} \begin{bmatrix} 0 & (I - P_2) \\ (I - P_2) & 0 \end{bmatrix} \begin{bmatrix} X_1 \\ X_2 \end{bmatrix} / n \\
  &= (1/2) \begin{bmatrix} X_1^T & X_2^T \end{bmatrix} \left( \begin{bmatrix} 0 & 1 \\ 1 & 0 \end{bmatrix} \otimes (I - P_2) \right) \begin{bmatrix} X_1 \\ X_2 \end{bmatrix} / n.
\end{align*}
We decorrelate the random vector $(X_1, X_2) \in \mathbb{R}^{2n}$ so that we can apply the Hanson-Wright inequality.  The covariance matrix used for decorrelation is 
\[
  \text{Cov}\left( \begin{bmatrix} X_1 \\ 
X_2 \end{bmatrix} \right) = \begin{bmatrix} a_{11}^* & a_{12}^* \\ 
a_{21}^* & a_{22}^* \end{bmatrix} \otimes B^* =: A_{\{1, 2\}}^* \otimes B^*, 
\]
with 
\[
  A_{\{1, 2\}}^* = \begin{bmatrix} a_{11}^* & a_{12}^* \\ a_{21}^* & a_{22}^* \end{bmatrix} \in \mathbb{R}^{2 \times 2}.
\]
Decorrelating the quadratic form yields
\[
  S_{12}(A^*) = Z^T \Phi Z,
\]
where $Z \in \mathbb{R}^{2n}$, with $E[Z] = 0$ and $\text{Cov}(Z) = I_{2n \times 2n}$, and 
\begin{equation} \label{matrixHWforA}
  \Phi = (1 / 2n) \left( (A_{\{1, 2\}}^*)^{1/2} \begin{bmatrix} 0 & 1 \\ 1 & 0 \end{bmatrix} (A_{\{1, 2\}}^*)^{1/2} \right) \otimes B^{1/2}(I - P_2)B^{1/2}.
\end{equation}

To apply the Hanson-Wright inequality, we first find the trace and Frobenius norm of $\Phi$.  For the trace, note that 
\begin{equation}
\text{tr}\left( (A_{\{ 1, 2 \}}^*)^{1/2} \begin{bmatrix} 0 & 1 \\ 
1 & 0 \end{bmatrix} (A_{\{ 1, 2 \}}^*)^{1/2} \right) =
\text{tr}\left( \begin{bmatrix} 
0 & 1 \\ 
1 & 0 \end{bmatrix} A_{\{ 1, 2 \}}^* \right) = 2a_{12}^*.
\end{equation}
For the Frobenius norm, note that 
\begin{align*}
\left \lVert (A_{\{ 1, 2 \}}^*)^{1/2} \begin{bmatrix} 0 & 1 \\ 
1 & 0 \end{bmatrix} (A_{\{ 1, 2 \}}^*)^{1/2} \right \rVert_F^2 &=
\text{tr}\left( \begin{bmatrix} 0 & 1 \\ 
1 & 0 \end{bmatrix} A_{\{ 1, 2 \}}^* \begin{bmatrix} 0 & 1 \\ 
1 & 0 \end{bmatrix} A_{\{ 1, 2 \}}^* \right) \\
&= \text{tr}\left( \begin{bmatrix} a_{12}^{*2} + a_{11}^* a_{22}^* 
& 2 a_{12}^* a_{22}^* \\ 
2 a_{12}^* a_{22}^* 
& a_{12}^{*2} + a_{11}^* a_{22}^* \end{bmatrix} \right) \\
&= 2 a_{12}^{*2} + 2a_{11}^* a_{22}^*,
\end{align*}
Therefore the trace of $\Phi$ is
\begin{equation}
  \tr{\Phi } = a_{12}^* \tr{\widetilde{B}} / n,
\end{equation}
and the Frobenius norm of $\Phi$ is
\begin{equation}
  \lVert \Phi \rVert_F = (1/n) \sqrt{a_{12}^{*2} + a_{11}^* a_{22}^*} \lVert \widetilde{B} \rVert_F.
\end{equation}
Applying the Hanson-Wright inequality yields
\begin{align*}
  &P\left( \left| S_{12}(A^*) - a_{12}^* \tr{B^*} / n \right| > \phi_{A, 12} \right) \\
  &\leq P\left( \left| S_{12}(A^*) - a_{12}^* \tr{\widetilde{B}} / n
    \right| + (a_{12}^* / n) \left| \tr{\widetilde{B}} - \tr{B^*} \right| > \phi_{A, 12} \right) \\
  &= P\left( \left| S_{12}(A) - a_{12}^* \tr{\widetilde{B}} / n \right| > d^{1/2} K \log^{1/2}(n \vee m) \lVert \Phi \rVert_F \right) \\
  &\leq 2/ (n \vee m)^d.
\end{align*}
By the union bound,
\bens
\lefteqn{P\left( \forall i, j  \left| S_{ij}(A^*) - a_{ij} \tr{B^*} /
      n \right| < \phi_{A, ij} \right) } \\
&\geq & 1 - \sum_{i = 1}^m \sum_{j = 1}^m P\left( \left| S_{ij}(A^*) - a_{ij} \tr{B^*} / n \right| > \phi_{A, ij} \right) \\
&\geq & 1 - 2m^2 / (n \vee m)^d \geq 2/(n \vee m)^{d - 2}.
\eens
\end{proofof2}

\subsection{Proof of Lemma \ref{lemma::large-devi-rep}} 
\label{proofSampleCorrRate}

\begin{proofof2}
For the event \eqref{eventB_cov} from Proposition~\ref{rateConvergenceSampleCovToB},
\begin{equation*}
\left| S_{ij}(B^*) - b_{ij}^* \right| < \phi_{B, ij} = 
K^2 d \frac{\log^{1/2}(m)}{\sqrt{m}} C_A \sqrt{\widetilde{b}_{ii} \widetilde{b}_{jj}} + \left|  b_{ij}^* - \widetilde{b}_{ij} \right|,
\end{equation*}
dividing by $\sqrt{b_{ii}^* b_{jj}^*}$ yields 
\begin{align} \label{intermediateX0B}
\left| \frac{q_i XX^T q_j}{\text{tr}(A^*) \sqrt{b_{ii}^* b_{jj}^* }} -
  \rho_{ij}(B) \right| 
< K^2 d C_A \frac{\log^{1/2}(m) }{\sqrt{m}} \sqrt{
  \frac{\widetilde{b}_{ii} \widetilde{b}_{jj}}{b_{ii}^* b_{jj}^*} } 
+ \frac{\left| b_{ij} - \widetilde{b}_{ij} \right|}{\sqrt{ b_{ii}^* b_{jj}^* }}.  
\end{align}
By Lemma \ref{lemma::BiasGroupCentering}, 
\[
  \widetilde{b}_{ij} = b_{ij}\left[ 1 + O\left( \frac{\lVert B \rVert_1}{n} \right) \right], 
\]
so the right-hand side of \eqref{intermediateX0B} is less than or
equal to $\widetilde{\alpha}$.  
Hence event $\eqref{eventB_cov}$ implies $\X_0(B)$. Therefore, we know that $P(\mathcal{X}_0(B)) \geq 1 - 2/m^{d-2}$.

Similarly, event \eqref{eventA_cov} in Proposition~\ref{rateConvergenceSampleCovToA}:
\begin{align*}
&\left| S_{ij}(A^*) - a_{ij}^* \tr{B^*} / n \right| < \phi_{A, ij} \\
&= (a_{ij}^* / n) \left| \tr{\widetilde{B}} - \tr{B} \right| +  d^{1/2} K \log^{1/2}(n \vee m) (1/n) \sqrt{a_{ij}^{*2} + a_{ii}^* a_{jj}^*} \lVert \widetilde{B} \rVert_F,
\end{align*}
implies that
\begin{align*}
&\left| \frac{X_j^T (I - P_2) X_t}{\text{tr}(B^*) \sqrt{a_{jj}^* a_{tt}^* }} - \rho_{jt}(A) \right| \\
&< |\rho_{jt}(A)| \frac{\left| \tr{\widetilde{B}} - \tr{B^*} \right|}{\text{tr}(B^*)} + d^{1/2} K \log^{1/2}(n \vee m) \sqrt{\rho_{jt}(A)^2 + 1} \frac{\lVert \widetilde{B} \rVert_F}{\text{tr}(B^*)} \\
&= |\rho_{jt}(A)| \frac{\left| \tr{\widetilde{B}} - \tr{B^*} \right|}{\text{tr}(B^*)} + d^{1/2} K C_{B} \frac{\lVert \widetilde{B} \rVert_F}{\lVert B^* \rVert_F} \sqrt{\rho_{jt}(A)^2 + 1} \frac{\log^{1/2}(n \vee m)}{\sqrt{n}} \\
&\leq \widetilde{\eta},
\end{align*}
which is the event $\X_0(A)$. Therefore, we get that $P(\X_0(A)) \geq 1 - 2/(n \vee m)^d$.

We can obtain the $P(\X_0)$ by using a union bound put together $P(\X_0(B))$ and $P(\X_0(A))$, completing the proof.
\end{proofof2}

\subsection{Proof of Lemma \ref{lemma::BiasGroupCentering}}
\label{sec::proofofBiasGroupCentering}

\begin{proofof2}
Recall that $\widetilde{B} = (I - P_2)B^*(I - P_2)$.  The matrix $\widetilde{B} - B^*$ can be expressed as 
\[ 
	\widetilde{B} - B^* = (I - P_2)B^*(I - P_2) - B^* = - P_2 B^* - B^* P_2 + P_2 B^* P_2.
\] 
By the triangle inequality, $\lVert \widetilde{B} - B^* \rVert_{\max} \leq \lVert P_2 B^* \rVert_{\max} + \lVert B^* P_2 \rVert_{\max} + \lVert P_2 B^* P_2 \rVert_{\max}$.  We bound each term on the right-hand side.  

First we bound $\lVert P_2 B^* \rVert_{\max}$ and $\lVert B^* P_2 \rVert_{\max}$.  Let $p_i$ denote the $i$th column of $P_2$.  The $(i,j)$th entry satisfies
\[
  |p_i^T b_j^*| \leq \lVert B^*p_i \rVert_\infty \leq \lVert B^* \rVert_\infty \lVert p_i \rVert_\infty = \lVert B^* \rVert_1 \lVert p_i \rVert_\infty = \lVert B^* \rVert_1 / n_{\min},
\]
so $\lVert P_2 B^* \rVert_{\max} \leq \lVert B^* \rVert_1 / n_{\min}$.  
Because $P_2$ and $B^*$ are symmetric, $\lVert P_2 B^* \rVert_{\max} = \lVert B^* P_2 \rVert_{\max}$.

We now bound $\lVert P_2 B^* P_2 \rVert_{\max}$.  Let $B^{1/2}$ denote the symmetric square root of $B^*$.  We can express $p_i^T B^* p_j$ as an inner product $( B^{1/2} p_i)^T (B^{1/2} p_j)$, so 
\begin{align} 
	|(P_2 B^* P_2)_{ij}| &= | ( B^{1/2} p_i)^T (B^{1/2} p_j) | \leq \left( p_i^T B^* p_i \right)^{1/2} \left( p_j^T B^* p_j \right)^{1/2} \label{applyCauchySchwarzPBP} \\
	&\leq \lVert p_i \rVert_2 \lVert p_j \rVert_2 \lVert B \rVert_2 \leq \lVert B^* \rVert_2 / n_{\min}, \label{boundPBP}
\end{align} 
where \eqref{applyCauchySchwarzPBP} follows from the Cauchy Schwarz inequality, and \eqref{boundPBP} holds because 
\[ 
	\lVert p_i \rVert_2 = \begin{cases} 
	1 / \sqrt{n_1} & \text{if $i \in \{ 1, \ldots, n_1 \}$} \\
	1 / \sqrt{n_2} & \text{if $i \in \{ n_1 + 1, \ldots, n \}$}.
	\end{cases} 
\] 
\end{proofof2}

\subsection{Proof of Lemma~\ref{SampleCovProjectionToB}}
\label{sec::proofofSampleCovProjectionToB}
\begin{proofof2}
Let $B^{1/2}$ denote the symmetric square root of $B^*$.  Let $Z_j
= (a_{jj}^*B^*)^{-1/2}X_j$.  We express $S_{ij}(B^*)$ as a quadratic
form in order to use the Hanson-Wright inequality to prove a large
deviation bound.  
That is, we show that $S_{ij}(B^*) = \operatorname{vec}(Z)^T \HWMatProj^{ij} \operatorname{vec}(Z)$, with 
\ben
\label{HWMatrixSampleCovProjection}
  \HWMatProj^{ij} & =&  (1/m)A^* \otimes B^{1/2} (e_j - p_j) (e_i - p_i)^T
  B^{1/2}.
\een
We express $S_{ij}(B^*)$ as a quadratic form, as follows: 
\bens
S_{ij}(B^*) &= & \frac{1}{m} \sum_{k = 1}^m (e_i - p_i)^T X_k X_k^T (e_j - p_j) = \frac{1}{m} \sum_{k = 1}^m  \operatorname{tr}\left[ (e_i - p_i)^T X_k X_k^T (e_j - p_j) \right] \\
&= & \frac{1}{m} \sum_{k = 1}^m  X_k^T (e_j - p_j)(e_i - p_i)^T X_k \\
&= & \frac{1}{m} \operatorname{vec}(X)^T \left( I_{m \times m} \otimes (e_j - p_j)(e_i - p_i)^T \right) \operatorname{vec}(X) \\
&= & \operatorname{vec}(Z)^T \HWMatProj^{ij} \operatorname{vec}(Z) 
\eens
where
\ben
\label{TraceHWMatrixSampleCovProjection}
\operatorname{tr}(\HWMatProj^{ij}) &= & 
\operatorname{tr}(B^{1/2}(e_j - p_j) (e_i - p_i)^T B^{1/2}) = (e_i -
p_i)^T B^* (e_j - p_j) = \widetilde{b}_{ij}, \\
\label{FroHWMatrixSampleCovProjection}
\lVert \HWMatProj^{ij} \rVert_F &=  &\frac{1}{m} \lVert A^* \rVert_F
\lVert B^{1/2} (e_j - p_j) (e_i - p_i)^T B^{1/2} \rVert_F \\
 \nonumber
&= & \frac{1}{m} \lVert A^* \rVert_F \left((e_i - p_i)^T B^* (e_i -
  p_i) \right)^{1/2} \left((e_j - p_j)^T B^* (e_j - p_j) \right)^{1/2} 
= \frac{1}{m} \lVert A^* \rVert_F \sqrt{\widetilde{b}_{ii} \widetilde{b}_{jj}}.
\een
Therefore, we get that
\bens
\lefteqn{
P\left( \forall i, j \, \left| S_{ij}(B^*) - \widetilde{b}_{ij}
  \right| \leq K^2 d \log^{1/2}(m) \lVert \HWMatProj^{ij} \rVert_F /
  c' \right)} \\ 
&= &  
 P\left( \forall i, j \,  \left| \operatorname{vec}(Z)^T
     \HWMatProj^{ij} \operatorname{vec}(Z) - \tr{\HWMatProj^{ij}}
   \right| \leq K^2 d \log^{1/2}(m) \lVert \HWMatProj^{ij} \rVert_F /
   c'\right)\\
&\geq & 1 - 2 m^2 \exp\left( -c \min\left( d^2 \log(m) / c'^2, \frac{d \log^{1/2}(m) \lVert \HWMatProj^{ij} \rVert_F / c' }{\lVert \HWMatProj^{ij} \rVert_2} \right) \right) \\
&\geq & 1 - 2 / m^{d - 2}. \label{ProbEntrywiseErrorBTilde}
\eens
If the event 
$\left\{\forall i, j \, \left| S_{ij}(B^*) -  \widetilde{b}_{ij}
  \right| \leq K^2 d \log^{1/2}(m) \lVert \HWMatProj^{ij} \rVert_F /
  c'\right\}$ holds, it follows that
\bens
\left| S_{ij}(B^*) - b_{ij}^* \right| \leq \left| S_{ij}(B^*) - \widetilde{b}_{ij} \right| + | b_{ij}^* - \widetilde{b}_{ij} | 
\leq K^2 d \log^{1/2}(m) \lVert \HWMatProj^{ij} \rVert_F / c' + |
b_{ij} - \widetilde{b}_{ij} |. 
\label{equivalentToX0B}
\eens
The Lemma is thus proved.
\end{proofof2}

%!TEX root = submit.arxiv.tex

\pagebreak
\section{Proof of Theorem \ref{mainTheoremModSel} } \label{sec::proofTheorem3}

\subsection{Notation}

\begin{center}
\begin{tabular}{| c c c |}
\hline
Notation & Meaning & \\
\hline \hline
Mean structure & & \\
\hline
 $\mu \in \mathbb{R}^m$ & Vector of grand means of each gene &  \\
$\gamma \in \mathbb{R}^m$ & Vector of mean differences for each gene &  \\
$\nu = \inv{2} \begin{bmatrix} \inv{n_1} \Ones{n_1}^T & \inv{n_2} \Ones{n_2}^T \end{bmatrix}^T \in \mathbb{R}^n$ & Inner product with $\nu$ computes global mean & \\
\hline
Outcome of model selection step & & \\
\hline
$\GroupCenIndices \subset \Braces{ 1, 2, \ldots, m}$ & Indices selected for group centering & \\
$\GlobalCenIndices \subset \Braces{ 1, 2, \ldots, m}$ & Indices selected for global centering  & \\
% $\FixedTrueNeg = \Braces{ j \in \GlobalCenIndices : \gamma_j = 0}$ & True negatives in model selection step &  \\
% $\FixedFalseNeg = \Braces{ j \in \GlobalCenIndices : \gamma_j \neq 0}$ & False negatives in model selection step & \\
\hline
Sizes of gene subsets & & \\
\hline
$m_0 = \abs{\GroupCenIndices}$ & Number of group centered genes & \\
$m_1 = \abs{\GlobalCenIndices}$ & Number of globally centered genes & \\
% $\NumFixedTrueNeg = \abs{\FixedTrueNeg}$ & Number of true negatives & \\
% $\NumFixedFalseNeg = \abs{\FixedFalseNeg}$ & Number of false negatives & \\
% Note that $\NumGlobal = \NumFixedTrueNeg + \NumFixedFalseNeg$ & & \\
 \hline
 Projection matrices & & \\
 \hline
 $\PGlobal = \Ones{n} \nu^T$ & Projection matrix that performs global centering & \\
  $\PGroup$ (as in \eqref{suppPGroup})& Projection matrix that performs group centering & \\
 \hline
Sample covariance matrices & & \\
 \hline
 $S(B, J_0, J_1) = \frac{\NumGlobal}{m} \SBGlobal + \frac{\NumGroup}{m} \SBGroup$ & Model selection sample covariance matrix & \\
$S_1(B, J_1) = \frac{1}{\NumGlobal} \sum_{j \in \GlobalCenIndices} (\ResidGlobal) X_j X_j^T (\ResidGlobal)$ & Globally centered sample covariance matrix & \\
$S_2(B, J_0) = \frac{1}{\NumGroup} \sum_{j \in \GroupCenIndices} (\ResidGroup) X_j X_j^T (\ResidGroup)$ & Group centered sample covariance matrix & \\
 \hline
Decomposition of $S(B, J_0, J_1)$ &  & \\
 \hline
 $S_{\operatorname{I}}  = S(B, J_0, J_1) - \mathbb{E}\Brackets{S(B, J_0, J_1)}$ & Bias &  \\
%  & Rate: $\log(m) / n$ & \\
 $S_{\operatorname{II}} = \inv{m} (\ResidGlobal) \MGlobal \MGlobal^T (\ResidGlobal)$ & False negatives (deterministic) & \\
% & Rate: $\log(m) / n$ & \\
$S_{\operatorname{III}} = \inv{m} (\ResidGlobal) \MGlobal \varepsilon^T (\ResidGlobal)$ & False negatives (random) &  \\
% & Rate: $\log(m) / \sqrt{n \NumGlobal}$ &  \\
$S_{\operatorname{IV}} = m^{-1} (I - P_2) \varepsilon_{J_0} \varepsilon_{J_0}^T (I - P_2) + $ & True negatives & \\
\qquad \qquad $m^{-1} (I - P_1) \varepsilon_{J_1} \varepsilon_{J_1}^T (I - P_1)$ &  & \\
\hline
\end{tabular}
\end{center}

\subsection{Two-Group Model and Centering}

We begin by introducing some relevant notation for the two-group model and centering. Define the group membership vector $\GroupIndicators \in \mathbb{R}^n$ as
\begin{equation} \label{def:GroupIndicators}
\GroupIndicators := \begin{bmatrix} \Ones{n_1}^T & -\Ones{n_2}^T \end{bmatrix}^T \in \mathbb{R}^n.
\end{equation}

In the two-group model, the mean matrix $M$ can be expressed as
\begin{equation} \label{meanDecomposition}
M = \Ones{n} \mu^T + (1/2) \GroupIndicators \gamma^T,
\end{equation}
where $\mu \in \mathbb{R}^m$ is a vector of grand means, and $\gamma \in \mathbb{R}^m$ is the vector of mean differences.  According to \eqref{meanDecomposition}, the $(i, j)$th entry of $M$ can be expressed as
\begin{equation}
m_{ij} = \begin{cases}
\mu_j + \gamma_j / 2 & \text{if sample $i$ is in group one} \\
\mu_j - \gamma_j /  2 & \text{if sample $i$ is in group two}.
\end{cases}
\end{equation}

Define the vector $\nu \in \mathbb{R}^n$ as
\begin{equation} \label{def:VecAvgSampleMeans}
	\nu = \inv{2} \begin{bmatrix} \inv{n_1} \Ones{n_1}^T & \inv{n_2} \Ones{n_2}^T \end{bmatrix}^T \in \mathbb{R}^n,
\end{equation}
so that for the $j$th column of the data matrix $X_j \in \mathbb{R}^n$,
\begin{equation}
\expct{\nu^T X_j} = \inv{2} \expct{  \inv{n_1} \sum_{k = 1}^{n_1} X_{jk} + \inv{n_2} \sum_{k = n_1 + 1}^{n} X_{jk} } = \mu_j.
\end{equation}
Note that
\begin{equation} \label{innerProductsNu}
	\nu^T \Ones{n} = (1/2) ( 1 + 1) = 1, \quad \text{ and } \quad \nu^T \GroupIndicators = (1/2) (1 - 1) = 0.
\end{equation}
Next we define a projection matrix that performs global centering.  Define the non-orthogonal projection matrix
\begin{equation} \label{def:PGlobal}
\PGlobal := \Ones{n} \nu^T \in \mathbb{R}^{n \times n}.
\end{equation}
Applying the projection matrix to the mean matrix yields
\begin{equation}
\PGlobal M = \Ones{n} \nu^T \left( \Ones{n} \mu^T + (1/2) \GroupIndicators \gamma^T \right) = \Ones{n} \mu^T + (1/2) (\nu^T \GroupIndicators) \Ones{n} \gamma^T = \Ones{n} \mu^T,
\end{equation}
with residuals
\begin{equation} \label{ResdualsM}
	(\ResidGlobal) M = M - \PGlobal M = M - \Ones{n} \mu^T = (1/2) \GroupIndicators \gamma^T.
\end{equation}

Define
\begin{equation} \label{suppPGroup}
P_2 = \begin{bmatrix}
n_1^{-1} \Ones{n_1} \Ones{n_1}^T & \\
& n_2^{-1} \Ones{n_2} \Ones{n_2}^T
\end{bmatrix}.
\end{equation}
Note that $P_2 \Ones{n} = \Ones{n}$ and $P_2 \delta_n = \delta_n$, so
\begin{equation} 
P_2 M = P_2 \Ones{n} \mu^T + (1/2) P_2 \GroupIndicators \gamma^T = \Ones{n} \mu^T + (1/2) \GroupIndicators \gamma^T = M,
\end{equation}
and therefore $(I - P_2)M = 0$.

Define
\begin{align}
\widecheck{B} &= (I - P_1) B (I - P_1) \label{BP1P1} = \Parens{ \widecheck{b}_{ij}} \\
\widetilde{B} &= (I - P_2) B (I - P_2) \label{BP2P2} = \Parens{ \widetilde{b}_{ij}}  \\
\breve{B} &= (I - P_1) B (I - P_2) \label{BP1P2} = \Parens{ \breve{b}_{ij}}.
\end{align}
Let $\widecheck{b}_{\max}$, $\widetilde{b}_{\max}$, and $\breve{b}_{\max}$ denote the maximum diagonal entries of $\widecheck{B}$, $\widetilde{B}$, and $\breve{B}$, respectively.  

\subsection{Model Selection Centering}

For a subset $J \subset \{ 1, \ldots, m\}$, let $X_J$ denote the submatrix of $X$ consisting of columns indexed by $J$.  For the fixed sets of genes $J_0$ and $J_1$, define the sample covariance
\begin{equation} \label{SampleCovFixedGeneSets}
S(B, J_0, J_1) = m^{-1} \sum_{k \in J_0} (I - P_2) X_k X_k^T (I - P_2)^T + m^{-1} \sum_{k \in J_1} (I - P_1) X_k X_k^T (I - P_1)^T =: \operatorname{I} + \operatorname{II}.
\end{equation}
Note that $\mathbb{E}\Brackets{S(B, J_0, J_1)} = B^\sharp$, with
\begin{equation} \label{expectedValueModSelSB}
B^\sharp = \frac{\Trace{A_{J_0}}}{m} (I - P_2) B(I - P_2) + \frac{\Trace{A_{J_1}}}{m} (I - P_1) B (I - P_1).
\end{equation}

Define the sample correlation matrix,
\begin{equation}
\label{defGammaBModSel}
\widehat{\Gamma}_{ij}(B) = \frac{(S(B, J_0, J_1))_{ij} }{ \sqrt{(S(B, J_0, J_1))_{ii}(S(B, J_0, J_1))_{jj}}}.
\end{equation}

The baseline Gemini estimators \cite{Zhou14a} are then defined as follows,
using a pair of penalized estimators for the correlation matrices $\rho(A) = (a_{ij}/\sqrt{a_{ii} a_{jj}})$
and $\rho(B) = (b_{ij}/\sqrt{b_{ii} b_{jj}})$:
\begin{subeqnarray}
\label{geminiObjectiveFnASupplement}
	\widehat{A}_\rho &= &
\argmin_{A_\rho \succ 0} \left\{ \tr{\widehat{\Gamma}(A) A_\rho^{-1}}
  + \log |A_\rho|
+ \lambda_B |A_\rho^{-1}|_{1, \text{off}} \right\}, \\
\label{geminiObjectiveFnBSupplement}
\widehat{B}_\rho &= &
\argmin_{B_\rho \succ 0} \left\{ \tr{\widehat{\Gamma}(B) B_\rho^{-1}}
  + \log |B_\rho| + \lambda_A |B_\rho^{-1}|_{1, \text{off}} \right\}.
\end{subeqnarray}
We will focus on $\widehat{B}_\rho$ using the input as defined in \eqref{defGammaBModSel}.
% where the input are a pair of sample correlation matrices as defined in \eqref{defGammaAGammaB}.

The proof proceeds as follows. Lemma~\ref{entrywiseErrorFixedSetGenes}, the equivalent of Proposition~\ref{rateConvergenceSampleCovToB} for Algorithm 1, establishes entry-wise convergence rates of the sample covariance matrix for fixed sets of group and globally centered genes. We use this to prove Theorem~\ref{mainTheoremFixedGenes} below in Section~\ref{MainTheoremFixedGenes} and to prove Theorem~\ref{mainTheoremModSel} in Section~\ref{proofMainThmAlgTwo}.

\subsection{Convergence for fixed gene sets}
% \subsection{Theorem \ref{mainTheoremFixedGenes} for fixed sets of group centered and globally centered genes}
\label{MainTheoremFixedGenes}
% \section{Algorithm and Proof of Main Theorem for Model Selection Centering}

We first state a standalone result, Theorem~\ref{mainTheoremFixedGenes}, which provides rates of convergence when $S(B, J_0, J_1)$ as in \eqref{SampleCovFixedGeneSets} is calculated using fixed sets of group centered and globally centered genes, $J_0$ and $J_1$, respectively.  This result shows how the algorithm used in the preliminary step to choose which genes to group center can be decoupled from the rest of the estimation procedure. The proof is presented below in Section~\ref{proofMainTheoremFixedGenes}.

\begin{theorem} \label{mainTheoremFixedGenes}
Suppose that (A1), (A2'), and (A3) hold.  Let $J_0$ and $J_1$ denote sets such that $J_0 \cap J_1 = \emptyset$ and $J_0 \cup J_1 = \{ 1, \ldots, m \}$.  Let $m_0 = \abs{J_0}$ and $m_1 = \abs{J_1}$ denote the sizes of the sets.  Let $\tau_{\text{global}} > 0$ satisfy
\begin{equation} \label{smallFalseNegatives}
\max_{j \in J_1} \abs{\gamma_j} \leq \tau_{\text{global}} ,
\end{equation}
for $\tau_{\text{global}} = C \sqrt{\log(m)} \lVert (D^T B^{-1}D)^{-1} \rVert_2^{1/2} \asymp \sqrt{\frac{\log(m)}{n}}$.

  Consider the data as generated from model \eqref{meanDecomposition} with $\varepsilon = B^{1/2} Z A^{1/2}$,  where $A \in \mathbb{R}^{m \times m}$ and $B \in \mathbb{R}^{n \times n}$ are positive definite matrices, and $Z$ is an $n \times m$ random matrix as defined in Theorem \ref{thm::GLSFixedB}.
%with independent entries $Z_{ij}$ satisfying
%$\E Z_{ij} = 0$, $1 = \E Z_{ij}^2 \le \norm{Z_{ij}}_{\psi_2} \leq K$.
 Let $\lambda_A$ denote the penalty parameter for estimating $B$.  Suppose the penalty parameter $\lambda_A$ in \eqref{geminiObjectiveFnBSupplement} satisfies
\begin{equation}
\lambda_A \geq  C'' \Brackets{ C_A K \frac{\log^{1/2}(m \vee n)}{\sqrt{m}} + \frac{\lVert B \rVert_1}{n_{\min}} }.
\end{equation}
where $C''$ is an absolute constant.

\textbf{(I)} Let $\mathcal{E}_4(J_0, J_1)$ be the event such that
\begin{equation} \label{GoodAlgTwoOpErrorBinv}
\twonorm{ \Trace{A} \Parens{\widehat{W}_2 \widehat{B}_{\rho} \widehat{W}_2}^{-1} - B^{-1} } \leq \frac{C' \lambda_A \sqrt{\abs{B^{-1}}_{0, \operatorname{off}} \vee 1 } }{b_{\min} \varphi_{\min}^2(\rho(B))}.
\end{equation}
Then $P(\mathcal{E}_4(J_0, J_1)) \geq 1 - C / m^d$.

\textbf{(II)}
% Let $\widehat{\beta}$ be defined as in (\ref{GLSestimator}) with $\widehat{B}^{-1}$ being defined as in \eqref{BiHat} and $D$ as in (\ref{meanMatrixTwoGroups}).
With probability at least $1 - C' / m^d$, for all $j$,
\begin{equation}
\label{mainThmBoundTailCutpointSimplified}
	\lVert \widehat{\beta}_j(\widehat{B}^{-1}) - \beta_j^* \rVert_2 \leq C_1 \lambda_A \sqrt{ \frac{n_{\text{ratio}} \left(|B^{-1}|_{0, \text{off}} \vee 1\right)}{n_{\min}}} + C_2 \sqrt{\log(m)} \lVert (D^T B^{-1} D)^{-1} \rVert_2^{1/2}.
\end{equation}
\end{theorem}

\subsubsection{Decomposition of sample covariance matrix}
The error in the sample covariance $S(B, J_0, J_1)$ can be decomposed as
\begin{equation}
S(B, J_0, J_1) - B = \Brackets{B^\sharp - B} + \Brackets{S(B, J_0, J_1) - B^\sharp},
\end{equation}
where the first term corresponds to bias and the second term to variance.  We now further decompose the variance term.  The first term of $S(B, J_0, J_1)$ in \eqref{SampleCovFixedGeneSets} can be decomposed as,
\begin{align}
\operatorname{I} &= m^{-1} (I - P_2) X_{J_0} X_{J_0}^T (I - P_2) \notag \\
&= m^{-1} (I - P_2) (M_{J_0} + \varepsilon_{J_0})(M_{J_0} + \varepsilon_{J_0})^T (I - P_2) \notag \\
&=  m^{-1} (I - P_2) \varepsilon_{J_0} \varepsilon_{J_0}^T (I - P_2) + m^{-1} (I - P_2) M_{J_0} \varepsilon_{J_0}^T (I - P_2)  \notag \\
&\qquad + m^{-1} (I - P_2) \varepsilon_{J_0} M_{J_0}^T (I - P_2) + m^{-1} (I - P_2) M_{J_0} M_{J_0}^T (I - P_2), \label{SampleCovGroupDecomp}
\end{align}
and the second term can be decomposed analogously, as
\begin{align}
\operatorname{II} &=  m^{-1} (I - P_1) \varepsilon_{J_1} \varepsilon_{J_1}^T (I - P_1) + m^{-1} (I - P_1) M_{J_1} \varepsilon_{J_1}^T (I - P_1) \notag \\
&\qquad + m^{-1} (I - P_1) \varepsilon_{J_1} M_{J_1}^T (I - P_1) + m^{-1} (I - P_1) M_{J_1} M_{J_1}^T (I - P_1). \label{SampleCovGlobalDecomp}
\end{align}
By the above decompositions, it follows that $S(B, J_0, J_1)$ can be expressed as
\begin{equation} \label{eq::modSelSampleCovDecomposition}
	S(B, J_0, J_1) = S_{\operatorname{II}} + S_{\operatorname{III}} + S_{\operatorname{III}}^T + S_{\operatorname{IV}},
\end{equation}
with
\begin{align}
S_{\operatorname{II}} &= m^{-1} (I - P_2) M_{J_0} M_{J_0}^T (I - P_2) + m^{-1} (I - P_1) M_{J_1} M_{J_1}^T (I - P_1). \label{SampleCovTermThreeFull} \\
S_{\operatorname{III}} &= m^{-1} (I - P_2) M_{J_0} \varepsilon_{J_0}^T (I - P_2) + m^{-1} (I - P_1) M_{J_1} \varepsilon_{J_1}^T (I - P_1)  \label{SampleCovTermTwoFull} \\
S_{\operatorname{IV}} &= m^{-1} (I - P_2) \varepsilon_{J_0} \varepsilon_{J_0}^T (I - P_2) + m^{-1} (I - P_1) \varepsilon_{J_1} \varepsilon_{J_1}^T (I - P_1) \label{SampleCovTermOne}
% \GlobalThree &= \inv{\NumGlobal} (\ResidGlobal) \varepsilon M^T (\ResidGlobal) \\
\end{align}
For each of $S_{\operatorname{II}}$, $S_{\operatorname{III}}$, and $S_{\operatorname{IV}}$, the first term comes from \eqref{SampleCovGroupDecomp} and the second term comes from \eqref{SampleCovGlobalDecomp}.

The terms $S_{\operatorname{II}}$ and $S_{\operatorname{III}}$ can be simplified, as follows.  Because $(I - P_2)M_{J_0} = 0$, it follows that the first term of $S_{\operatorname{II}}$ is zero:
\[
	m^{-1} (I - P_2) M_{J_0} M_{J_0}^T (I - P_2) = 0.
\]
and the first term of $S_{\operatorname{III}}$ is also zero,
\[
	m^{-1} (I - P_2) M_{J_0} \varepsilon_{J_0}^T (I - P_2) = 0,
\]

Therefore the terms $S_{\operatorname{II}}$ and $S_{\operatorname{III}}$ are equal to
\begin{align}
S_{\operatorname{II}} &= m^{-1} (I - P_1) M_{J_1} M_{J_1}^T (I - P_1), \label{SampleCovTermThree} \\
S_{\operatorname{III}} &=  m^{-1} (I - P_1) M_{J_1} \varepsilon_{J_1}^T (I - P_1).  \label{SampleCovTermTwo} 
\end{align}

Let $S_{\operatorname{I}} = B^\sharp - B$.  We have thus decomposed the error in the sample covariance as
\begin{equation} \label{ModSelBiasVarDecomp}
S(B, J_0, J_1) - B = \underbrace{S_{\operatorname{I}}}_{\text{bias}} +\underbrace{\Brackets{ \Parens{S_{\operatorname{IV}} - B^{\sharp}} + S_{\operatorname{III}} + S_{\operatorname{II}}}}_{\text{variance}}.
\end{equation}
In Lemma \ref{LemmaEntrywiseSBModSel}, we provide an error bound for each term in the decomposition \eqref{ModSelBiasVarDecomp}.

We next state Lemma~\ref{entrywiseErrorFixedSetGenes}, which establishes the maximum of entry-wise errors for estimating $B$ using the sample covariance for fixed gene sets as defined in \eqref{ModSelBiasVarDecomp}.  Lemma~\ref{entrywiseErrorFixedSetGenes} is used in the proof of Theorem \ref{mainTheoremFixedGenes}.  Following, we state Lemma~\ref{LemmaEntrywiseSBModSel}, which is used in the proof of Lemma~\ref{entrywiseErrorFixedSetGenes}.
\begin{lemma} \label{entrywiseErrorFixedSetGenes}
Suppose the conditions of Theorem \ref{mainTheoremFixedGenes} hold.  Let $\mathcal{E}_6(J_0, J_1)$ denote the event
\begin{equation} \label{entrywiseBoundModSel}
\mathcal{E}_6(J_0, J_1) = \Braces{ \InfNorm{S(B, J_0, J_1) - B} \leq  C_A K \frac{\log^{1/2}(m \vee n)}{\sqrt{m}} + \frac{\lVert B \rVert_1}{n_{\min}} }.
\end{equation}
Then $\mathcal{E}_6(J_0, J_1)$ holds with probability at least $1 - \frac{8}{(m \vee n)^2}$.
\end{lemma}

\begin{lemma} \label{LemmaEntrywiseSBModSel}
Let the model selection-based sample covariance $S(B, J_0, J_1)$ be as defined in \eqref{SampleCovFixedGeneSets}, where $\GlobalCenIndices$ and $\GroupCenIndices$ are fixed sets of variables that are globally centered, and group centered, respectively.  Let $m_0 = \abs{J_0}$ and $m_1 = \abs{J_1}$.  Define the rates
\begin{align}
r_1 &= \frac{3 \OneNorm{B}}{n_{\min}}, \\
r_2 &= (4 m)^{-1} \twonorm{\gammaGlobal}^2, \label{r2Rate} \\
r_3 &= C_3 d^{1/2} K^2 \log^{1/2}(m) m^{-1} \Parens{ \gammaGlobal^T A_{J_1} \gammaGlobal}^{1/2} \widecheck{b}_{\max}^{1/2}, \label{r3Rate} \\
r_4 &= C_4 d^{1/2} K \log^{1/2}(m) m^{-1} \fnorm{A} \twonorm{B}.
\end{align}

\textbf{(I)} Deterministically,
\begin{align} \label{entrywiseBoundSThreeSFour}
\InfNorm{B^\sharp - B} \leq r_1\quad \text{ and } \quad \InfNorm{S_{\operatorname{II}}} &\leq r_2.
\end{align}

\textbf{(II)} Define the events
\begin{equation} \label{entrywiseProbs}
\SOne = \Braces{ \InfNorm{S_{\operatorname{IV}} - B^{\sharp}} \leq r_4 } \quad \text{ and }  \quad \STwo = \Braces{ \InfNorm{S_{\operatorname{III}}} \leq r_3}.
\end{equation}
Then $\SOne$ and $\STwo$ occur with probability at least $1 - 2/m^d$.
% \begin{equation} \label{entrywiseProbs}
% P\Parens{ \SOne } \geq 1 - \frac{2}{m^d} \quad \text{ and } \quad P\Parens{\STwo} \geq 1 - \frac{2}{m^d}.
% \end{equation}

\end{lemma}

Lemmas~\ref{entrywiseErrorFixedSetGenes} and~\ref{LemmaEntrywiseSBModSel} are proved in Section~\ref{sec::LemmasForTheorem3}. We analyze term $S_{\operatorname{I}}$ in Section \ref{entrywiseTermI}, term $S_{\operatorname{II}}$ in Section \ref{entrywiseTermII}, term $S_{\operatorname{III}}$ in Section \ref{entrywiseTermIII}, and term $S_{\operatorname{IV}}$ in Section \ref{entrywiseTermIV}.

\subsubsection{Proof of Theorem \ref{mainTheoremFixedGenes}}
\label{proofMainTheoremFixedGenes}
Let us first define the event $\mathcal{E}_{\text{global}}$, that is, the GLS error based on the true $B^{-1}$ is small:
\begin{equation} \label{EGlobal}
\mathcal{E}_{\text{global}} = \Braces{ \InfNorm{\widehat{\gamma}(B^{-1}) - \gamma} < \sqrt{\log(m)} \lVert (D^T B^{-1}D)^{-1} \rVert_2^{1/2}}.
\end{equation}

% \begin{proof}[Proof of Theorem \ref{mainTheoremFixedGenes}]
Let $\mathcal{E}_4(J_0, J_1)$ be defined as in \eqref{GoodAlgTwoOpErrorBinv}, denoting small operator norm error in estimating $B^{-1}$:
\begin{equation}  
\mathcal{E}_4(J_0, J_1) = \Braces{
\twonorm{ \Trace{A} \Parens{\widehat{W}_2 \widehat{B}_{\rho} \widehat{W}_2}^{-1} - B^{-1} } \leq \frac{C' \lambda_A \sqrt{\abs{B^{-1}}_{0, \operatorname{off}} \vee 1 } }{b_{\min} \varphi_{\min}^2(\rho(B))} }.
\end{equation}
Note that $\mathcal{E}_4(J_0, J_1)$ holds deterministically under event $\mathcal{E}_{6}(J_0, J_1)$ as defined in \eqref{entrywiseBoundModSel} of Lemma~\ref{entrywiseErrorFixedSetGenes}.
%, which itself holds with high probability by Lemma \ref{entrywiseErrorFixedSetGenes}.

Define the event bounding the perturbation in mean estimation due to error in estimating $B^{-1}$:
\begin{equation} \label{GoodMeanFixedB}
\mathcal{E}_5(J_0, J_1) = \Braces{ \InfNorm{ \widehat{\gamma}(\widehat{B}^{-1}) - \widehat{\gamma}(B^{-1})  } <  C n_{\min}^{-1/2} \twonorm{\widehat{B}^{-1} - B^{-1}} }.
\end{equation}
Conditional on a fixed matrix $\widehat{B}^{-1}$ that satisfies $\mathcal{E}_4(J_0, J_1)$, event $\mathcal{E}_5(J_0, J_1)$ holds with probability at least $1 - C / m^d$, by Lemma \ref{BetaBtildeBetaBStarHansonWright} (used in the proof of Theorem \ref{thm::GLSFixedB}).

The overall rate of convergence follows by applying the union bound to the events $\mathcal{E}_{\text{global}} \cap \mathcal{E}_4(J_0, J_1) \cap \mathcal{E}_5(J_0, J_1)$, as follows:
\begin{align*}
&P(\mathcal{E}_{\text{global}}^c \cup \mathcal{E}_4(J_0, J_1)^c \cup \mathcal{E}_5(J_0, J_1)^c) \\
&\leq P(\mathcal{E}_{\text{global}}^c) +  P(\mathcal{E}_4(J_0, J_1)^c) +  P(\mathcal{E}_5(J_0, J_1)^c \mid \mathcal{E}_4(J_0, J_1))P(\mathcal{E}_4(J_0, J_1)) \\
&\qquad + P(\mathcal{E}_5(J_0, J_1)^c \mid \mathcal{E}_4(J_0, J_1)^c) P(\mathcal{E}_4(J_0, J_1)^c) \\
&\leq P(\mathcal{E}_{\text{global}}^c) +  P(\mathcal{E}_4(J_0, J_1)^c) + P(\mathcal{E}_4(J_0, J_1)^c) + P(\mathcal{E}_5(J_0, J_1)^c \mid \mathcal{E}_4(J_0, J_1)) \\
&= P(\mathcal{E}_{\text{global}}^c) +  2P(\mathcal{E}_4(J_0, J_1)^c) + P(\mathcal{E}_5(J_0, J_1)^c \mid \mathcal{E}_4(J_0, J_1)),
\end{align*}
where $P(\mathcal{E}_{\text{global}}^c)$ and $P(\mathcal{E}_5(J_0, J_1)^c \mid \mathcal{E}_4(J_0, J_1))$ are bounded in Theorem \ref{thm::GLSFixedB}, and $P(\mathcal{E}_4(J_0, J_1)^c)$ has high probability under Lemma \ref{entrywiseErrorFixedSetGenes}.
%, and $P(\mathcal{E}_5(J_0, J_1)^c \mid \mathcal{E}_4(J_0, J_1))$

\subsection{Proof of Theorem \ref{mainTheoremModSel}}
\label{proofMainThmAlgTwo}

Let $\widehat{\gamma}^{\operatorname{init}}$ denote the output from Algorithm 1.  By our choice of the threshold parameter $\tau_{\text{init}}$ as in \eqref{modSelThresh}, that is,
\[
	\tau_{\text{init}} =C \left(
          \frac{\log^{1/2}(m)}{\sqrt{m}} + \frac{\lVert B
            \rVert_1}{n_{\min}} \right)
\sqrt{ \frac{n_{\text{ratio}} \left(|B^{-1}|_{0, \text{off}} \vee
      1\right)}{n_{\min}}}
+ C\sqrt{\log(m)} \lVert (D^T B^{-1}D)^{-1} \rVert_2^{1/2},
\]
we have a partition $( \widetilde{J}_0, \widetilde{J}_1)$ such that $\widetilde{J}_0$ is the set of variables selected for group centering and $\widetilde{J}_1$ is the set of variables selected for global centering.   The partition results in a sample covariance matrix $S(B, \widetilde{J_0}, \widetilde{J_1})$ as defined in \eqref{SampleCovFixedGeneSets}.  Define the event that the Algorithm 1 estimate $\widehat{\gamma}^{\operatorname{init}}$ is close to $\gamma$ in the sense that  
\begin{equation} \label{AlgOneGood}
	\mathcal{E}_{A1} = \Braces{ \InfNorm{\widehat{\gamma}^{\operatorname{init}} - \gamma} < \tau_{\text{init}}}.
\end{equation}
Note that the event $\mathcal{E}_{A1}$ implies that the false negatives have small true mean differences.  That is, on event $\mathcal{E}_{A1}$, by the triangle inequality,
 \begin{align}
	\InfNorm{\gamma_{\widetilde{J}_1}} \leq \InfNorm{\gamma_{\widetilde{J}_1} - \widehat{\gamma}^{\operatorname{init}}_{\widetilde{J}_1} } + \InfNorm{\widehat{\gamma}^{\operatorname{init}}_{\widetilde{J}_1}} \leq \tau_{\text{init}} + \tau_{\text{init}} = 2\tau_{\text{init}},
 \end{align}
 where $\InfNorm{\widehat{\gamma}^{\operatorname{init}}_{\widetilde{J}_1}} < \tau_{\text{init}}$ by definition of $\mathcal{E}_{A1}$, and $\InfNorm{\gamma_{\widetilde{J}_1} - \widehat{\gamma}^{\operatorname{init}}_{\widetilde{J}_1} } < \tau_{\text{init}}$ by definition of the thresholding set $\widetilde{J}_1$.

 Under the assumptions of Theorem \ref{mainTheoremFixedGenes}, $\tau_{\text{init}} \leq \tau_{\text{global}}$ with $\tau_{\text{global}}$ as defined in  \eqref{smallFalseNegatives}, so condition \eqref{smallFalseNegatives} of Theorem \ref{mainTheoremFixedGenes} is satisfied.  Under the conditions of Theorem \ref{mainTheoremFixedGenes}, event $\mathcal{E}_6(J_0, J_1)$ as defined in Lemma \ref{entrywiseErrorFixedSetGenes} holds with high probability; that is, the entrywise error in the sample covariance matrix is small.

 Let $\mathcal{E}_B$ denote event \eqref{mainThmErrorCov} in Theorem \ref{mainTheoremModSel}.  In view of Theorem \ref{thm::frob} and Lemma \ref{boundW1W2errorOpFro}, event $\mathcal{E}_B$ holds on $\mathcal{E}_6(J_0, J_1)$.  Hence
 \begin{align*}
P\Parens{ \mathcal{E}_B^c } &= P\Parens{ \mathcal{E}_6(J_0, J_1)^c \mid \mathcal{E}_{A1}} P\Parens{ \mathcal{E}_{A1} } + P\Parens{ \mathcal{E}_6(J_0, J_1)^c \mid \mathcal{E}_{A1}^c } P\Parens{ \mathcal{E}_{A1}^c } \\
&\leq P\Parens{ \mathcal{E}_6(J_0, J_1)^c \mid \mathcal{E}_{A1}} + P\Parens{ \mathcal{E}_{A1}^c } \\
&\leq 2 / m^d + 2 / m^d,
 \end{align*}
 where the first term is bounded in Lemma \ref{entrywiseErrorFixedSetGenes} and the second in Theorem \ref{mainTheoremGroupCentering}.

Recall the event $\mathcal{E}_{\text{global}}$ as defined in \eqref{EGlobal}.  Event \eqref{mainThmBoundTailCutpoint} in Theorem \ref{mainTheoremModSel} holds under the intersection of events $\mathcal{E}_{\text{global}} \cap \mathcal{E}_5(\widetilde{J}_0, \widetilde{J}_1) \cap \mathcal{E}_B \cap \mathcal{E}_{A1}$.  Hence the probability of \eqref{mainThmBoundTailCutpoint} can be bounded as follows:
\begin{align*}
&P(\mathcal{E}_{\text{global}}^c \cup \mathcal{E}_5(\widetilde{J}_0, \widetilde{J}_1)^c \cup \mathcal{E}_B^c \cup \mathcal{E}_{A1}^c ) \\
&\leq P(\mathcal{E}_{\text{global}}^c) +  P(\mathcal{E}_B^c) +  P(\mathcal{E}_5(\widetilde{J}_0, \widetilde{J}_1)^c \mid \mathcal{E}_B)P(\mathcal{E}_B) \\
&\qquad + P(\mathcal{E}_5(\widetilde{J}_0, \widetilde{J}_1)^c \mid \mathcal{E}_B^c) P(\mathcal{E}_B^c) + P\Parens{ \mathcal{E}_{A1}^c} \\
&\leq P(\mathcal{E}_{\text{global}}^c) +  P(\mathcal{E}_B^c) + P(\mathcal{E}_B^c) + P(\mathcal{E}_5(\widetilde{J}_0, \widetilde{J}_1)^c \mid \mathcal{E}_B) + P\Parens{ \mathcal{E}_{A1}^c} \\
&= P(\mathcal{E}_{\text{global}}^c) +  2P(\mathcal{E}_B^c) + P(\mathcal{E}_5(\widetilde{J}_0, \widetilde{J}_1)^c \mid \mathcal{E}_B) + P\Parens{ \mathcal{E}_{A1}^c},
\end{align*}
where $P(\mathcal{E}_{\text{global}}^c)$ and $P(\mathcal{E}_5(\widetilde{J}_0, \widetilde{J}_1)^c \mid \mathcal{E}_B)$ are bounded in Theorem 1, $P(\mathcal{E}_B^c)$ is bounded above, and $P\Parens{ \mathcal{E}_{A1}^c}$ is bounded in Theorem \ref{mainTheoremGroupCentering}.

% \section{Additional lemmas used in the proof of Theorem \ref{mainTheoremModSel}}
% \subsection{Unified analysis of entrywise error in model selection sample covariance}

\section{Proof of Lemmas \ref{entrywiseErrorFixedSetGenes} and \ref{LemmaEntrywiseSBModSel}} 
\label{sec::LemmasForTheorem3}

We first prove Lemma~\ref{entrywiseErrorFixedSetGenes} in Section~\ref{proofEntrywiseErrorFixedSetGenes}. The rest of the section contains the proof of Lemma \ref{LemmaEntrywiseSBModSel}, where part I is proved in Sections \ref{entrywiseTermI} and \ref{entrywiseTermII} and part II in Sections \ref{entrywiseTermIII} and \ref{entrywiseTermIV}.

\subsection{Proof of Lemma \ref{entrywiseErrorFixedSetGenes}}
\label{proofEntrywiseErrorFixedSetGenes}

% \textbf{Proof of Lemma \ref{entrywiseErrorFixedSetGenes}} %\label{proofEntrywiseRateModSel}

The entrywise error in the sample covariance matrix \eqref{SampleCovFixedGeneSets} can be decomposed as
\begin{align}
\InfNorm{ S(B, J_0, J_1) - B} &\leq \InfNorm{S(B, J_0, J_1) - B^\sharp} + \InfNorm{B^\sharp - B} \\
&\leq \InfNorm{ S_{\operatorname{IV}} - B^\sharp} + 2\InfNorm{ S_{\operatorname{III}}} + \InfNorm{S_{\operatorname{II}}} + \InfNorm{B^\sharp - B}. \label{SampleCovDecomp}
\end{align}
Let $r_{n, m} = r_1 + r_2 + 2 r_3 + r_4$. By parts I and II of Lemma \ref{LemmaEntrywiseSBModSel},
\begin{align*}
&P\Parens{ \InfNorm{ S(B, J_0, J_1) - B} \geq r_{n, m} } \\
&\leq P\Parens{ \InfNorm{ S_{\operatorname{IV}} - B^\sharp} + 2\InfNorm{ S_{\operatorname{III}}} + \InfNorm{S_{\operatorname{II}}} + \InfNorm{B^\sharp - B} \geq r_{n, m} } \quad \text{(by \eqref{SampleCovDecomp})} \\
&\leq P\Parens{ \InfNorm{ S_{\operatorname{IV}} - B^\sharp} + 2\InfNorm{ S_{\operatorname{III}}} + r_2 + r_1 \geq r_{n, m} } \quad \text{(by \eqref{entrywiseBoundSThreeSFour})} \\
&= P\Parens{ \InfNorm{ S_{\operatorname{IV}} - B^\sharp} + 2\InfNorm{ S_{\operatorname{III}}} \geq r_4 + 2r_3 }  \\
&\leq P\Parens{ \InfNorm{ S_{\operatorname{IV}} - B^\sharp} \geq r_4} + P\Parens{2\InfNorm{ S_{\operatorname{III}}} \geq 2r_3} \quad \text{(by \eqref{entrywiseProbs})} \\
&\leq \frac{2}{m^d} + \frac{2}{m^d} = \frac{4}{m^d}.
\end{align*}

We show that under the assumptions of Theorem \ref{mainTheoremFixedGenes}, the entrywise error in terms $S_{\operatorname{II}}$ and $S_{\operatorname{III}}$ is $O\Parens{C_A \sqrt{\frac{\log(m)}{m}}}$.  Recall that the entrywise rates of convergence of $\GlobalTwo$ and $\GlobalThree$ are stated in equations \eqref{r2Rate} and \eqref{r3Rate}, respectively.  Let $s = \abs{\text{supp}(\gamma)}$ denote the sparsity of $\gamma$.  Let $m_{01} = \abs{\text{supp}\Parens{ \gamma_{J_1} }}$ denote the number of false negatives.  

First, we express the entrywise rate of convergence of $S_{\operatorname{II}}$ in terms of $\tau_{\text{global}}$.  By \eqref{smallFalseNegatives}, $\InfNorm{\gamma_{J_1}} \leq \tau_{\text{global}}$, which implies that $\twonorm{\gamma_{J_1}}^2 \leq m_{01} \tau_{\text{global}}^2 \leq s \tau_{\text{global}}^2$, where the last inequality holds because $m_{01} \leq s$ by definition.  Therefore,     
\begin{equation} \label{r2BoundtauGlobal}
r_2 = (4 m)^{-1} \twonorm{\gammaGlobal}^2 \leq \frac{s \tau_{\text{global}}^2 }{4m} \leq C \frac{s \log(m)}{4 n m} \twonorm{B} ,
\end{equation}
where the last step holds because $\tau_{\text{global}} = C \sqrt{\log(m)} \lVert (D^T B^{-1}D)^{-1} \rVert_2^{1/2} \asymp \sqrt{\frac{\log(m)}{n}} \twonorm{B}^{1/2}$ by assumption.  Applying (A3) to the right-hand side of \eqref{r2BoundtauGlobal} implies that $r_2 = O\Parens{C_A \sqrt{\frac{\log(m)}{m}}}$.  

 Next, consider term $S_{\operatorname{III}}$.  First note that 
\begin{equation}
\gammaGlobal^T \AGlobal \gammaGlobal \leq \twonorm{\gammaGlobal}^2 \twonorm{\AGlobal} \leq \NumFixedFalseNeg \tau_{\text{global}}^2 \twonorm{\AGlobal},
\end{equation}
where the last inequality holds by \eqref{smallFalseNegatives}.  This implies that $r_3$ is on the order 
\begin{align} 
\frac{\log^{1/2}(m)}{m} \Parens{ \widecheck{b}_{\max} \gammaGlobal^T \AGlobal \gammaGlobal}^{1/2} &\leq \widecheck{b}_{\max}^{1/2} \twonorm{\AGlobal}^{1/2} \Parens{ \frac{\log^{1/2}(m) \NumFixedFalseNeg^{1/2}}{m} } \tau_{\text{global}} \notag  \\ 
&\leq C \frac{\log(m)}{\sqrt{n}} \frac{\sqrt{s}}{m} \twonorm{A_{J_1}}^{1/2} \twonorm{B}^{1/2} \widecheck{b}_{\max}^{1/2}, \label{entrywiseSThreeUnderA2primeBound}
\end{align}
where the last inequality holds because $m_{01} \leq s \leq m$ and $\tau_{\text{global}} \asymp \sqrt{\frac{\log(m)}{n}} \twonorm{B}^{1/2}$.  Under (A2'), the right-hand side of \eqref{entrywiseSThreeUnderA2primeBound} satisfies
\begin{equation}
\frac{\log(m)}{\sqrt{n}} \frac{\sqrt{s}}{m} \twonorm{A_{J_1}}^{1/2} \twonorm{B}^{1/2} \widecheck{b}_{\max}^{1/2} \leq \sqrt{\log(m)} \frac{\sqrt{s}}{m} C_A \frac{\twonorm{A_{J_1}}^{1/2} }{\twonorm{A}^{1/2} } \leq C_A \sqrt{ \frac{\log(m)}{m} }, 
\end{equation}
where the last inequality holds because $s \leq m$.

\subsection{Proof of part I of Lemma \ref{LemmaEntrywiseSBModSel}, term I}
\label{entrywiseTermI}

We bound the entrywise bias, 
\begin{align}
\MaxNorm{B^\sharp - B} &= \MaxNorm{ \frac{\Trace{A_{J_0}}}{m} \widetilde{B} + \frac{\Trace{A_{J_1}} }{m} \widecheck{B}  - B} \notag \\
% &= \MaxNorm{ \frac{\Trace{A_{J_0}}}{m}  \Parens{\widecheck{B} - B} + \frac{\Trace{A_{J_1}}}{m}  \Parens{\widetilde{B} - B}} \notag \\
&\leq \frac{\Trace{A_{J_0}}}{m} \MaxNorm{ \widetilde{B} - B}  + \frac{\Trace{A_{J_1}}}{m} \MaxNorm{\widecheck{B} - B}.\label{modSelBBiasMaxNorm}
\end{align}
% To bound the first term of \eqref{modSelBBiasMaxNorm}, note that
Note that 
\begin{align}
\MaxNorm{ \widecheck{B} - B } &= \MaxNorm{ (I - P_1) B (I - P_1) - B } =  \MaxNorm{P_1 B P_1 - P_1 B - BP_1} \notag \\
&\leq \MaxNorm{P_1 B P_1} + \MaxNorm{P_1 B} + \MaxNorm{BP_1}. \label{maxEntryBGlobalCen}
\end{align}
We bound the first term of \eqref{maxEntryBGlobalCen} as follows:
\begin{align*}
\abs{ \Parens{ P_1 B P_1 }_{ij} } &\leq \twonorm{p^{(1)}_i } \twonorm{ p^{(1)}_j } \twonorm{ B } \leq \frac{\twonorm{ B }}{n_{\min}}.
\end{align*}
For the second term of \eqref{maxEntryBGlobalCen},
\begin{align*}
\Parens{P_1 B}_{ij} &= \abs{ b_i^T p_j^{(1)} } \leq \OneNorm{ b_i } \InfNorm{p_j^{(1)}} \leq \OneNorm{ B } \InfNorm{p_j^{(1)}} \leq \frac{\OneNorm{ B }}{n_{\min}},
\end{align*}
where $\InfNorm{p_j^{(1)}} \leq \frac{1}{n_{\min}}$ by the definition of $P_1$ in \eqref{def:PGlobal}.  We have shown $\MaxNorm{ BP_1} \leq \frac{\OneNorm{ B }}{n_{\min}}$.  Likewise, $\MaxNorm{ B P_1 } \leq \frac{\OneNorm{ B }}{n_{\min}}$.  Therefore,
\begin{equation} \label{biasGlobalMaxNormNmin}
\MaxNorm{ \widecheck{B} - B} \leq 3 \frac{ \OneNorm{B} }{n_{\min}}.
\end{equation}

Because the projection matrix $P_2$ satisfies $\InfNorm{p_j^{(2)}} \leq \frac{1}{n_{\min}}$, an analogous proof shows that
\begin{equation} \label{biasGroupMaxNormNmin}
\MaxNorm{\widetilde{B} - B} \leq \frac{3 \OneNorm{B}}{n_{\min}}.
\end{equation}

Substituting \eqref{biasGlobalMaxNormNmin} and \eqref{biasGroupMaxNormNmin} into \eqref{modSelBBiasMaxNorm} yields
\begin{align}
\MaxNorm{B^\sharp - B} &\leq \frac{\Trace{A_{J_0}}}{m} \MaxNorm{ \widecheck{B} - B}  + \frac{\Trace{A_{J_1}}}{m} \MaxNorm{\widetilde{B} - B} \notag \\
&\leq \Parens{\frac{\Trace{A_{J_0}}}{m} + \frac{\Trace{A_{J_1}}}{m} } \frac{3 \OneNorm{B}}{n_{\min}} \notag \\
&= \frac{\Trace{A}}{m} \frac{3 \OneNorm{B}}{n_{\min}} \notag \\
&= \frac{3 \OneNorm{B}}{n_{\min}}.
\end{align}

\subsection{Proof of part I of Lemma \ref{LemmaEntrywiseSBModSel}, term II}
\label{entrywiseTermII}

In this section we prove a deterministic entrywise bound on $\GlobalTwo$.  By \eqref{ResdualsM}, it follows that
\begin{align*}
(\ResidGlobal) M_{J_1} M_{J_1}^T (\ResidGlobal) = (1/4) \twonorm{\gamma_{J_1}}^2 \GroupIndicators \GroupIndicators^T,
\end{align*}
which implies
\[
	\InfNorm{ (\ResidGlobal) M_{J_1} M_{J_1}^T (\ResidGlobal) } = \InfNorm{ (1/4) \twonorm{\gamma_{J_1}}^2 \GroupIndicators \GroupIndicators^T } = (1/4) \twonorm{\gamma_{J_1}}^2.
\]
Therefore $S_{\operatorname{II}}$ satisfies the maximum entrywise bound
\[
	\InfNorm{S_{\operatorname{II}}} = \InfNorm{ m^{-1} (\ResidGlobal) M_{J_1} M_{J_1}^T (\ResidGlobal) } = \InfNorm{ (4 m)^{-1} \twonorm{\gamma_{J_1}}^2 \GroupIndicators \GroupIndicators^T } = (4 m)^{-1} \twonorm{\gamma_{J_1}}^2,
\]
so
\[
	\InfNorm{S_{\operatorname{II}}} = r_2.
\]

Note that if $J_1$ is chosen so that $\InfNorm{\gamma_{J_1}} \leq \tau$, then $\twonorm{\gammaGlobal}^2 \leq \NumFixedFalseNeg \tau^2$, where $\NumFixedFalseNeg$ is the number of false negatives, so
\begin{equation} \label{boundTauGlobalThree}
\frac{\twonorm{\gamma_1}^2}{4 m} \leq \frac{\NumFixedFalseNeg}{4 m} \tau^2 \leq \frac{\tau^2}{4}.
\end{equation}
which implies that the entrywise rate of convergence of $S_{\operatorname{II}}$ is $O(\tau^2)$.

\subsection{Proof of part II of Lemma \ref{LemmaEntrywiseSBModSel}, term III} \label{entrywiseTermIII}

Let $p_i$ denote the $i$th column of $\PGlobal^T$, for $i = 1, \ldots, n$.  Let $m_k$ denote the $k$th column of $M$.  Let $\varepsilon_k$ denote the $k$th column of $\varepsilon$.  The term $\GlobalThree$ can be expressed as
\begin{align*}
(\GlobalThree)_{ij} &= m^{-1} (e_i - p_i)^T M_{J_1} \varepsilon_{J_1}^T (e_j - p_j) \\
&= m^{-1} \Trace{\varepsilon_{J_1}^T (e_j - p_j)(e_i - p_i)^T M_{J_1}} \\
&= m^{-1} \sum_{k \in J_1} \varepsilon_k^T (e_j - p_j)(e_i - p_i)^T m_k \\
&= m^{-1} \Vectorize{\varepsilon_{J_1}}^T \Parens{ \Identity{m_1} \otimes (e_j - p_j)(e_i - p_i)^T } \Vectorize{M_{J_1}} \\
&= m^{-1} \Vectorize{Z}^T \Parens{ A_{J_1}^{1/2} \otimes \TrueB^{1/2} (e_j - p_j)(e_i - p_i)^T } \Vectorize{M_{J_1}} \\
&= \Vectorize{Z}^T \VecGlobalTwo,
\end{align*}
where
\begin{equation} \label{def:VecGlobalTwo}
\VecGlobalTwo := m^{-1} \Parens{ A_{J_1}^{1/2} \otimes \TrueB^{1/2} (e_j - p_j)(e_i - p_i)^T } \Vectorize{M_{J_1}}.
\end{equation}

The squared Euclidean norm of $\VecGlobalTwo$ is
\begin{align}
\twonorm{\VecGlobalTwo}^2 &= \Vectorize{M_{J_1}}^T \Parens{ A_{J_1} \otimes (e_i - p_i) (e_j - p_j)^T \TrueB (e_j - p_j)(e_i - p_i)^T } \Vectorize{M_{J_1}} / m^2 \notag \\
&= \Vectorize{M_{J_1}}^T \Parens{A_{J_1} \otimes \widecheck{b}_{jj} (e_i - p_i) (e_i - p_i)^T} \Vectorize{M_{J_1}} / m^2 \notag \\
&= \widecheck{b}_{jj} \sum_{k \in J_1} \sum_{\ell \in J_1} \EntryTrueA{k \ell} m_k^T (e_i - p_i)(e_i - p_i)^T m_{\ell} / m^2 \notag \\
&= \widecheck{b}_{jj} \sum_{k \in J_1} \sum_{\ell \in J_1} \EntryTrueA{k \ell} (\GroupIndicators)_i \gamma_k (\GroupIndicators)_i \gamma_\ell / \Parens{4  m^2} \notag \\
&= \widecheck{b}_{jj} \sum_{k \in J_1} \sum_{\ell \in J_1} \EntryTrueA{k \ell} \gamma_k \gamma_\ell / \Parens{ 4 m^2} \notag \\
&= \widecheck{b}_{jj} \gammaGlobal^T A_{J_1} \gammaGlobal / \Parens{4 m^2}. \label{squaredNormPsiijRate3}
\end{align}

By the Hanson-Wright inequality (Theorem 2.1),
\begin{equation} \label{HansonWrightGlobalTwo}
\prob{ \abs{ \Vectorize{Z}^T \VecGlobalTwo - \twonorm{\VecGlobalTwo} } >  d^{1/2} K^2 \sqrt{\log(m)} \twonorm{\VecGlobalTwo} } \leq 2 \expf{ -d \log(m)} = 2 / m^d.
\end{equation}
Therefore
\begin{align*}
\prob{ \abs{ \EntryGlobalThree{ij} }  > \Parens{1 + d^{1/2} K^2 \sqrt{\log(m)} } \twonorm{\VecGlobalTwo} } &= \prob{ \abs{ \Vectorize{Z}^T \VecGlobalTwo } > \twonorm{\VecGlobalTwo} +  d^{1/2} K^2 \sqrt{\log(m)} \twonorm{\VecGlobalTwo}} \\
&\leq \prob{ \abs{ \Vectorize{Z}^T \VecGlobalTwo - \twonorm{\VecGlobalTwo} } > d^{1/2} K^2 \sqrt{\log(m)} \twonorm{\VecGlobalTwo} } \\
&\leq 2/m^d,
\end{align*}
where the last step follows from \eqref{HansonWrightGlobalTwo}.  By \eqref{squaredNormPsiijRate3}, it follows that 
\begin{equation} \label{r3UpperBoundEntrywise}
\Parens{ 1 + d^{1/2} K^2 \sqrt{\log(m)} } \twonorm{\VecGlobalTwo} \leq r_3, 
\end{equation} 
so 
\begin{equation} 
\prob{ \abs{ \EntryGlobalThree{ij} }  >  r_3 } \leq \prob{ \abs{ \EntryGlobalThree{ij} }  > \Parens{1 + d^{1/2} K^2 \sqrt{\log(m)} } \twonorm{\VecGlobalTwo} } \leq 2 / m^d, 
\end{equation} 
by \eqref{r3UpperBoundEntrywise}.  By the union bound, 
\begin{equation*} 
\prob{ \InfNorm{ S_{\operatorname{III}} } > r_3 } \leq \sum_{i = 1}^m \sum_{j = 1}^m  \prob{ \abs{ \EntryGlobalThree{ij} }  >  r_3 } \leq 2 / m^{d - 2}.  
\end{equation*} 

\subsection{Proof of part II of Lemma \ref{LemmaEntrywiseSBModSel}, term IV}
\label{entrywiseTermIV}

We now analyze term $S_{\operatorname{IV}}$.  To do so, we express $S_{\text{IV}}$ as a quadratic form in order to apply the Hanson-Wright inequality.

Let $p_i^{(1)}$ denote the $i$th column of $\PGlobal^T$.  Let $p_i^{(2)}$ denote the $i$th column of $\PGroup^T$.  Define
\begin{equation}
H_{\operatorname{group}}^{ij} = I_{m_0} \otimes \Parens{ e_j - p_j^{(2)} } \Parens{ e_j - p_j^{(2)} } ^T \qquad \text{ and } \quad H_{\operatorname{global}}^{ij} = I_{m_1} \otimes \Parens{ e_j - p_j^{(1)} } \Parens{ e_j - p_j^{(1)} } ^T,
\end{equation}
and let
\begin{equation}
H^{ij}(J_0, J_1) = \begin{bmatrix}
H_{\operatorname{group}}^{ij} \\
& H_{\operatorname{global}}^{ij}
\end{bmatrix},
\end{equation}
where $H_{\operatorname{group}}^{ij} \in \mathbb{R}^{m_0 n \times m_0 n}$, $H_{\operatorname{global}}^{ij} \in \mathbb{R}^{m_1 n \times m_1 n}$, and $H^{ij}(J_0, J_1) \in \mathbb{R}^{mn \times mn}$.  Recall that
\begin{equation*}
S_{\operatorname{IV}} = m^{-1} (I - P_2) \varepsilon_{J_0} \varepsilon_{J_0}^T (I - P_2) + m^{-1} (I - P_1) \varepsilon_{J_1} \varepsilon_{J_1}^T (I - P_1).
\end{equation*}
The second term of $S_{\operatorname{IV}}$ can be expressed as a quadratic form, as follows (where $\varepsilon_k$ denotes the $k$th column of $\varepsilon \in \mathbb{R}^{n \times m}$):
\begin{align}
m^{-1} (I - P_1) \varepsilon_{J_1} \varepsilon_{J_1}^T (I - P_1) &= m^{-1} \sum_{k \in J_1} \Parens{ e_i - p_i^{(1)} }^T \varepsilon_k \varepsilon_k^T \Parens{ e_j - p_j^{(1)} }  \notag \\
&= m^{-1} \sum_{k \in J_1} \Trace{ \Parens{ e_i - p_i^{(1)} }^T \varepsilon_k \varepsilon_k^T \Parens{ e_j - p_j^{(1)} } } \notag \\
&= m^{-1} \sum_{k \in J_1} \varepsilon_k^T \Parens{ e_j - p_j^{(1)} } \Parens{ e_i - p_i^{(1)} }^T \varepsilon_k \notag \\
&= m^{-1} \Vectorize{\varepsilon_{J_1}}^T \Parens{ I_{m_1} \otimes \Parens{ e_j - p_j^{(1)} } \Parens{ e_i - p_i^{(1)} } ^T } \Vectorize{\varepsilon_{J_1}}^T \notag \\
&= m^{-1} \Vectorize{\varepsilon_{J_1}}^T H_{\operatorname{global}}^{ij} \Vectorize{\varepsilon_{J_1}}^T. \label{termOneGlobalQuadFrom}
% &= m^{-1} \Vectorize{Z_{\widetilde{J}_0}}^T \Parens{A \otimes B^{1/2} H_{\operatorname{global}}^{ij} B^{1/2}} \Vectorize{Z_{\widetilde{J}_0}}^T
\end{align}

Analogously, the first term of $S_{\operatorname{IV}}$ can be expressed as a quadratic form:
\begin{align}
m^{-1} (I - P_2) \varepsilon_{J_0} \varepsilon_{J_0}^T (I - P_2) &= m^{-1} \sum_{k \in J_0} \Parens{ e_i - p_i^{(2)} }^T \varepsilon_k \varepsilon_k^T \Parens{ e_j - p_j^{(2)} } \notag \\
&= m^{-1} \Vectorize{\varepsilon_{J_0}}^T H_{\operatorname{group}}^{ij} \Vectorize{\varepsilon_{J_0}}^T. \label{termOneGroupQuadFrom}
\end{align}

We now express $S_{\operatorname{IV}}$ as a quadratic form.  Let $\pi(X)$ denote the matrix $X$ with reordered columns:
\begin{equation} \label{permuted}
\pi( X ) = \begin{bmatrix}
X_{J_0} & X_{J_1}
\end{bmatrix}
\quad \text{ and } \quad
\pi(A) = \operatorname{Cov}\Parens{ \Vectorize{\pi(X)}}.
\end{equation}
Then by \eqref{termOneGlobalQuadFrom} and \eqref{termOneGroupQuadFrom},
\begin{align*}
(S_{\operatorname{IV}})_{ij} &= m^{-1} \Vectorize{\varepsilon_{J_0}}^T H_{\operatorname{group}}^{ij} \Vectorize{\varepsilon_{J_0}}^T + m^{-1} \Vectorize{\varepsilon_{J_1}}^T H_{\operatorname{global}}^{ij} \Vectorize{\varepsilon_{J_1}}^T \\
&= m^{-1}  \Vectorize{\pi \Parens{ \varepsilon }}^T H^{ij}(J_0, J_1) \Vectorize{\pi \Parens{ \varepsilon }} \\
&= m^{-1}  \Vectorize{Z}^T \Parens{ \Parens{ \pi(A)^{1/2} \otimes B^{1/2}} H^{ij}(J_0, J_1) \Parens{ \pi(A)^{1/2} \otimes B^{1/2}}} \Vectorize{Z},
\end{align*}
where the last step holds by decorrelation, with $Z \in \mathbb{R}^{n \times m}$ as a random matrix with independent subgaussian entries.

Note that the $(i, j)$th entry of $S_{\operatorname{IV}}$ can be expressed as
\begin{equation} \label{SOneQuadraticForm}
\Parens{S_{\operatorname{IV}}}_{ij} = \Vectorize{Z}^T \Phi_{i, j} \Vectorize{Z},
\end{equation}
with
\begin{equation} \label{DecorrHij}
\Phi_{i, j} = m^{-1} \Parens{ \pi(A)^{1/2} \otimes B^{1/2}} H^{ij}(J_0, J_1) \Parens{ \pi(A)^{1/2} \otimes B^{1/2}}.
\end{equation}
Having expressed $\Parens{S_{\operatorname{IV}}}_{ij}$ as a quadratic form in \eqref{SOneQuadraticForm}, we find the trace and Frobenius norm of $\Phi_{i, j}$, then apply the Hanson-Wright inequality.  First we find the trace of $\Phi_{i, j}$.  Let
\begin{equation}
\mathcal{I}_0 = \begin{bmatrix} I_{m_0 \times m_0} & 0_{m_0 \times m_1} \\
0_{m_1 \times m_0} & 0_{m_1 \times m_1} \end{bmatrix} \quad \text{ and } \quad \mathcal{I}_1 = \begin{bmatrix} 0_{m_0 \times m_0} & 0_{m_0 \times m_1} \\
0_{m_1 \times m_0} & I_{m_1 \times m_1} \end{bmatrix}.
\end{equation}

Note that $H^{ij}(J_0, J_1)$ can be written as a sum of Kronecker products,
\begin{equation} \label{HijSumKronecker}
H^{ij}(J_0, J_1) = \mathcal{I}_0 \otimes \Parens{ e_j - p_j^{(2)} } \Parens{ e_i - p_i^{(2)} } ^T + \mathcal{I}_1 \otimes \Parens{ e_j - p_j^{(1)} } \Parens{ e_i - p_i^{(1)} }^T,
\end{equation}
hence \eqref{DecorrHij} can be expressed as
\begin{align}
&m^{-1} \Parens{ \pi(A)^{1/2} \otimes B^{1/2}} \Parens{\mathcal{I}_0 \otimes \Parens{ e_j - p_j^{(2)} } \Parens{ e_i - p_i^{(2)} } ^T } \Parens{ \pi(A)^{1/2} \otimes B^{1/2}}  \label{DecorrHijTermGlobal} \\
&\qquad + m^{-1} \Parens{ \pi(A)^{1/2} \otimes B^{1/2}} \Parens{\mathcal{I}_1 \otimes \Parens{ e_j - p_j^{(1)} } \Parens{ e_i - p_i^{(1)} }^T }  \Parens{ \pi(A)^{1/2} \otimes B^{1/2}} \label{DecorrHijTermGroup}.
\end{align}
The trace of the term \eqref{DecorrHijTermGlobal} is
\begin{align*}
& m^{-1} \Trace{\Parens{\pi(A)^{1/2} \otimes B^{1/2}} \Parens{\mathcal{I}_0 \otimes \Parens{ e_j - p_j^{(2)} } \Parens{ e_i - p_i^{(2)} } ^T} \Parens{\pi(A)^{1/2} \otimes B^{1/2}}} \\
&= m^{-1} \Trace{ \pi(A)^{1/2} \mathcal{I}_0 \pi(A)^{1/2} \otimes B^{1/2} \Parens{ e_j - p_j^{(2)} } \Parens{ e_i - p_i^{(2)} }^T B^{1/2}} \\
&= m^{-1} \Trace{\pi(A)^{1/2} \mathcal{I}_0 \pi(A)^{1/2}} \Trace{B^{1/2} \Parens{ e_j - p_j^{(2)} } \Parens{ e_i - p_i^{(2)} }^T B^{1/2} } \\
&= m^{-1} \Trace{\mathcal{I}_0 \pi(A)} \Parens{ \Parens{ e_i - p_i^{(2)} }^T B \Parens{ e_j - p_j^{(2)} } } \\
&= m^{-1} \Trace{A_{J_0} } \Brackets{ (I - P_2) B (I - P_2)) }_{ij} \\
&= m^{-1} \Trace{A_{J_0} } \widetilde{b}_{ij}.
\end{align*}
Analogously, the trace of the term \eqref{DecorrHijTermGroup} is
\begin{align*}
&m^{-1} \Trace{\Parens{\pi(A)^{1/2} \otimes B^{1/2}} \Parens{\mathcal{I}_1 \otimes \Parens{ e_j - p_j^{(1)} } \Parens{ e_i - p_i^{(1)} } ^T} \Parens{\pi(A)^{1/2} \otimes B^{1/2}}} \\
&= m^{-1} \Trace{A_{J_1}} \Brackets{ (I - P_1) B (I - P_1)) }_{ij} \\
&= m^{-1} \Trace{A_{J_1}} \widecheck{b}_{ij}.
\end{align*}
Let $b^\sharp_{ij}$ denote the $(i, j)$th entry of $B^\sharp$ defined in \eqref{expectedValueModSelSB}.  We have shown that the trace of $\Phi_{i, j}$ (as defined in \eqref{DecorrHij}) is
\begin{equation} \label{traceQuadFromTermOne}
\Trace{\Phi_{i, j}} = m^{-1} \Trace{A_{J_0} } \widetilde{b}_{ij} + m^{-1} \Trace{A_{J_1}} \widecheck{b}_{ij} = b^\sharp_{ij}.
\end{equation}

Next, we find the Frobenius norm of $\Phi_{i, j}$.  For convenience, define
\begin{align}
\mathcal{A}_{0} = \pi(A)^{1/2} \mathcal{I}_0 \pi(A)^{1/2} \quad &\text{ and } \quad \mathcal{A}_{1} = \pi(A)^{1/2} \mathcal{I}_1 \pi(A)^{1/2} \\
\mathcal{B}_{2, ij}  = B^{1/2} \Parens{ e_j - p_j^{(2)} } \Parens{ e_i - p_i^{(2)} }^T B^{1/2} \quad &\text{ and } \quad \mathcal{B}_{1, ij}  = B^{1/2} \Parens{ e_j - p_j^{(1)} } \Parens{ e_i - p_i^{(1)} }^T B^{1/2}.
\end{align}
Then
\begin{align}
\FroNorm{ \Phi_{i, j} }^2 &= \FroNorm{ m^{-1} \Parens{ \pi(A)^{1/2} \otimes B^{1/2}} H^{ij}(J_0, J_1) \Parens{ \pi(A)^{1/2} \otimes B^{1/2}} }^2 \notag \\
&= m^{-2} \FroNorm{ \mathcal{A}_0 \otimes \mathcal{B}_{2, ij}  + \mathcal{A}_1 \otimes \mathcal{B}_{1, ij}  }^2 \notag \\
&= m^{-2} \Trace{\Parens{\mathcal{A}_0 \otimes \mathcal{B}_{2, ij}  + \mathcal{A}_1 \mathcal{B}_{1, ij} }^T \Parens{\mathcal{A}_0 \otimes \mathcal{B}_{2, ij}  + \mathcal{A}_1 \otimes \mathcal{B}_{1, ij} } } \notag \\
&= m^{-2} \Trace{ \mathcal{A}_0^T \mathcal{A}_0 \otimes \mathcal{B}_{2, ij} ^T \mathcal{B}_{2, ij}   } +  m^{-2} \Trace{ \mathcal{A}_1^T \mathcal{A}_1 \otimes \mathcal{B}_{1, ij} ^T \mathcal{B}_{1, ij}   } \notag \\
&\qquad + m^{-2} \Trace{ \mathcal{A}_0^T \mathcal{A}_1 \otimes \mathcal{B}_{2, ij} ^T \mathcal{B}_{1, ij}   } +  m^{-2} \Trace{ \mathcal{A}_1^T \mathcal{A}_0 \otimes \mathcal{B}_{1, ij} ^T \mathcal{B}_{2, ij}  }. \label{TermOneFroDecomp}
\end{align}
We now find the traces of each of the terms in \eqref{TermOneFroDecomp}.  First, note that
\begin{align} \label{A0A0}
\Trace{ \mathcal{A}_0^T \mathcal{A}_0 } &= \Trace{ \mathcal{I}_0 \pi(A) \mathcal{I}_0 \pi(A) } = \Trace{ A_{J_0}^2 } = \FroNorm{ A_{J_0} }^2.
\end{align}
Analogously,
\begin{equation} \label{A1A1}
\Trace{ \mathcal{A}_1^T \mathcal{A}_1} = \FroNorm{ A_{J_1} }^2.
\end{equation}
For the cross-term, let $A_{J_0 J_1}$ denote the $m_0 \times m_1$ submatrix of $\pi(A)$ given by columns of $A$ in $J_0$ and rows of $A$ in $J_1$.  Then
\begin{align}
\Trace{ \mathcal{A}_0^T \mathcal{A}_1} &= \Trace{ \mathcal{I}_0 \pi(A) \mathcal{I}_1 \pi(A) } \notag \\
&= \Trace{ \begin{bmatrix} 0_{m_0 \times m_0} & A_{J_0 J_1} \\ 0_{m_1 \times m_0} & 0_{m_1 \times m_1} \end{bmatrix} \pi(A) } \notag \\
&= \Trace{ A_{J_0 J_1}^T A_{J_0 J_1}} \notag \\
&= \FroNorm{ A_{J_0 J_1}}^2. \label{A0A1}
\end{align}

Next,
\begin{align}
\Trace{ \mathcal{B}_{1, ij} ^T \mathcal{B}_{1, ij}  } &= \Trace{B^{1/2} \Parens{ e_i - p_i^{(1)} } \Parens{ e_j - p_j^{(1)} }^T B\Parens{ e_j - p_j^{(1)} } \Parens{ e_i - p_i^{(1)} }^T B^{1/2} } \notag \\
&= \Parens{ \Parens{ e_j - p_j^{(1)} }^T B\Parens{ e_j - p_j^{(1)} } } \Parens{ \Parens{ e_i - p_i^{(1)} }^T B\Parens{ e_i - p_i^{(1)} } } \notag \\
&= \widecheck{b}_{jj} \widecheck{b}_{ii}. \label{B1B1}
\end{align}
Analogously,
\begin{align}
\Trace{ \mathcal{B}_{2, ij} ^T \mathcal{B}_{2, ij}  } &= \Parens{ \Parens{ e_j - p_j^{(2)} }^T B\Parens{ e_j - p_j^{(2)} } } \Parens{ \Parens{ e_i - p_i^{(2)} }^T B\Parens{ e_i - p_i^{(2)} } } \notag \\
&= \widetilde{b}_{jj} \widetilde{b}_{ii}. \label{B2B2}
\end{align}
The cross-terms yield
\begin{align} \label{B1B2}
\Trace{ \mathcal{B}_{1, ij} ^T \mathcal{B}_{2, ij}  } &= \Parens{ \Parens{ e_j - p_j^{(1)} }^T B\Parens{ e_j - p_j^{(2)} } } \Parens{ \Parens{ e_i - p_i^{(2)} }^T B\Parens{ e_i - p_i^{(1)} } } = \breve{b}_{ii} \breve{b}_{jj}.
\end{align}

The squared Frobenius norm of $\Phi_{i, j}$ is
\begin{align*}
\FroNorm{\Phi_{i, j}}^2 &= \frac{1}{m^2} \Parens{ \FroNorm{A_{J_0}}^2 \widecheck{b}_{ii} \widecheck{b}_{jj} + \FroNorm{A_{J_1}}^2 \widetilde{b}_{ii} \widetilde{b}_{jj} + 2 \FroNorm{A_{J_0, J_1}}^2 \breve{b}_{ii} \breve{b}_{jj} } \\
&\leq \frac{1}{m^2} C \Parens{ \FroNorm{A_{J_0}}^2 + \FroNorm{A_{J_1}}^2 + 2 \FroNorm{A_{ J_0 J_1 }}^2 } \twonorm{B}^2 \\
&= C \frac{1}{m^2} \FroNorm{A}^2 \twonorm{B}^2.
\end{align*}

We now apply the Hanson-Wright inequality,
\begin{align*}
\prob{ \abs{ \EntryGlobalOne{ij} -  b^\sharp_{ij} } > r_4} &= \prob{ \abs{ \Vectorize{Z}^T \QuadraticFormGlobalCen \Vectorize{Z} - \Trace{\QuadraticFormGlobalCen} } > r_4} \\
&\leq 2 \exp\left( -c \min\left\{ d \log(m) , d^{1/2} \sqrt{\log(m)} \frac{\fnorm{\QuadraticFormGlobalCen} }{ \twonorm{\QuadraticFormGlobalCen} } \right\}  \right) \\
&\leq 2 \max\left( m^{-d} , \exp\left( d^{1/2} \sqrt{\log(m)} r^{1/2}(\QuadraticFormGlobalCen) \right) \right).
\end{align*}
The first step holds by \eqref{SOneQuadraticForm} and \eqref{traceQuadFromTermOne}.

%!TEX root = submit.arxiv.tex

\section{Comparisons to related methods} \label{secComparison}

The most similar existing method to ours is the sphering approach from \citet{allen2012inference}. Both methods use a preliminary demeaned version of the data to generate covariance estimates, then use these estimates to adjust the gene-wise $t$-tests. The largest difference between the procedures lies in this last step. The sphering approach produces an adjusted data set based on decorrelating residuals from a preliminary mean estimate and performs testing and mean estimation on this adjusted data using traditional OLS techniques. Though their approach is well-motivated at the population level, they do not provide theoretical support for their plug-in procedure, and in particular do not explore how noise in the initial mean estimate may complicate their decorrelation procedure. In contrast, our approach uses a generalized least squares approach motivated by classical statistical results including the Gauss Markov theorem.

The sphering approach also involves decorrelating a data matrix along both axes.  Our work, including the theoretical analysis in \citet{Zhou14a}, suggests that when the data matrix is non-square, attempting to decorrelate along the longer axis generally degrades performance. For genetics applications, where there are usually many more genes than samples, this suggests that decorrelating along the genes may hurt the performance of the sphering method. Fortunately, for gene-level analyses it is not necessary to decorrelate along the gene axis, since inference methods like false discovery rate are robust to a certain level of dependence among the variables (genes) ~\citep{benjamini2001control}. Therefore, we also consider a modification of the sphering algorithm that only decorrelates along the samples.

Confounder adjustment is another related line of work that deals with similar issues when attempting to discover mean differences. In particular, a part of that literature posits models where row-wise connections arise from the additive effects of potential latent variables. \citet{sun2012multiple} and \citet{wang2015confounder} use models of the form
\begin{align*}
X_{n \times m} &= D_{n \times 1} \beta_{m \times 1}^T  + Z_{n \times r} \Gamma_{m \times r}^T + E_{n \times m} \\
Z_{n \times r} &= D_{n \times 1} \alpha_{r \times 1}^T + W_{n \times r}
\end{align*}
where $Z$ is an unobserved matrix of $r$ latent factors. Rewriting these equations into the following form lets us better contrast the confounder model to our matrix-variate setup:
\begin{equation} \label{cateModelFactoredForm}
X = D (\beta + \Gamma \alpha)^T + W \Gamma^T + E.
\end{equation}
These models are generally estimated by using some form of factor analysis to estimate $\Gamma$ and then using regression methods with additive outlier detection to identify $\beta$, methodology that is quite different from our GLS-based methods.

For the two-group model, in the case of a globally centered data matrix $X$, the design matrix $D$ in \eqref{cateModelFactoredForm} takes the form
\begin{equation} \label{cateModelGloballyCentered}
D_{n \times 1}^T = \begin{bmatrix} -1 & \cdots & -1 & 1 \cdots & 1 \end{bmatrix} = \begin{bmatrix} -1_{n_1}^T & 1_{n_2}^T \end{bmatrix},
\end{equation}
and $2\beta$ represents the vector of true mean differences between the groups.  The vector $\beta$ is estimated via OLS, yielding $\hat{\beta}_{\text{OLS}}$, and CATE considers whether the residual $X - D_{n \times 1} \hat{\beta}_{\text{OLS}}$ has a low-rank covariance structure plus noise.  If so, $\hat{\Gamma} \hat \alpha$
aims to take out the residual low-rank structure
through $D (\widehat{\Gamma \alpha})^T$.
As illustrated in simulation and data analysis, this improves upon inference based only on $\hat{\beta}_\text{OLS}$.
 When applying the CATE and related methods to data originated from the generative model as described in the present paper,
CATE (and in particular, the related LEAPP) method essentially seeks a sparse approximation of $\hat{\beta}_\text{OLS}$; Moreover in LEAPP, this is essentially achieved via hard thresholding of coefficients of  $\hat{\beta}_\text{OLS}$, leading to improvements in  performance  in variable selection and its subsequence inference when the vector of true mean differences is presumed to be sparse. In our setting, we improve upon OLS using GLS.

\subsection{Simulation results}

\begin{figure}[tbp]
\includegraphics[width=\textwidth]{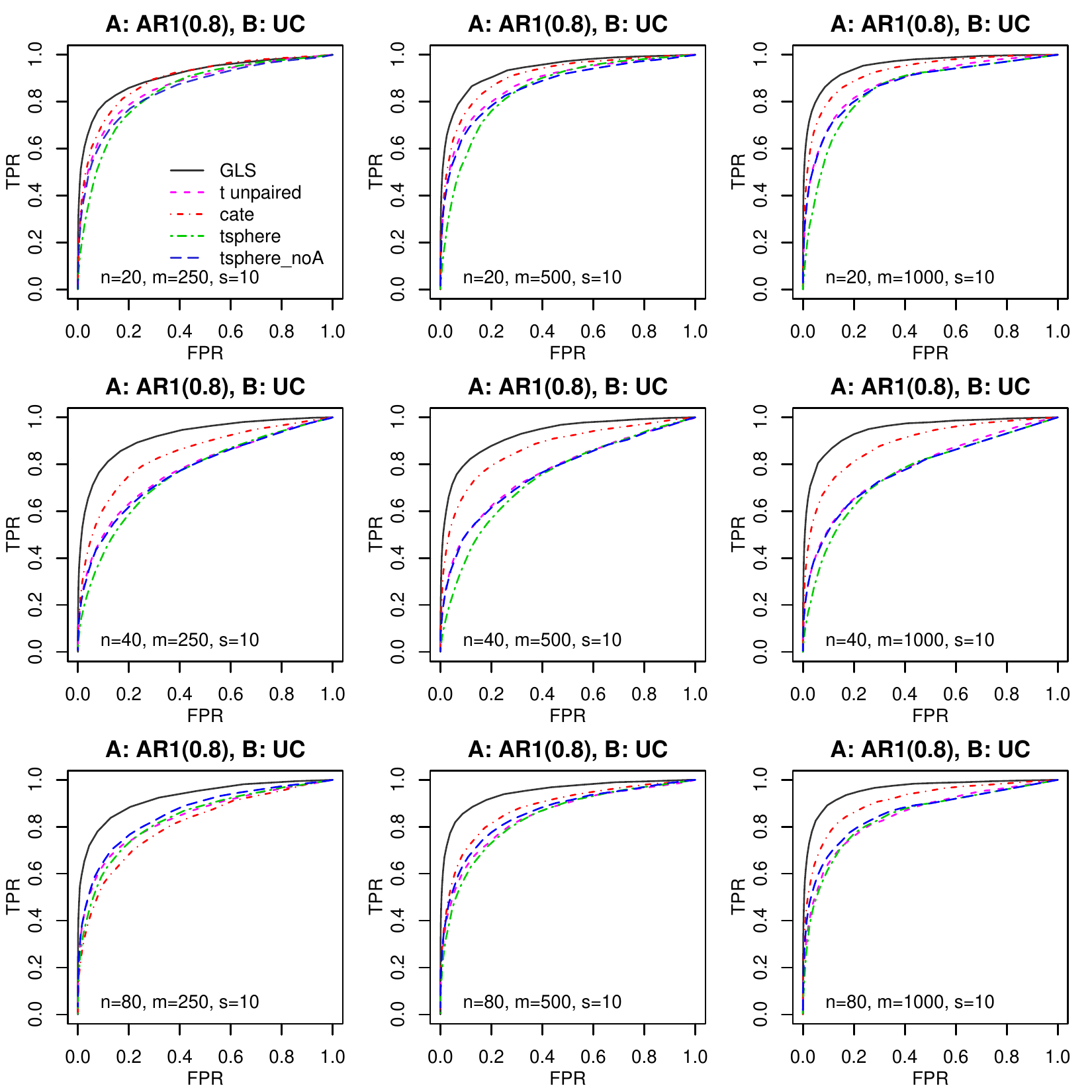}
\caption{Performance of Algorithm 2 (GLS) relative to sphering and confounder adjustment methods, labeled as \texttt{tsphere} and \texttt{cate}, respectively. These are ROC curves for identifying true mean differences. An implementation of the sphering algorithm that does not adjust for $A$ is also included, labeled as \texttt{tsphere\_noA}. Each panel shows the average ROC curves over 200 simulations. We simulate matrix variate data with gene correlations from an $\text{AR1}(0.8)$ model and let $s=10$ genes have true mean differences of $0.8$, $0.6$, and $0.4$ for the first, second and third rows, respectively. For all of these the true $B$ is set to $\widehat{B}$ from the ulcerative colitis data (using a repeated block structure for larger $n$ values), described in Section~\ref{sec::UCData} and evenly-sized groups are assigned randomly.}
\label{fig::roc_comparison_UC}
\end{figure}

\begin{figure}[htbp]
\includegraphics[width=\textwidth]{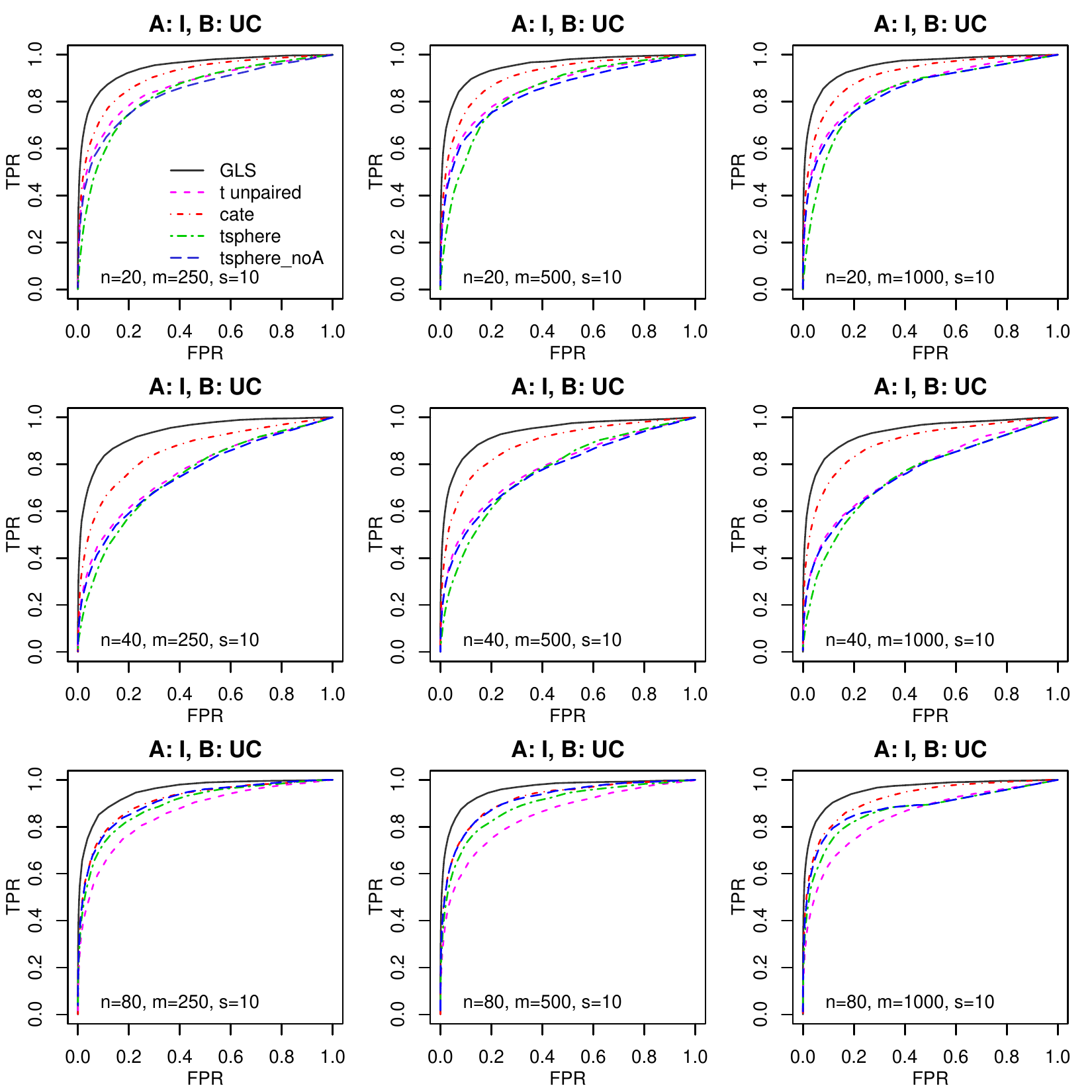}
\caption{Performance of Algorithm 2 (GLS) relative to sphering and confounder adjustment, labeled as \texttt{tsphere} and \texttt{cate}, respectively. These are ROC curves for identifying true mean differences. An implementation of the sphering algorithm that does not adjust for $A$ is also included, labeled as \texttt{tsphere\_noA}. Each panel shows the average ROC curves over 200 simulations. We simulate matrix variate data with no gene-wise correlations ($A=I$) and let $s=10$ genes have true mean differences of $0.8$, $0.6$, and $0.4$ for the first, second and third rows, respectively. For all of these the true $B$ is set to $\widehat{B}$ from the ulcerative colitis data (using a repeated block structure for larger $n$ values), described in Section~\ref{sec::UCData} and evenly-sized groups are assigned randomly.}
\label{fig::roc_comparison_UC_baseline}
\end{figure}

Figure~\ref{fig::roc_comparison_UC} compares the performance of Algorithm 2 to the sphering method of \citet{allen2012inference} and the robust regression confounder adjustment method of \citet{wang2015confounder} on simulated matrix variate data motivated by the ulcerative colitis dataset described in Section~\ref{sec::UCData}. Note that this robust regression confounder adjustment is a minor modification of the LEAPP algorithm introduced in \citet{sun2012multiple}. As discussed above, we also consider a modification of \citet{allen2012inference} that only decorrelates along the rows.

\begin{figure}[htb]
\includegraphics[width=\textwidth]{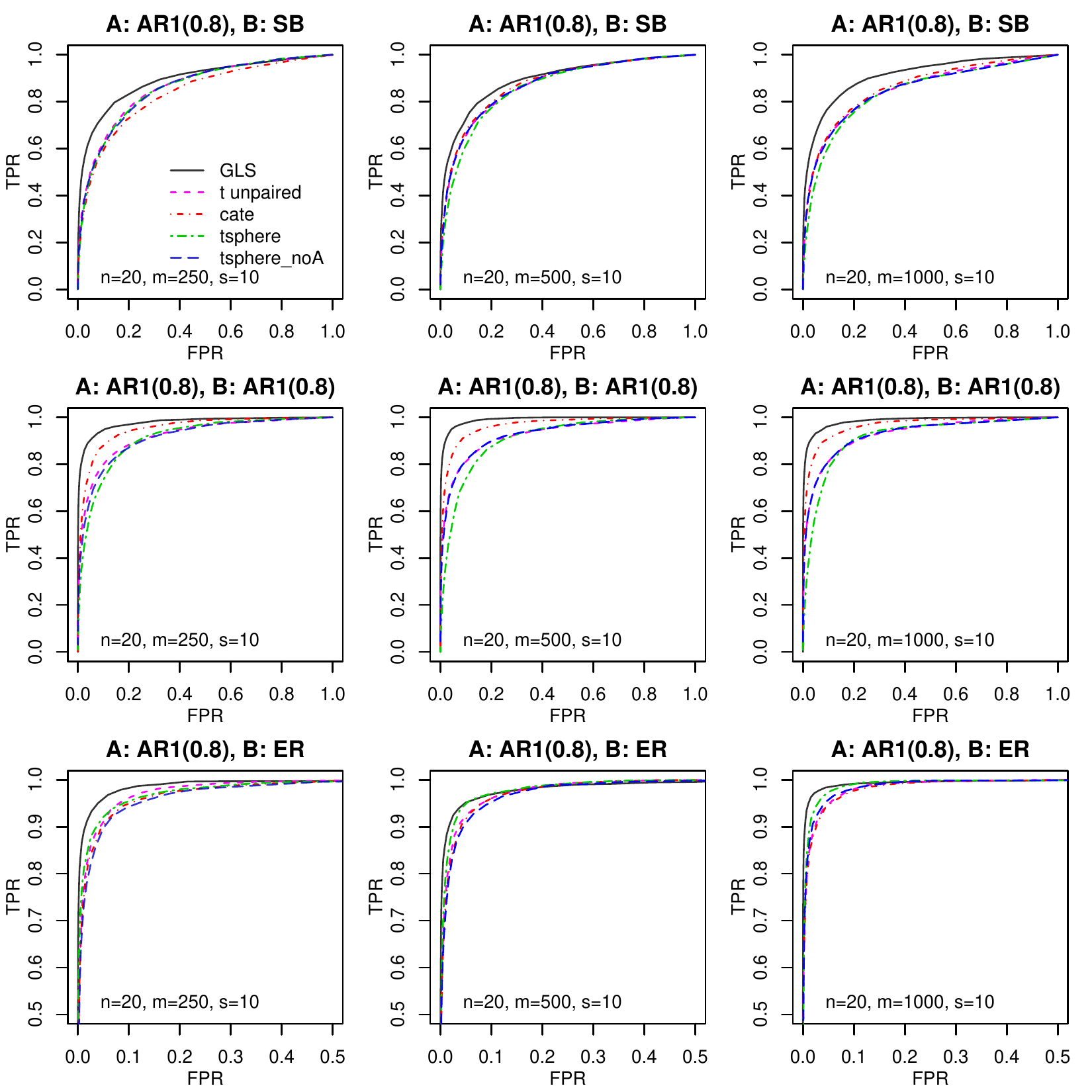}
\caption{Performance of Algorithm 2 (GLS) relative to sphering and confounder adjustment, labeled as \texttt{tsphere} and \texttt{cate}, respectively. These are ROC curves for identifying true mean differences. An implementation of the sphering algorithm that does not adjust for $A$ is also included, labeled as \texttt{tsphere\_noA}. Each panel shows the average ROC curves over 200 simulations. We simulate matrix variate data with an $\text{AR1}(0.8)$ model for $A$ and let $s=10$ genes have true mean differences of $0.8$. $B$ is constructed according to a Star-Block model with blocks of size 4, an $\text{AR1}(0.8)$, and an Erd\H{o}s-R\'{e}nyi random graph with $d=n \log n$ edges. All of these use $n=20$ and randomly assign 10 observations to each group.}
\label{fig::roc_comparison_other}
\end{figure}

\begin{figure}[htb]
\includegraphics[width=\textwidth]{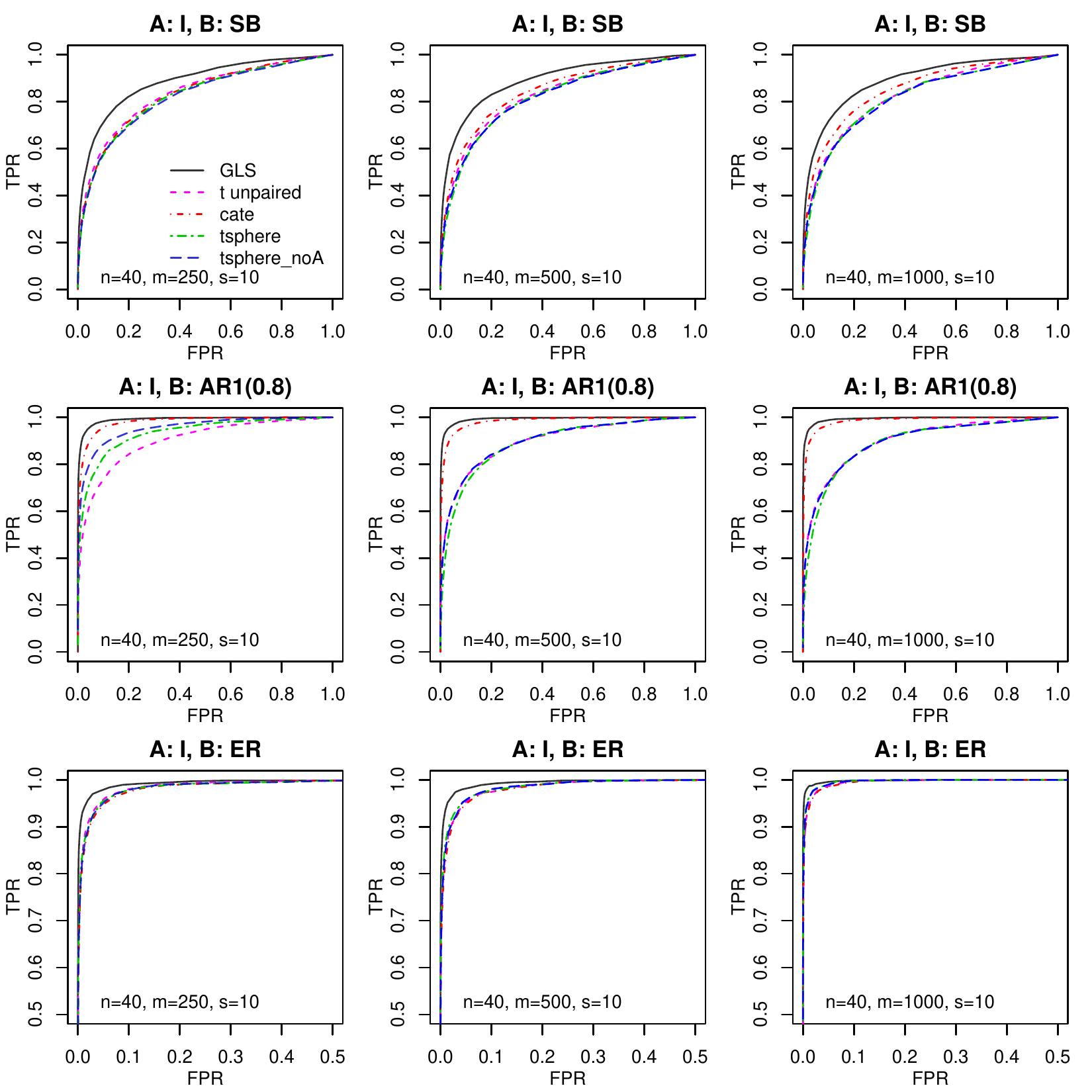}
\caption{Performance of Algorithm 2 (GLS) relative to sphering and confounder adjustment, labeled as \texttt{tsphere} and \texttt{cate}, respectively. These are ROC curves for identifying true mean differences. An implementation of the sphering algorithm that does not adjust for $A$ is also included, labeled as \texttt{tsphere\_noA}. Each panel shows the average ROC curves over 200 simulations. We simulate matrix variate data with no gene-wise correlations ($A=I$) and let $s=10$ genes have true mean differences of $0.6$. $B$ is constructed according to a Star-Block model with blocks of size 4, an $\text{AR1}(0.8)$, and an Erd\H{o}s-R\'{e}nyi random graph with $d=n \log n$ edges. All of these use $n=40$ and randomly assign 20 observations to each group.}
\label{fig::roc_comparison_I_other}
\end{figure}

We can see that across a range of dataset sizes our method consistently outperforms sphering in terms of sensitivity and specificity for identifying mean differences.  In some settings, CATE improves on Tsphere and $t$-statistics despite being applied on misspecified models, because CATE takes out the additional rank two structure from the mean after OLS regression and does some approximate thresholding on the coefficients.  Our method using GLS performs significantly better than CATE in the setting of non-identity $B$, with edges present both within and between groups.

Figure~\ref{fig::roc_comparison_other} fixes the sample size and repeats these comparisons on different sample correlation structures (which are described in Section~\ref{sec::Simulations}).  Figure~\ref{fig::roc_comparison_I_other} is analogous to Figure~\ref{fig::roc_comparison_other}, but with $A$ as the identity matrix.  Algorithm 2 is competitive or superior to the competing methods across a range of topologies.

\begin{figure}[tb] \centering
\includegraphics[width=0.65\textwidth]{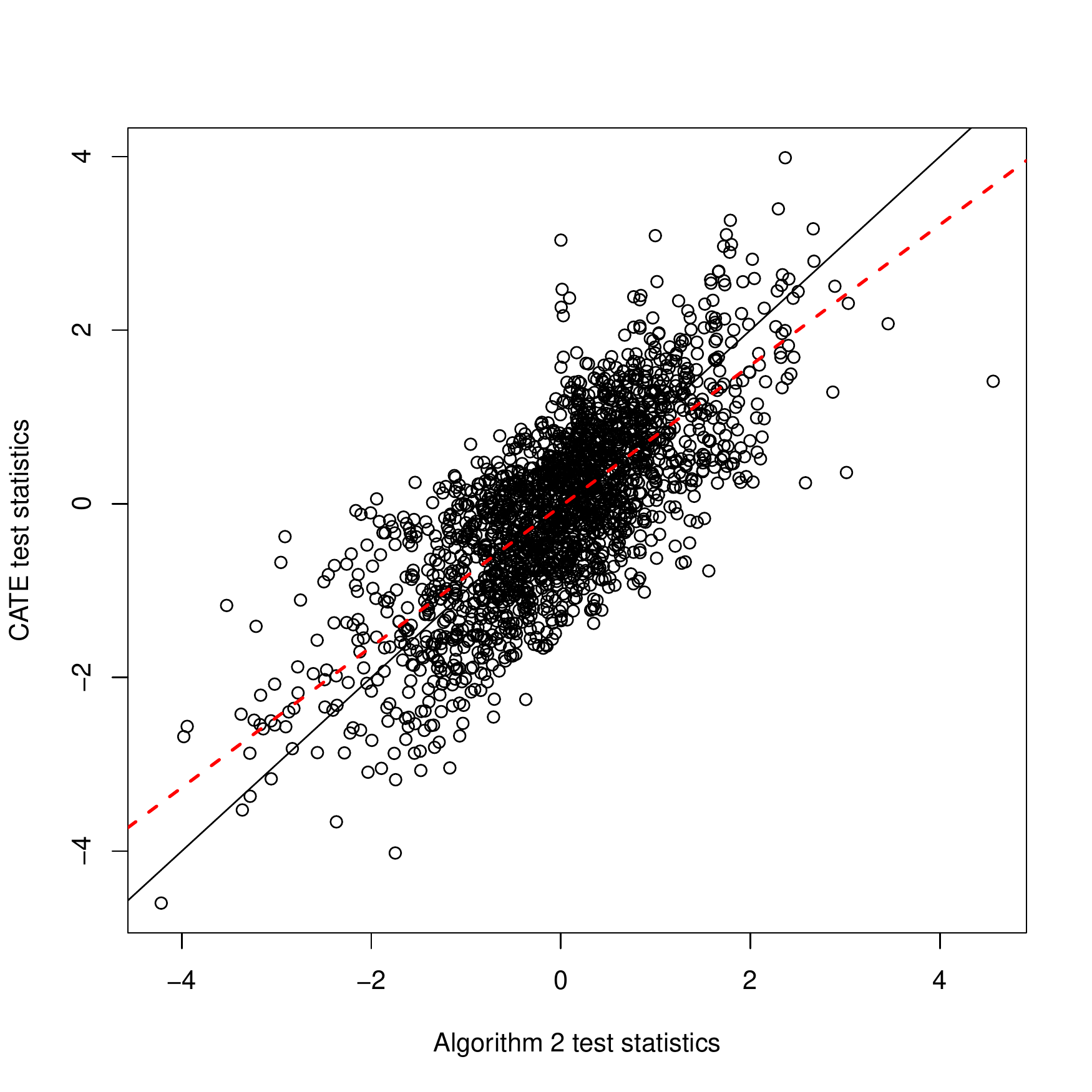}
\caption{Scatterplot of $t$-statistics for CATE and Algorithm 2 applied on the ulcerative colitis data. The 45-degree line is included in black while red dashed line is the linear fit.}
\label{fig::scatterplotCateAlg2}
\end{figure}

\subsection{Comparison on UC data} \label{secComparisonUC}

We apply both Algorithm 2 and CATE on the ulcerative colitis data to compare their respective findings on real data. Figure~\ref{fig::scatterplotCateAlg2} presents the test statistics from these algorithms. The test statistics have a correlation of $0.75$. As expected, both methods find that the bulk of genes have small test statistics. Note that the regression line of the CATE test statistics on Algorithm 2's test statistics has a slope less than 1. This implies that Algorithm 2 generates more dispersed test statistics than CATE, and, given that we have shown in Figures~\ref{fig::QQAR} and~\ref{fig::qqdata} that Algorithm 2 provides well-calibrated test statistics, that it also has more power in this situation.

Using a threshold of FDR adjusted p-values smaller than $0.1$, both methods find four genes with significant mean differences. However, there is only one gene (DPP10-AS1) that both methods identify. So, although there is significant correlation between the test statistics, the methods do not necessarily identify the same genes.

% \section{Other Comparisons} \label{secOtherComparison}

\clearpage

\bibliographystyle{ims}
\bibliography{subgaussian}{}
%\clearpage

\end{document}